\documentclass[11pt]{article}

\usepackage[preprint]{acl}

\usepackage{times}
\usepackage{latexsym}

\usepackage[T1]{fontenc}
\usepackage[utf8]{inputenc}

\usepackage{microtype}

\usepackage{inconsolata}

\usepackage{graphicx}

\usepackage{booktabs}
\usepackage{amssymb}
\usepackage{comment}
\usepackage{multirow}
\usepackage{arydshln}
\usepackage{subcaption}
\usepackage{fvextra}
\usepackage{amsmath}
\usepackage{hyperref}
\usepackage{placeins}

\title{\re{Does Every Instance Count Equally?\\Order Effects in LLM Multi-Instance Processing}}

\author{Jingxuan Chen \quad \quad Mohammad Taher Pilehvar \quad \quad Jose Camacho-Collados  \\
        School of Computer Science and Informatics, Cardiff University \\
        \texttt{\{ChenJ192,PilehvarMT,CamachoColladosJ\}@cardiff.ac.uk}}

\newcommand{\jose}[1]{{\color{black}#1}}
\newcommand{\archivejose}[1]{{\color{black}#1}}
\newcommand{\taher}[1]{{\color{black}#1}}
\newcommand{\archivetaher}[1]{{\color{black}#1}}
\newcommand{\jx}[1]{{\color{black}#1}}
\newcommand{\feedback}[1]{{\color{black}#1}}
\newcommand{\final}[1]{{\color{black}#1}}
\newcommand{\update}[1]{{\color{black}#1}}
\newcommand{\re}[1]{{\color{black}#1}}
\newcommand{\last}[1]{{\color{black}#1}}
\newcommand{\chen}[1]{{\color{black}#1}}

\begin{document}
\maketitle
\begin{abstract}
Users increasingly rely on large language models (LLMs) to analyse collections of independent inputs, such as product reviews or social media posts, and return a single aggregated answer. This setting requires multi-instance processing (MIP), where
% a model must process all instances and combine their evidence into one prediction. Since MIP inputs are unordered sets,
% the correct answer should be invariant to
% any permutation of the instances.
% \last{instance order}.
\last{a model combines information from multiple instances into one prediction}.
We
% use this property as a necessary reliability criterion and design a controlled probe to test whether current LLMs satisfy this permutation invariance prerequisite.
% Specifically \jose{in our setting},
\last{study a task in which}
% each instance requires
\last{the model must infer a binary label for each instance},
% a
% binary classification%
% into one of two labels
% , and
and
% the model must
identify the \jose{most} frequent label \jose{among the full instance set}. The correct answer should be invariant to
% any permutation of the instances.
\last{instance order}.
% \last{For example, a collection of reviews may contain more positive than negative reviews, regardless of their order.}
% Evaluating
\last{By evaluating} \re{fifteen} LLMs
% on 
% a \archivejose{simple} majority-label 
% \re{this} task 
across five datasets and seven instance order patterns\re{, with each instance set held fixed across patterns%
% ,
\last{,}}
% \re{, while holding the instances in each set fixed and varying only their order across} seven patterns,
% and seven order patterns,
\last{we} find
% that models are far from permutation invariant.
\last{substantial order sensitivity}.
Accuracy
% differs by up to 90
\last{can differ by about 90} \re{percentage points} between the best and worst order patterns, with
% majorities placed
\last{models generally more accurate when majority-label instances appear}
at the start
% being consistently easier to identify
than
% those placed
\last{when they appear}
in the middle or at the end.
% By examining the reasoning traces
% \jose{This behaviour is further confirmed through reasoning-trace} \jose{and interpretability analyses showing how LLMs tend to pay an increased attention to instances placed at the beginning, as well as other specific positions}.
% \re{This order sensitivity is accompanied by non-uniform positional patterns in both explicit instance mentions from \jose{reasoning traces} and gradient-based attribution signals.}
\jose{%
% We further analyse this behaviour by exploring LLM
% \last{Analyses of available}
\last{For subsets of models, available}
reasoning traces and
% using
% gradient-based attribution
% methods that highlight how LLMs treat instances differently depending on their position.
% \last{reveal non-uniform positional patterns in explicit instance mentions and attribution scores.}
% and
\last{gradient-based attribution reveal uneven emphasis across positions. Traces disproportionately mention early instances, while attribution scores are generally higher near the beginning and end than in the middle.}}
% \re{Available reasoning traces often mention only part of the input, frequently no more than around a quarter of the instances, despite instructions to consider them all.}
% \archivejose{that we refer to \update{descriptively} as} ``laziness''.
% Models often process only part of the input, frequently around a quarter or \archivejose{fewer instances}, despite being instructed to consider all instances.
These results show that
% current LLMs are not yet reliable aggregators for MIP, as trivial
\last{instance}
order
% changes
can systematically
% steer their
\last{bias aggregated}
predictions%
% \archivejose{\jx{in} a specific direction}.
\last{, highlighting the need to evaluate MIP across different input arrangements.}%
% PREPRINT-ONLY \footnote{Data and code are available at \update{\url{https://github.com/jingxuanchen916/LLM-Data-Understanding/tree/open-laziness}}.}
% PREPRINT-ONLY 
\footnote{Data and code are available at \update{\url{https://github.com/jingxuanchen916/mip-order}}.}
\end{abstract}

\section{Introduction}
\jose{LLMs} are increasingly used to assist users in analysing large \jose{collections} of independent inputs, such as product reviews, social media posts,
user records,
or news snippets~\citep{Sun16102025,tang2025llmagentasdataanalystsurvey}. This scenario requires multi-instance processing (MIP): the model must process each instance and aggregate
% the evidence
% \last{their information}
\last{the information}
into one prediction~\citep{bertsch2025oolongevaluatinglongcontext,chen2026understandingllmperformancedegradation}.
\last{Existing work on MIP}
% explores efficient batch processing}~\citep{cheng-etal-2023-batch,lin2024batchprompt} and
\last{reports performance degradation as instance count and context length increase}~\citep{bertsch2025oolongevaluatinglongcontext,chen2026understandingllmperformancedegradation}. 
\chen{Instance count and context length are closely related, as increasing the number of instances also lengthens the context, creating a natural overlap between MIP and long-context settings.}
% MIP also intersects with long-context settings, as more instances naturally increase context length. 

\taher{However, many long-context evaluations focus on tasks where only a small subset of the input is relevant~\citep{hsieh2024ruler,NEURIPS2020_6b493230,liu2025comprehensivesurveylongcontext}.
On such tasks, models are sensitive to where the relevant information appears~\citep{liu-etal-2024-lost}, and similar order effects have been reported for in-context learning and multiple-choice questions~\citep{cobbina-zhou-2025-show,pezeshkpour-hruschka-2024-large}.
Less is known about how the positions of differently labelled instances affect aggregation over the full set of instances.}

\last{Because these aggregated predictions often drive downstream decisions, their reliability matters in practice}~\citep{Zhu_Zhong_Zhang_Zhang_Qiao_Luo_Du_Zheng_Zhang_Chen_2026}.
\last{When 
% aggregation over an 
\chen{aggregating over the full,}
unordered set of instances, omitting} even a small fraction \last{of them} can change
\last{the correct answer, but reordering them cannot}.
% Given that these aggregated predictions frequently drive downstream decisions, ensuring their reliability is of real practical importance~\citep{Zhu_Zhong_Zhang_Zhang_Qiao_Luo_Du_Zheng_Zhang_Chen_2026}.
\last{We therefore treat permutation invariance as a necessary, though not sufficient, condition for reliable aggregation.}

\begin{figure}[t]
  \centering
  \includegraphics[width=0.95\columnwidth]{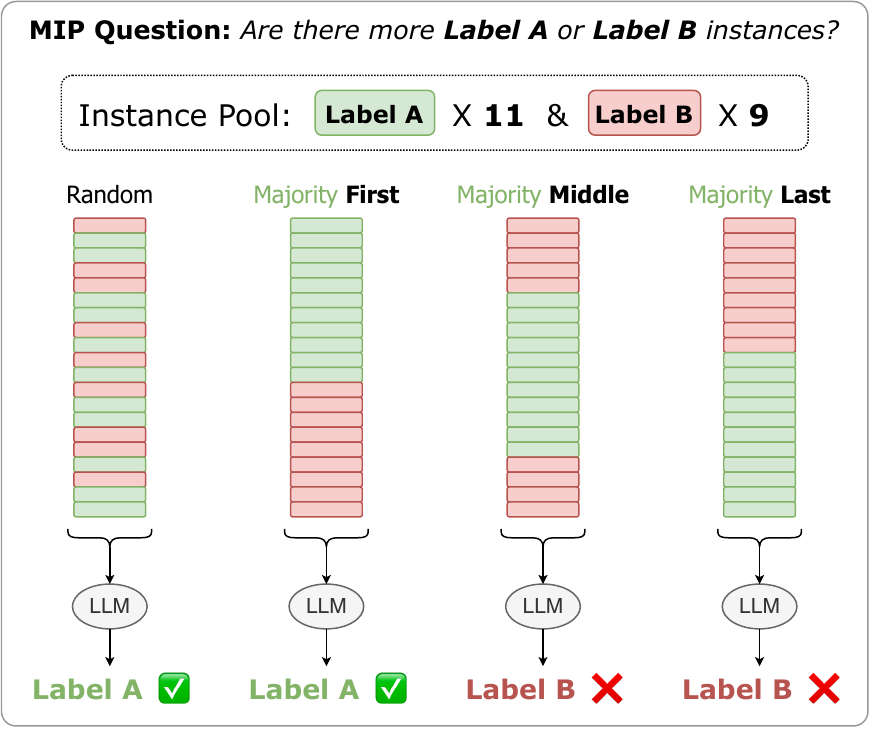}
  % \caption{A \jx{toy} example of 20 instances with two labels. \archivejose{Each instance corresponds to a text input (e.g. a movie review) and labels are binary (e.g. positive or negative).} On the left \archivejose{of the figure}, instances are randomly distributed, while on the right, the instances with the majority label (i.e. Label A) are placed in the middle.}
  \caption{An illustrative example of our MIP setting, with 20 textual instances and two \jose{pre-defined} labels, arranged in four order patterns. \jose{In this example,} the model correctly predicts the majority label when instances are randomly shuffled or majority-label instances are grouped at the beginning, but not when they are grouped in the middle or at the end. \jose{Note that while all instances are color-coded for informative purposes, LLMs receive instances without label information.}}
  \label{fig:intro}
\end{figure}

%The remaining three patterns are shown in Figure~\ref{fig:order}.

To test whether LLMs are permutation invariant in MIP, we build a controlled \update{diagnostic} probe%
% to explore \archivejose{whether they are} robust to instance order.
.
\jose{Given a set of instances
% (presented in different order patterns)
and two
\last{possible}
labels,
% the task consists of identify
\last{the model must infer each instance's label and identify}
the majority label across instances.} %We instantiate a majority-label task, where each instance \jx{carries} a binary label that the model \jx{must infer, after which it needs to answer}
% needs to infer and finally answer 
%which label is in the majority, a target that is order-independent by construction. 
Figure~\ref{fig:intro} illustrates 
% an example of our setting.
this setting with four representative instance order patterns.
We evaluate fifteen LLMs across five classification tasks, holding each instance set fixed while varying its order across seven patterns.

%(pp)
\jose{Our results suggest that}
\last{the evaluated}
LLMs are broadly not robust to instance order. Across models, accuracy
% swings by up to around 90
\last{can differ by about 90}
\re{percentage points} between the best and worst order patterns, and the effect is systematic and directional. A majority placed at the start is
% far
\last{generally}
easier to identify than one placed
% later in the \archivejose{middle}.
\last{in the middle or at the end}.
\last{This advantage remains visible across different instance counts and label distributions.}
\last{For subsets of models, we further examine reasoning traces and gradient-based attribution.}
% \taher{We find that reasoning traces often mention only a small part of the input (frequently under a quarter of instances), mostly from the beginning, and gradient-based attribution likewise drops in the middle compared with the start and end.}
We find that reasoning traces often mention only a small fraction of instances (frequently fewer than a quarter), primarily those near the beginning, while gradient-based attribution scores are lower in the middle than at the beginning and end.

\section{Related Work}
% \re{TODO: add a few latest work, intro as well?}
% NEED TO READ MORE!

% From Jose: A quick idea for the related work as a thread could be start with MIP settings (we can talk about previous paper dealing with MIP, we have several already cited in the previous paper, and in fact we should cite ours too), but then say that while they have studied the limitations of LLMs in MIP, they haven't studied the impact of instance order, which is what we do here. And then this can lead to the next subsection where we can talk about the retrieval lost-in-the-middle papers, and mention that our paper fills the gap and explain the differences 
\paragraph{Multi-Instance \archivejose{Processing}.}
\archivejose{Several} lines of work study \archivejose{LLM} prompts that contain multiple independent inputs, problems, or evidence units.
One line of research
\jx{%
% frames such inputs primarily through the lens of
\last{studies} batch processing and multi-task inference}
% , which investigate whether
\last{to process}
multiple items
% can be processed
jointly
% to reduce repeated instructions, API calls, or inference cost
\last{more efficiently}
while
% largely
preserving task performance~\citep{cheng-etal-2023-batch,lin2024batchprompt,kolawole2026parallelprompt,son-etal-2024-multi-task}.
% \last{These settings generally require separate predictions, and some also examine order sensitivity~\citep{lin2024batchprompt}.}
Another line
% treats aggregation as one task inside broader long-context benchmarks, includ\jx{ing} tasks requiring
\last{examines}
information fusion, aggregation,
% or
\last{and}
deep reasoning
% over long inputs
\last{in long-context benchmarks}~\citep{bai-etal-2025-longbench,hsieh2024ruler,qiu-etal-2024-clongeval,shaham-etal-2023-zeroscrolls}.
The \jose{studies closest to ours} measure accuracy degradation as instance count and context length increase%
% , but crucially, it does not investigate \re{the impact of} instance order
~\citep{bertsch2025oolongevaluatinglongcontext,chen2026understandingllmperformancedegradation}.
% \last{These include tests of random shuffling~\citep{chen2026understandingllmperformancedegradation}.}
\jose{In our case, we isolate 
% the instance order factor as a cause for scoring discrepancies 
\chen{the effect of instance order on aggregation accuracy}
in MIP.
} %settings

%Compared to them, our setting instead isolates a clean MIP regime,
%\jose{C}
%

\paragraph{Position Bias and Lost-in-the-Middle Effects.}
A substantial body of work shows that LLMs are highly sensitive to where relevant information appears in the input.~\citet{liu-etal-2024-lost} established the ``lost-in-the-middle'' effect, showing that
\re{LLMs can struggle to access task-relevant information when it appears in the middle of a long context rather than near the beginning or end.}
Subsequent work
% broadened this picture by studying multi-evidence and position-aware settings, including the effects of
\last{examines}
absolute position, relative spacing between relevant pieces, and multi-evidence retrieval~\citep{gao-etal-2024-insights,hsieh-etal-2024-found,song-etal-2025-counting,tian-etal-2025-distance,veseli2025positional}.
Related order effects also arise outside retrieval\jx{:} few-shot in-context learning is sensitive to demonstration order, and multiple-choice reasoning is brittle to option permutations and selection bias~\citep{cobbina-zhou-2025-show,pezeshkpour-hruschka-2024-large,zheng2024large}.
\last{Recent work examines order sensitivity in retrieval and multimodal settings~\citep{NEW_FacetProbe2026,NEW_StableRAG2026} and explores training and architectural approaches to order robustness~\citep{NEW_SetLLM2025,NEW_DGAO2026}.} \last{Instead, our work focuses on aggregation tasks that require considering information from every instance in the final judgement. Although the correct answer is independent of instance order, we show that the positions of instances with different labels can systematically bias the model's prediction.}

\section{\re{Experimental Setting}}
\label{sec:setting}
% \section{Are LLMs Robust to Instance Order?}
% \label{sec:main_results}

% \feedback{the main thing: if order has an effect}
% \feedback{simplify the results - only around instance order}

%\archivejose{The main goal of our experiments is to test whether instance order has an effect}
% - statistical significant, t-test, mistake is not due to the chance (though 55 and 45 close, actually 200 instances difference)
% We have three runs
% - any distribution, given std

% \subsection{Performance Analysis}

%\archivejose{We used the experimental setting de}

To study
% if
\last{whether} LLMs can \archivejose{perform MIP robustly under instance permutations},
% \last{how instance order affects majority-label predictions},
we use five heterogeneous \archivejose{classification} tasks.
% . \archivejose{The common goal in our setting is to find the} majority label \archivejose{across a fixed instance pool.} %to evaluate fourteen LLMs.
\archivejose{The goal in our setting is to find the} majority label
\archivejose{across a fixed instance pool}. Within each pool, we present the instances in \last{different} order patterns (cf. Section \ref{sec:instance_construction}).
% \last{, keeping each instance set fixed across order patterns}.
% Within each pool, we present the instances in various order patterns, expecting similar accuracy across all of them if the LLMs are truly permutation invariant.

% \subsection{Experimental Setting}
% \label{sec:setting}

%understand that instances are permutation invariant

\subsection{Individual Tasks}
We design five \archivejose{classification} tasks from existing datasets: \textit{Math} 
% identifying odd and even numbers
\jose{(\chen{arithmetic reasoning})}~\citep{cobbe2021trainingverifierssolvemath}, \textit{NER} ({n}amed-{e}ntity {r}ecognition)~\citep{rahimi-etal-2019-massively},
% \footnote{For task simplicity, we only sample instances that contain exactly one entity, either PER(son) or LOC(ation).}
\textit{Occurrence} \jose{(\chen{target}-word counting)}~\citep{barbieri-etal-2020-tweeteval}, \textit{Sentiment} \chen{(sentiment \jose{analysis})}~\citep{socher-etal-2013-recursive} and \textit{WSD} ({w}ord {s}ense {d}isambiguation)~\citep{10.1162/coli_a_00405}. \jose{The tasks and datasets are selected so as to be diverse, relatively simple and solvable by current LLMs \chen{when each instance is processed individually}.} 
% in the single instance setting
We treat each instance within a given task as an independent binary classification problem.
In the MIP setting, we ask the model to identify the majority label across all instances%
% , allowing it to output \jx{``}tie\jx{''} if both labels appear with equal frequency.
.
Table~\ref{tab:tasks-labels}
% shows the \chen{overview of}
\chen{provides an overview of the five tasks}%
% information about
% \last{lists the labels for}
% each task%
% (
\last{, with}
\archivejose{further} details
% \archivejose{about each specific task} can be found
in Appendix~\ref{append:task}%
% )
.
%\taher{We evaluate 2,500 instances per task, split evenly across the two labels.}
\taher{To reduce individual-instance difficulty as a confound, we keep only instances that both \textit{Qwen3.6-35B-A3B} and \textit{Gemini 3.1 Flash-Lite} classify correctly in isolation (\chen{see} Appendix~\ref{append:filtering} for more details on filtering).}

% \todoTAB{task information, name, content, label}
\begin{table}
  \centering
  \scalebox{0.83}{
  \begin{tabular}{lll}
    \toprule
    \textbf{Name} & \textbf{Target} & \textbf{Label \chen{A}} : \textbf{Label \chen{B}} \\
    \midrule
    {Math} &
    {Answer} &
    odd\_number : even\_number \\
    
    NER & Entity & PER : LOC \\

    {Occurrence} &
    {\jx{``}the\jx{''}} &
    exactly\_once : more\_than\_once \\

    Sentiment & Polarity & positive : negative \\

    {WSD} &
    {\jx{``}java\jx{''}} &
    programming\_language : place
     \\

    \bottomrule
  \end{tabular}
  }
  \caption{Overview of our binary-labelled tasks.
  % where the MIP objective is to identify the majority label. 
  \chen{Each instance's target is assigned either Label A or Label B, and the MIP objective is to identify the majority label across all instances.}
  % \jx{For example, the \textit{Math} task requires the LLM to 
  % \chen{solve an arithmetic word problem and}
  % classify \chen{its} answer as either ``odd\_number'' or ``even\_number''.} 
  }
  \label{tab:tasks-labels}
\end{table}

%More details \archivejose{of each task} can be found in Appendix~\ref{append:task}.

\subsection{Models and Prompting}
% \re{TODO -- may change a3b to local version as well}
We evaluate \re{fifteen} LLMs, including six \re{group}\jx{s of} models from the same \jose{family}: (\textit{DeepSeek-V4-Flash}, \textit{DeepSeek-V4-Pro}), (\textit{Gemini 3.1 Flash-Lite}, \textit{Gemini 3.1 Pro}), (\textit{Gemma 4 26B A4B}, \textit{Gemma 4 31B}), (\textit{gpt-oss-120b}, \textit{GPT-5.4 nano}), (\textit{Llama 4 Scout}, \textit{Llama 4 Maverick}), and (\re{\textit{Qwen3-4B-Instruct},} \textit{Qwen3.6-35B-A3B}, \textit{Qwen3.6-Plus});\footnote{\chen{Unlike other models, \textit{Qwen3.6-Plus} has separate reasoning and answer token budgets, so it is forced to produce a final answer even when its reasoning budget is exhausted %though such answers may be less accurate 
(cf. 
Appendix~\ref{append:error_analysis}).}} and two other LLMs: \textit{Grok 4.1 Fast} and \textit{MiniMax M2.7} (more details in Appendix~\ref{append:model_configuration}).

For \re{generation}, we use a temperature of 0 and a maximum output length of 20K tokens,\footnote{
\last{We 
% also examine a 100K-token budget for \textit{Qwen3.6-35B-A3B} in 
\chen{additionally test a 100K-token output budget
for \textit{Qwen3.6-35B-A3B} to examine the effect of
allowing longer generations. The invalid rate decreases substantially, although order sensitivity remains
pronounced (cf.}
Appendix~\ref{append:budget_sensitivity}).}
}
% \footnote{\re{For \textit{Qwen3.6-Plus}, the 20K-token output budget is split into 20 completion tokens and 19,980 reasoning tokens.}} for consistency across all models.
\last{including reasoning tokens where applicable}. %\footnote{\update{
% Some providers accessed through OpenRouter did not consistently
% enforce this limit: 213 of 3,780 core experiments returned
% completions exceeding 20K tokens. These cases were not
% concentrated in any particular order pattern and are therefore
% unlikely to account for the observed order-pattern gaps; see
% Appendix~\ref{append:error_analysis}.
% }} 
% To allow limited tolerance for invalid model responses, we permit up to three attempts when a model fails to produce a valid output.
\last{We permit up to three attempts in total, retaining the first valid output.}
% The full set of prompting templates is provided in Appendix~\ref{append:prompt}.
\feedback{Unless otherwise specified, we use}
% \final{along with}
the \re{standard} prompts  \jose{that simply state the task and pre-defined labels, and require models to find the majority label among the input instance set. \jose{Full prompts are available} in Appendix~\ref{append:general_prompt}}.
% . %in %Appendix~\ref{append:general_prompt}.

%%Taher: I removed the whole subsection (and added a line to the end of 3.1)

%\subsection{Multi-Instance Input Construction}
% \subsection{\jose{Order Patterns}}
\subsection{\chen{Instance Set Construction and Ordering}}
\label{sec:instance_construction}
% We sample from the filtered instances to construct MIP inputs. 
Each instance set $\mathcal{X}$ is constructed \chen{from filtered instances} according to: the instance count $n$ (i.e., $n = |\mathcal{X}|$), the majority proportion $p$, the majority label $y_{\mathrm{maj}}$, and the instance selection seed $s$. Given $n$ and $p$, we use $s$ to sample $n_{\mathrm{maj}} = p \cdot n$ instances with label $y_{\mathrm{maj}}$ and $n_{\mathrm{min}} = (1-p) \cdot n$ instances with the corresponding minority label $y_{\mathrm{min}}$. We then hold $\mathcal{X}$ fixed and vary the order pattern $\pi$ across seven predefined choices to construct the experimental inputs.

\paragraph{\jose{Order patterns.}} \jose{We explore seven order patterns that specifically place labelled instances across different positions in the input.} 
% We then arrange the instances in $\mathcal{X}$ according to each of seven order patterns $\pi$, always using the same selected instances.
\jose{First,} the \textit{random} pattern \jose{acts as a baseline and} shuffles the full instance set \jose{randomly} using a separate order seed $s_{\mathrm{order}}$.
Four patterns group instances into contiguous blocks by label.
\textit{Majority-first} places all majority instances before the minority ones, and \textit{majority-last} reverses this order.
\textit{Majority-edges} splits the majority instances into two halves placed around a central minority block, and \textit{majority-middle} instead places the majority block between two halves of the minority instances.
\jose{The \chen{remaining} two are} \textit{soft} variants that interleave the labels.
They \chen{form an even-length alternating sequence of majority--minority pairs} and place the surplus majority instances either at both ends of this sequence (\textit{soft-majority-edges}) or in its middle (\textit{soft-majority-middle}).
\jose{Figure \ref{fig:intro} provides a visual illustration of four representative patterns (\textit{random}, \textit{majority-first}, \textit{majority-last} and \textit{majority-middle}), while}
Appendix~\ref{append:order_pattern} provides formal definitions of all seven patterns.

\subsection{Evaluation Metric}
\label{sec:eval_metrics}
For each experiment $e$, we \last{attempt to parse} the model output $o_e$
into a predicted majority label $\hat{y}_e$.
\last{If no label can be parsed within the allowed attempts, the experiment is invalid.\footnote{A parseable ``tie'' is treated as valid but incorrect.} Otherwise, we compare $\hat{y}_e$ with the}
ground-truth majority label $y_{\mathrm{maj}}^{(e)}$ \last{and consider the experiment}
correct if $\hat{y}_e = y_{\mathrm{maj}}^{(e)}$.
We report \jose{valid-only} \textbf{accuracy (\chen{$\mathrm{Acc}$})} as the fraction of \re{valid} experiments that produce a correct answer within each selection seed.
Overall accuracy pools runs across tasks, majority labels and order patterns. We then report its mean and standard deviation across selection seeds.
\re{\last{We select the best and worst order patterns, $\pi^{+}$ and $\pi^{-}$, using their mean accuracies. We compute their gap as
$\Delta_{\chen{best-worst}} = \underset{s}{\operatorname{mean}}
\bigl[\mathrm{Acc}_{s}(\pi^{+})-\mathrm{Acc}_{s}(\pi^{-})\bigr]$,
where $\mathrm{Acc}_{s}(\pi)$ is the accuracy of pattern $\pi$ at selection seed $s$.
% The mean and standard deviation are computed over seeds with valid results for both patterns.
% The mean uses only seeds with valid results for both patterns. 
Larger values indicate greater variation in accuracy across order patterns.}}
% \last{Further details are provided in Appendix~\ref{append:metric_details}.}
% the weighting and missing-value rules.}
% Appendix~\ref{append:invalid_rate} also reports accuracy with invalid experiments counted as incorrect.}
% Overall part in append:invalid_rate

\subsection{Core Setting}
% Given that the research focus\jx{es on} permutation invariance in MIP, our
\last{Our}
\textit{core} setting is based on $n = 2, 000$ instances with proportion $p =0.55$%
% \archivejose{(i.e.
% 55\% of instances with label 1 ($y_{maj}$) and 45\% of instances with label 0
% \jx{instances with the majority label account for 55\% of all instances})}
. 
Although this setting is near-balanced in percentage terms, the majority signal
is still well-defined. It corresponds to $n_{\mathrm{maj}}=1100$ and
$n_{\mathrm{min}}=900$, giving an absolute margin of 200 instances. 
% As a
% conservative reference point, even if independent instance-level predictions had
% only $90\%$ accuracy,\footnote{However, \jx{since we only keep the filtered instances,} we expect all LLMs to achieve nearly perfect accuracy for each individual instance after our filtering procedure as described in Section~\ref{sec:filtering}.} the expected predicted-majority count would still be 1080,
% about 6 standard deviations above the 50/50 tie boundary at 1000. 
\chen{In probabilistic terms, the chance of reversing the majority under a 55/45 distribution, even with a model predicting at only 90\% accuracy,\footnote{However, \jx{since we only keep the filtered instances,} we expect all LLMs to achieve nearly perfect accuracy for each individual instance after our filtering procedure.}
% as described in Section~\ref{sec:filtering}.} 
is on the order of one in a billion (assuming independent instance-level predictions with equal accuracy across instances). We provide the calculation in Appendix~\ref{append:majority_sanity_check}.}
% \last{Appendix~\ref{append:majority_sanity_check} gives an illustrative calculation showing that the expected predicted-majority count remains well above the tie boundary under independent classification errors with a common accuracy of 90\%.}

\archivetaher{
% ??
% \archivejose{For each experiment}, we 
\jx{We conduct each experiment twice by swapping which binary label serves
% use each of the two binary labels 
as $y_{\mathrm{maj}}$. For each configuration, we test all seven order patterns over three} selection seeds $s \in \{0,1,2\}$%
% to compute the standard deviation
.}
% (used for the standard deviation calculation), and all seven order patterns;} for 
\jx{For the} \textit{random} \jx{pattern}, we additionally vary the order seed \last{three times with} $s_{\mathrm{order}} \in \{0,1,2\}$
% \archivejose{%to % randomly 
% \jx{using each seed to shuffle the} instance \jx{set and 
to obtain a more robust estimate under random orderings.
% } three times to provide a \jx{more robust} estimate of \jx{pure} random \jx{orderings}}. 

% We conduct every experiment twice, swapping which of the two binary labels serves as the majority (....).  selection seeds (s {0,1,2}) 
% For the \textit{random} pattern, we additionally vary the order seed (sorder {0,1,2}), shuffling the instance set three times to provide a more robust estimate of a purely random distribution. 

%\archivejose{three independent runs with different samples with} 

%\section{Main Results: 55/45 Majority Split with 2,000 Instances}
% \feedback{use RQ instead? naming stuff, or two sections}

\begin{figure*}[t]
  \centering
  \includegraphics[trim={0 1.5cm 0 3cm},width=0.94\textwidth]{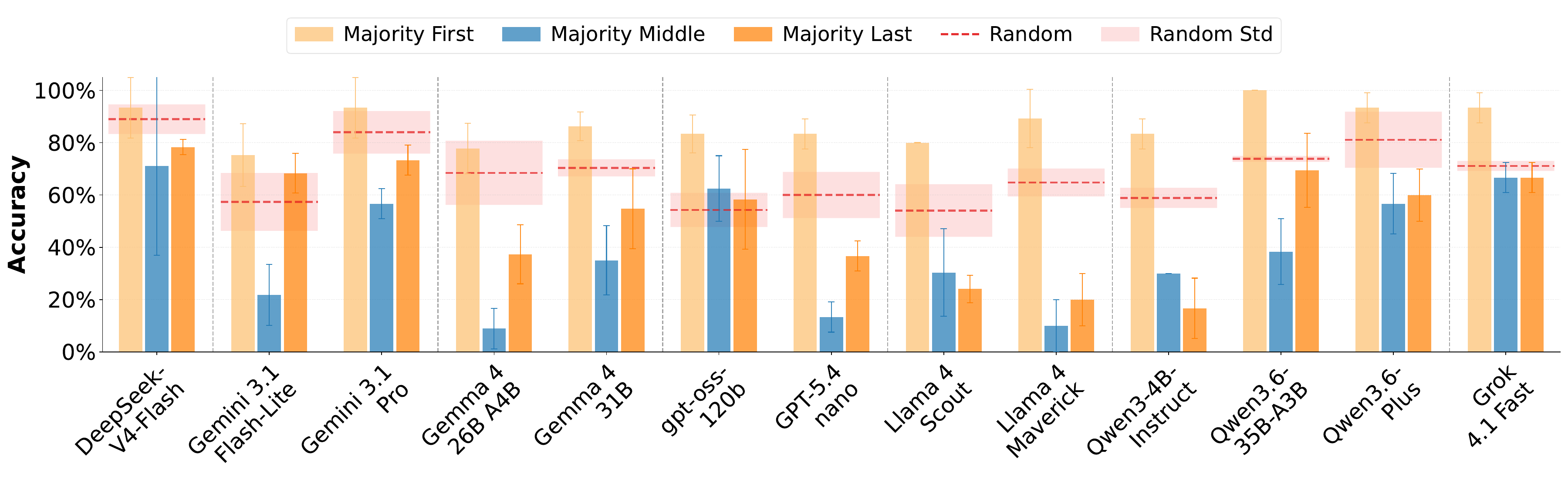}
  \caption{Per-model accuracy on four representative order patterns in the \textit{core} setting. \re{Error bars indicate standard deviations over three instance-selection seeds.}}
  \label{fig:four_pattern_model_acc}
\end{figure*}

\section{\re{Robustness of LLMs to Instance Order}}
\label{sec:main_results}

% In this section, we explore the sensitivity of LLMs to the order in which a set of instances are presented to them, and in particular to specific order patterns where instances with different pre-fixed labels are placed in strategic positions across the instance sets (see Section \ref{sec:instance_construction} for more details on these order patterns).

We examine LLM sensitivity to instance order, focusing on specific order patterns that place instances with different labels at predefined positions within the input, \jose{as explained in Section~\ref{sec:setting}.} %(see Section \ref{sec:instance_construction} for more details on these order patterns).

% \re{TODO: trim}
\subsection{\re{Core} \archivejose{Results}}
\label{sec:core_performance}

%\paragraph{Performance \re{Overview} \jose{and General Variation}.}
% \paragraph{Overall LLM Performance}
Table~\ref{tab:overall-success-pattern-gap} reports each model's accuracy \jose{averaged across tasks and order patterns}\last{, invalid rate,} and the corresponding \chen{accuracy} gap \last{$\Delta$} between its best and worst order patterns.
\re{A}ccuracy varies widely across models.
% Because our central question is whether LLMs are robust to instance order, the relevant comparison is the gap between order patterns.
% \last{For the remaining thirteen models,} 
\chen{Despite this variation, all models exhibit accuracy gaps at least 40.0 percentage points (pp), with models such as \textit{Gemma 4 26B A4B} and \textit{Llama 4 Maverick} reaching around 90.0 pp.}
% the accuracy gap ranges from \re{40.0} to \re{91.1} \jose{percentage points} \chen{(pp)}.
% \footnote{\last{We omit these two models from the pattern plots and generalisability analysis. The latter also excludes} \textit{Grok 4.1 Fast}\last{, which was deprecated during experimentation.}}
\re{Such} gap\re{s are} substantial across the board, \jx{and \re{they} exceed the corresponding variation across instance-selection seeds, indicating that the observed order sensitivity \re{is unlikely to} be explained by random variation over instance sets alone.}
%%%
\textit{DeepSeek-V4-Pro} and \textit{MiniMax M2.7}
\chen{have invalid rates near or above 90\%},
\chen{as they often exhaust the 20K-token output budget.
Their accuracy and $\Delta$ are therefore based on
relatively few valid experiments.}\footnote{\chen{As a result, \textit{DeepSeek-V4-Pro} and \textit{MiniMax M2.7} are excluded from subsequent experiments and analyses.}}
% % \textit{Grok 4.1 Fast}\last{, which was deprecated during experimentation, is included only in the \textit{core} setting. 
% The generalisability analysis \jose{and remaining experiments} therefore use \chen{thirteen} models. Section~\ref{sec:why_not} also excludes all three models.}
%%%
\last{In addition, error breakdowns, results} \last{treating invalid experiments as incorrect, and per-task accuracies are provided in Appendices~\ref{append:error_analysis}, \ref{append:invalid_rate}, and \ref{append:per_task_acc}.}

\begin{table}[t!]
  \centering
  \setlength{\tabcolsep}{3pt}
  \scalebox{0.85}{%
    \begin{tabular}{@{}lccc@{}}
\toprule
{\textbf{Model}} &
% \textbf{Acc. (\%)} &
% $\mathbf{Acc}_{\mathbf{valid}}$ &
$\mathbf{Acc}$ &
% \textbf{$\Delta$ Order Patt.} \\
% $\boldsymbol{\Delta}_{\mathbf{valid}}$ \\
$\boldsymbol{\Delta}_{best-worst}$ &
\textbf{Invalid} \\
\midrule

\re{DeepSeek-V4-Flash}
& 79.8{\footnotesize $\pm$6.9}
& 64.0{\footnotesize $\pm$16.0}
& 40.0{\footnotesize $\pm$5.8} \\

\re{DeepSeek-V4-Pro}
& 78.6{\footnotesize $\pm$25.8}
% & 50.0{\footnotesize $\pm$0.0}
& 50.0\last{$^{\dagger}$}
& 88.9{\footnotesize $\pm$4.4} \\

\cmidrule(lr){1-1}

Gemini 3.1 Flash-Lite
& 63.0{\footnotesize $\pm$4.1}
& 67.4{\footnotesize $\pm$12.2}
& 10.7{\footnotesize $\pm$3.4} \\

Gemini 3.1 Pro
& 81.7{\footnotesize $\pm$4.2}
& 40.0{\footnotesize $\pm$0.0}
& ~~1.1{\footnotesize $\pm$1.1} \\

\cmidrule(lr){1-1}

Gemma 4 26B A4B
& 64.0{\footnotesize $\pm$6.9}
& 91.1{\footnotesize $\pm$7.8}
& 28.1{\footnotesize $\pm$2.3} \\

Gemma 4 31B
& 68.2{\footnotesize $\pm$0.3}
& 51.3{\footnotesize $\pm$12.0}
& 10.4{\footnotesize $\pm$6.3} \\

\cmidrule(lr){1-1}

gpt-oss-120b
& 63.0{\footnotesize $\pm$1.5}
& 43.5{\footnotesize $\pm$9.8}
& 21.9{\footnotesize $\pm$0.6} \\

GPT-5.4 nano
& 58.1{\footnotesize $\pm$3.4}
& 70.0{\footnotesize $\pm$10.0}
& ~~0.0{\footnotesize $\pm$0.0} \\

\cmidrule(lr){1-1}

\re{Llama 4 Scout}
& 55.3{\footnotesize $\pm$2.9}
& 55.9{\footnotesize $\pm$5.3}
& ~~1.5{\footnotesize $\pm$1.7} \\

Llama 4 Maverick
& 61.0{\footnotesize $\pm$2.0}
& 90.0{\footnotesize $\pm$10.0}
& ~~2.2{\footnotesize $\pm$1.1} \\

\cmidrule(lr){1-1}

\re{Qwen3-4B-Instruct}
& 57.4{\footnotesize $\pm$6.1}
& 66.7{\footnotesize $\pm$5.8}
& ~~0.0{\footnotesize $\pm$0.0} \\

\re{Qwen3.6-35B-A3B}
& 75.5{\footnotesize $\pm$3.1}
& 61.7{\footnotesize $\pm$12.6}
& 33.7{\footnotesize $\pm$5.3} \\

\re{Qwen3.6-Plus}
& 75.2{\footnotesize $\pm$4.2}
& 50.0{\footnotesize $\pm$0.0}
& ~~0.0{\footnotesize $\pm$0.0} \\

\cmidrule(lr){1-1}

Grok 4.1 Fast
& 72.2{\footnotesize $\pm$2.9}
& 46.7{\footnotesize $\pm$15.3}
& ~~0.0{\footnotesize $\pm$0.0} \\

\cmidrule(lr){1-1}

MiniMax M2.7
& 36.1{\footnotesize $\pm$19.6}
& 100.0{\footnotesize $\pm$0.0}\last{$^{\ddagger}$}
& 95.2{\footnotesize $\pm$5.0} \\

\bottomrule
\end{tabular}
}
  % \caption{\re{Valid-only} accuracy and \re{INVALID RATE (\%)} the $\Delta$ between the corresponding accuracies of best and worst order patterns.}
\caption{Accuracy \last{(Acc, \%), best-worst order-pattern gap ($\Delta$, \chen{pp}), and invalid rate (\%) in the \textit{core} setting. Values are means $\pm$ standard deviations over three instance-selection seeds unless marked. Gaps marked $\dagger$ use one paired seed (no standard deviation), and $\ddagger$ use two.}}
  \label{tab:overall-success-pattern-gap}
\end{table}

\paragraph{\final{Results by Instance Order Patterns}.}
%\last{With} \textit{random} \last{ordering} as \last{the} baseline, \last{Figure~\ref{fig:four_pattern_model_acc} shows} a systematic positional bias \last{for each model}. 
\jose{Figure ~\ref{fig:four_pattern_model_acc} shows 
% the results of each LLM by 
\chen{the accuracy of each LLM across four representative} order patterns.}\footnote{\chen{Since several patterns show similar trends,
% the rest of \jx{this section} focuses
\last{Figure~\ref{fig:four_pattern_model_acc} and Section~\ref{sec:count_proportion}
% the pattern plots in Sections~\ref{sec:core_performance} and~\ref{sec:count_proportion} 
focus}
on four representative patterns, including \textit{random}, \textit{majority-first}, \textit{majority-middle} and \textit{majority-last}, with \chen{results for all seven patterns provided} in the appendix. }}
Models \last{generally} achieve their highest accuracy under the \textit{majority-first} pattern\last{, while} \chen{accuracy} under \textit{majority-last} \jose{and particularly} \textit{majority-middle} 
% \last{are more likely to} drop
\chen{often falls}
below the \textit{random} baseline.
This systematic \jx{variation shows} that the models' predictions are 
% heavily skewed
\chen{strongly influenced}
by where the majority signal is placed, rendering them 
% highly brittle
\chen{sensitive}
to order permutations.
Although no single pattern yields the lowest accuracy across all \jx{plotted} LLMs, \jose{over} half of the models perform worst under \textit{majority-middle}, suggesting a \archivetaher{\jx{``}lost-in-the-middle\jx{''} effect}~\citep{liu-etal-2024-lost}. 
% This systematic \jx{variation shows} that \last{reordering the same instances can steer the model's aggregate prediction.}
\last{Full results across tasks and all seven patterns are provided in Appendix~\ref{append:all_pattern_acc}.}

\subsection{\archivejose{\jx{Generalisability} Analysis:} Instance Count and \archivejose{Label Distribution}}
\label{sec:count_proportion}
% Our \final{main results were based on our} \textit{core} setting, \final{which is} 
% % based on 
% 2,000 instances with the 55/45 \jx{label distribution}. %, which is challenging for many LLMs. 
Our
% \final{main results were based on our}
\textit{core} setting
% , \final{which is}
\last{uses} 2,000 instances with
% the
\last{a} 55/45 \jx{label distribution}.
To investigate whether the \jx{observed order effects}
% observations made on the effect of order pattern 
generalise \final{beyond this specific setting}, we \jx{vary} the instance count (100, 200, 500, 1,000 and 2,000), and the \jx{label distribution} (\chen{majority/minority:} 90/10, 70/30, 60/40 and 55/45). 
The setting otherwise follows that of the \textit{core}, except that we use a single instance-selection seed ($s = 1$) to reduce computational cost.
% , and the \textit{random} \chen{pattern still uses the same three order seeds for shuffling}.

% Per-model results are given in Appendix~\ref{append:model_performance}.

%\paragraph{Order Pattern.}
% From Jose: Again, in this heatmap at the bottom I would call "label distribution" as more intuitive if we go for this slight name change
\begin{figure*}[t]
  \centering
  % AAA \includegraphics[trim={0 1cm 0 0},clip,width=0.95\textwidth]
  \includegraphics[width=0.94\textwidth]{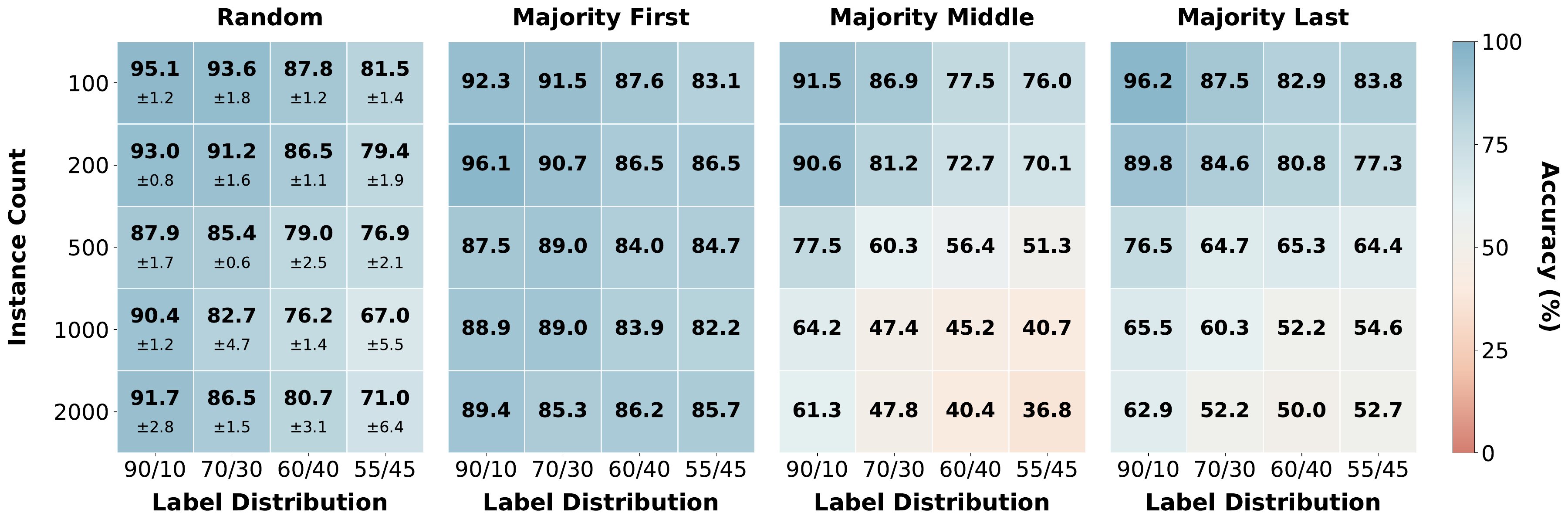}
  \caption{\jose{Accuracy} over different instance counts and label \jx{distributions}, for the four representative patterns. Standard deviation is computed over three order seeds for \textit{random} only.}
  \label{fig:count_proportion_acc}
\end{figure*}

Figure~\ref{fig:count_proportion_acc} shows accuracy across instance counts and \final{label} distributions for the four representative order patterns.
% As expected, accuracy increases as the instance count decreases and as the \jx{distribution} becomes more skewed.
The order effects observed in Section~\ref{sec:core_performance}
% holds across almost \re{all} combination\re{s, including the ones with only 100 instances}.
\last{remain visible across these settings,} \re{including those with only 100 instances.}
The \textit{majority-first} pattern
% \re{usually} remains the strongest, outperforming both the \textit{random} baseline and \final{particularly} the late-occurring \re{majority} patterns (\jx{i.e. }\textit{majority-middle} and \textit{majority-last}).
\last{generally outperforms} \textit{majority-middle} and \textit{majority-last}\last{, while its advantage over the} \textit{random} \last{baseline is clearest for more balanced label distributions.}
% Furthermore, this performance gap widens systematically:
\last{As} the instance count grows and the label \jx{distribution} becomes more balanced, accuracy under the \textit{majority-middle} and \textit{majority-last} configurations
% degrades sharply,
\last{generally declines,}
whereas \textit{majority-first}
% experiences only a mild decline.
\last{remains comparatively stable. The performance gap between these patterns therefore tends to widen.}
\update{The
% strongly
 directional nature of this effect suggests that models may rely disproportionately on instances appearing early in the context, motivating the behavioural analysis in Section~\ref{sec:why_not}.}
Full results for all seven patterns are given in Appendix~\ref{append:count_proportion_all}.

%\subsection{\re{Can Prompting Improve Robustness to Permutation?}}
\subsection{\jose{Prompting Variants}}
\label{sec:prompt_variant_can}
% \re{mentioning all different prompt strategies we have done, not splitting 2000 instances}
% Prompt can be found in Appendix~\ref{append:add_prompt}.
% %\paragraph{Prompt-Based Hints.}
% Since our analysis identifies instance skipping as a 
% % primary cause of 
% \update{recurring behaviour associated with} failure, 
% \update{limited explicit instance coverage as a recurring behaviour associated with failure,} we test whether explicitly reminding the models to consider every instance helps. Keeping the core setting, we add a hint prompt block instructing that the majority label must be based on all instances rather than a subset. The hint yields no consistent improvement and may even degrade accuracy. \update{It also produces only limited and inconsistent changes in instance coverage, with full coverage remaining rare.} \final{These results} indicate that 
% % these systemic MIP failures cannot be trivially resolved 
% \update{the failures observed in our controlled setting are not consistently resolved} through simple prompt engineering, \final{although this may be an interesting area for future work}. We provide the details \final{of this analysis, together with detailed results,} in Appendix~\ref{append:prompt_hints}.

\re{To assess whether prompting can reduce order sensitivity,
% for three models,
we compare the standard prompt with three prompting variants \last{for three models and all patterns in the \textit{core} setting,}
% while
keeping the input instances and their orderings unchanged:} 
\re{(i) \textit{hint prompting}, which
emphasises processing} all instances rather than relying on a subset;
\re{(ii) \textit{suggested strategy prompting}, which
additionally suggests processing instances in groups and summing the group-level label counts;
(iii) \textit{self-prompting},
where the model first generates its own task instructions and then uses them to solve the task.\footnote{\re{Both generation stages \chen{of \textit{self-prompting}} have a separate maximum output length of 20K tokens.}}
}

\begin{figure}[t]
  \centering
  \includegraphics[width=0.95\columnwidth]{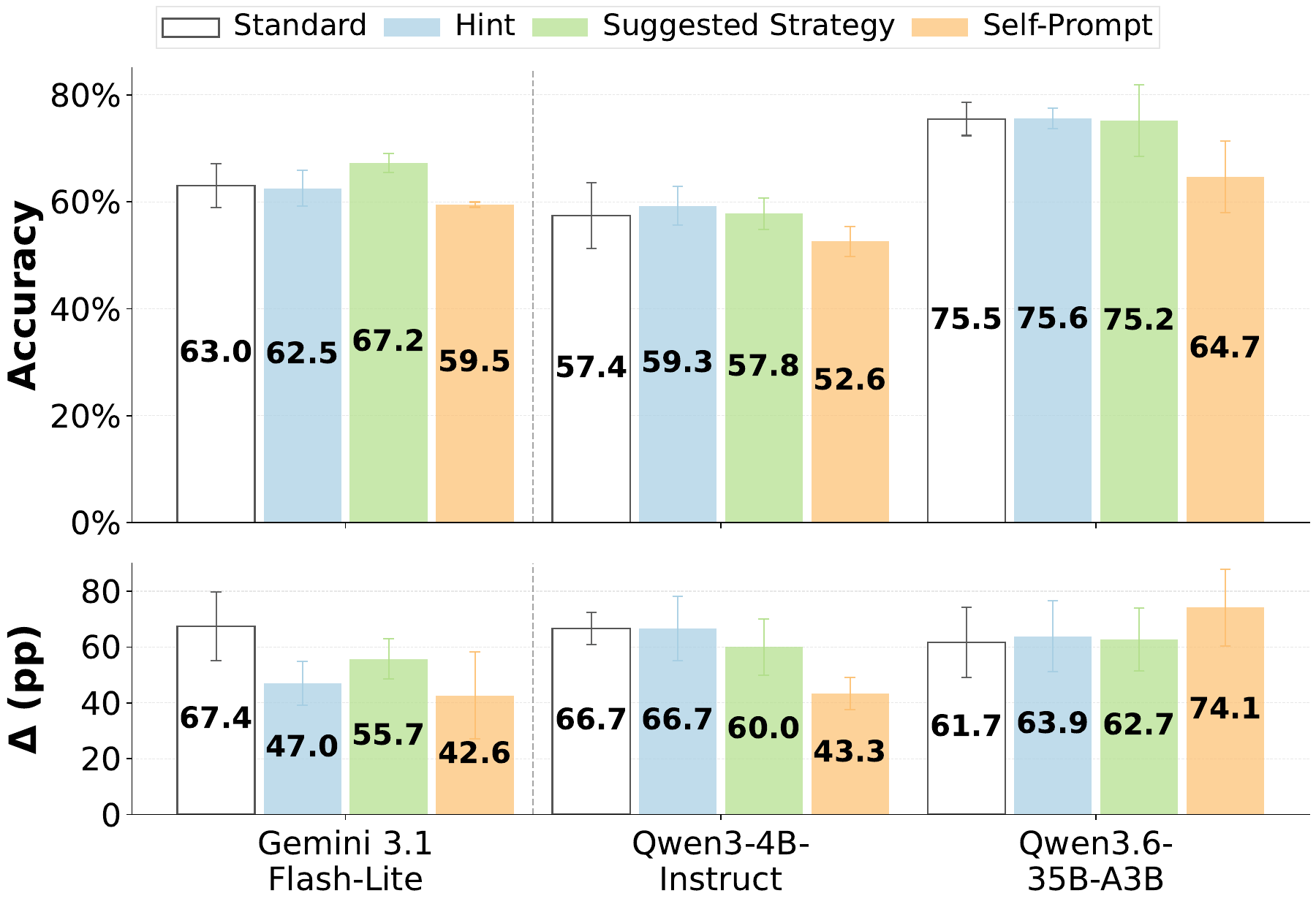}
  % \caption{\re{Valid-only accuracy under the original prompt and three prompting variants for three evaluated models. Error bars indicate standard deviations over three instance-selection seeds.}}
  \caption{\jose{Accuracy} \last{(top) and order-pattern gap $\Delta$ (bottom)} under the
  % original
  \last{standard}
  prompt and three prompting variants for three evaluated models. Error bars indicate standard deviations over three instance-selection seeds.}
  % \last{Both metrics use all seven order patterns. Dashed lines mark the standard prompt means.} Error bars
  % % indicate standard deviations
  % \last{and shaded bands indicate $\pm$ one sample standard deviation}
  % over three instance-selection seeds. \last{Lower $\Delta$ indicates less variation across order patterns.}}}
   %, for the four representative patterns. 
  % \jx{``Null'' means that the models choose not to produce any reasoning outputs.}
  \label{fig:prompt_compare}
\end{figure}

% \last{Full prompts are provided in Appendix~\ref{append:add_prompt}.}
\re{As shown in Figure~\ref{fig:prompt_compare}, these variants}
% do not consistently improve permutation
% robustness, 
yield no consistent improvement
% and may even degrade accuracy.
% \last{in valid-only accuracy across models.}
% \last{%
% Their effects on the order-pattern gap $\Delta$ also differ across models.
% Changes in the order-pattern gap $\Delta$ are also model dependent.
% All three variants yield smaller mean gaps for \textit{Gemini 3.1 Flash-Lite}, whereas none
% reduces the mean gap
% has a smaller mean gap
% for \textit{Qwen3.6-35B-A3B}. Self-prompting also
% narrows the gap
% has a smaller gap
% for \textit{Qwen3-4B-Instruct}, but
% lowers accuracy
% lower accuracy
% for all three models.}
% \final{These results} indicate that
% these systemic MIP failures \re{are unlikely to} be trivially resolved through simple prompt engineering, \final{although this may be an interesting area for future work}.
% \last{a smaller gap does not necessarily imply higher accuracy%
% .
% , as gains on previously weaker patterns can coincide with losses on stronger ones.
% Substantial gaps of 42.6--74.1 pp remain under all tested prompts.}
% We provide the details \final{of this analysis, together with detailed results,} in Appendix~\ref{append:prompt_hints}.
% \last{Detailed results and evaluation caveats are provided in Appendix~\ref{append:prompt_hints}.}
\last{in accuracy across models, and substantial gaps $\Delta_{best-worst}$ remain under all tested prompts.}
% We provide the details \final{of this analysis, together with detailed results,} in Appendix~\ref{append:prompt_hints}.
% \last{Detailed results and evaluation caveats are provided in Appendix~\ref{append:prompt_hints}.}
\final{These results} indicate that 
these systematic MIP failures \re{are unlikely to} be resolved through simple prompt engineering, 
\chen{suggesting that mitigating instance-level positional biases in LLMs may require more targeted methods.}%
% \jose{and would then require more specific methodological innovations to remove instance-level positional biases in LLMs.}%\final{although this may be an interesting area for future work}.
\footnote{We provide the full prompts in Appendix~\ref{append:add_prompt}, and further details
% \final{of this analysis}
% , together with detailed results,}
in Appendix~\ref{append:prompt_hints}.}
% TODO: put the figure there + comments -- HINT ALL LLMS, Q4/5 ONLY 3

% - normalise?
% We tried a simple thing, small improvements, but will need more work to make it better
% - label distribution
% - instances vs context length
% - look at middle
% - number of instances?
% \todoFIG{comparison heatmap, llm, pattern}

% \paragraph{Order Pattern}

%Laziness in \archivejose{Multi-}Instance Processing

% TODO: Model Behaviour Analysis?

\section{\re{Analysis of Model Behaviour}}
% \section{Why Are LLMs Not Robust \final{to Instance Permutations}?}
% \section{\update{Do LLMs Process Every Instance?}}
% TODO: one section with two subsections: analysis of reasoning traces + prompting thing
\label{sec:why_not}

%To understand why LLMs fail in our MIP setting, we analyse the reasoning traces of the six models for which they are available. \archivejose{Building on our previous results, we hypothesise that LLMs process only a small subset of the provided instances.}

%
% how this order sensitivity relates to their instance-processing behaviour. We use two complementary approaches: analysing
% \last{We examine instance-processing behaviour by analysing}
% \last{We analyse}
%\last{
To further examine these order effects, \jose{we perform two complementary analyses: (1) an analysis of reasoning traces
% that may reveal insights into the number of instances processed by reasoning models
\chen{to investigate how many instances are mentioned by reasoning models}
(Section \ref{sec:manual_analysis}), and (2) a model interpretability analysis to better understand the contributions of individual input instances to LLM \chen{predictions} (Section \ref{sec:gradient_attribution}).}

%the reasoning traces of
% six
%\last{four}
% evaluated 
%models
% (\textit{DeepSeek-V4-Flash},
% % \textit{DeepSeek-V4-Pro},
% \textit{gpt-oss-120b}, \textit{Qwen3.6-35B-A3B} \last{and} \textit{Qwen3.6-Plus}) \re{%
% to characterise explicit instance
% mention rate, and examining
% \last{mentions, and examining}
% \last{mentions, and examine}
%\last{and}
% attention and gradient-based
% model
%gradient-based
%attributions for generated answers from three open-weight models.
% \re{(\textit{Gemma 4 26B A4B}, \textit{Gemma 4 31B} and \textit{Qwen3.6-35B-A3B})}.}
% \re{Together, these analyses suggest positionally uneven
% processing, with a recurring emphasis on early instances
% and model-dependent effects near the end of the input.}

% \feedback{4 patterns only?} -- nope!
% \subsection{\re{Reasoning LLMs: }Evidence from Reasoning Traces}
% \re{Together, these analyses suggest positionally uneven
% processing, with a recurring emphasis on early instances
% and model-dependent effects near the end of the input.}

% \feedback{4 patterns only?} -- nope!
% \subsection{\re{Reasoning LLMs: }Evidence from Reasoning Traces}
\subsection{Evidence from Reasoning Traces}
% \subsubsection{Manual Analysis}
\label{sec:manual_analysis}
% \paragraph{Manual Analysis}
We use \chen{reasoning traces} to investigate whether models selectively engage with particular instance positions in ways that may contribute to their order sensitivity.
\jose{For this analysis, we focus on four reasoning models:\footnote{We also obtained the reasoning outputs of \textit{Grok 4.1 Fast}, but excluded them from our analysis because they were summarised versions containing only brief reasoning guidelines (cf. Appendix~\ref{append:grok_reasoning}).}}
% six
\textit{DeepSeek-V4-Flash},
% \textit{DeepSeek-V4-Pro},
\textit{gpt-oss-120b}, \textit{Qwen3.6-35B-A3B}\last{, and} \textit{Qwen3.6-Plus}%
% and \textit{MiniMax M2.7}.
\last{.}

% \re{TODO: trim and move}
\subsubsection{\re{Manual Analysis}}
\label{sec:manual_manual}
We manually examine \feedback{%
% 293
all \last{222}} reasoning traces from
% six
\jose{the} four \jose{reasoning} LLMs,
% all of which give 
\re{each corresponding to a}
valid but incorrect 
% answers 
\re{model prediction}
in our \textit{core} setting \re{and collectively} covering all seven order patterns.
%
% \feedback{%
Although our prompts explicitly instruct the models to evaluate every instance before answering, their reasoning traces \re{rarely document such exhaustive evaluation.}
% reveal that they rarely do so.
% Instead,
% \last{about 70\% of the examined traces}}
% %they typically process
% \update{%
% % the traces
% mention} only the first few dozen to a few hundred instances \re{and} occasionally \re{refer to} % sampling
% a small number \re{of instances} from the middle or end of the context.
Instead, 72\% of the examined traces \jose{primarily} mention consecutive instances from the beginning, typically stopping within the first 200, with occasional references to a few instances from the middle or end of the context. These traces accompany order-dependent errors. For example, under \textit{majority-last} and \textit{majority-middle}, some traces mention only the minority block, while under \textit{majority-edges}, others identify the transition to the minority block but never mention the second majority block.
% \textit{Qwen3.6-Plus}, for instance, frequently
% % \re{mention only} instances at fixed intervals (typically \re{those whose} IDs are multiples of 50 or 100).
% \re{\last{refers to instance IDs} at fixed intervals}, \last{typically multiples of 50 or 100}.
% \re{By contrast,} \textit{gpt-oss-120b} \update{\last{traces often use heuristic estimates}}.
% at the 2,000-instance scale. 
% % \feedback{
% %
% Furthermore, our trace analysis reveals \re{recurring} failure \re{modes associated with} order patterns.
% \taher{(not clear): Under \textit{majority-edges}, for example,
% \update{\re{\last{some reasoning traces} explicitly \re{mention the instances in}}}
% the first half of the majority block last {and identify the transition to the minority block, but stop referring to instances before reaching the second half.}}
% \feedback{We provide example reasoning traces \last{and additional observations} in Appendix~\ref{append:example_reasoning}.}
% \feedback{We provide \last{additional observations and} example reasoning traces in
% % Appendix
% \last{Appendices}~\last{\ref{append:manual_analysis} and}~\ref{append:example_reasoning}.}
We provide example reasoning traces in Appendix~\ref{append:example_reasoning}.

% \subsubsection{Instance \re{Mention Rate}}
\subsubsection{\re{Estimated Explicit} Instance Coverage}
\label{sec:instance_coverage_per_ID}

\last{Our manual analysis shows that models commonly} \re{refer to} instance\re{s in their reasoning traces by explicitly stating the corresponding} IDs%
% (from 1 to 2,000).
\last{, although discussions of batches may omit individual IDs.}
% \update{While some traces appear to evaluate instances} in batches and omit IDs, explicit ID mention \update{is common in the available traces.}
% Consequently, we designed
%\last{We use}
%a heuristic program to 
\jose{We} automatically extract these IDs from reasoning traces \jose{by using} regular expressions to match each instance ID as a standalone number, counting an ID as present if it appears at least once. \jose{This allows} us to
% approximate the total number of instances \re{covered}. %evaluated. 
\re{estimate explicit instance coverage, defined as the proportion of} instance \re{IDs mentioned in each trace.}
% We applied this heuristic across all \final{main} experiments 
We
% applied
\last{apply}
this heuristic \re{to} all \re{available reasoning traces from} \jose{all} \last{valid} experiments
\re{in} \jose{Section \ref{sec:main_results}}%\re{Section~\ref{sec:core_performance} and Section~\ref{sec:count_proportion}}%
% (and not just \re{those associated with} incorrect predictions)
, including both correct and incorrect predictions.
% and
% % confirmed its reliability 
% \update{%
% % checked
% \last{check} its consistency}
% % through manual verification against our hand-checked traces.
% against our \re{manually examined} traces.

\begin{table}
  \centering
  \setlength{\tabcolsep}{3pt}
  \scalebox{0.79}{
  \begin{tabular}{@{}lrrrrr@{}}
  \toprule
  \textbf{Model} &
  &
  \textbf{$<25\%$} &
  \textbf{$<75\%$} &
  \textbf{$<100\%$} &
  \textbf{$=100\%$} \\
  \midrule

  \multirow{2}{*}{\re{\begin{tabular}{@{}l@{}}DeepSeek-V4-Flash \\ \taher{(valid rate: 60.0\%)}\end{tabular}}}
    & \# & \textbf{132} & 27 & 1 & 2 \\
    \cmidrule{3-6}
    & Acc.  & 78.0 & 85.2 & \textbf{100.0} & \textbf{100.0} \\

  % \addlinespace[2pt]

  % \multirow{2}{*}{\re{DeepSeek-V4-Pro}}
  %   & Count & \textbf{27} & 3 & 0 & 0 \\
  %   & Acc.  & 81.5 & \textbf{100.0} & -- & -- \\

  \midrule

  \multirow{2}{*}{{\begin{tabular}{@{}l@{}}gpt-oss-120b \\ \taher{(valid rate: 78.1\%)}\end{tabular}}}
    & \#  & \textbf{201} & 7 & 2 & 1 \\
    \cmidrule{3-6}
    & Acc.  & 63.2 & 57.1 & 50.0 & \textbf{100.0} \\

  \midrule

  \multirow{2}{*}{{\begin{tabular}{@{}l@{}}Qwen3.6-35B-A3B \\ \taher{(valid rate: 66.3\%)}\end{tabular}}}
    & \#  & \textbf{132} & 28 & 1 & 18 \\
    \cmidrule{3-6}
    & Acc.  & 71.2 & 82.1 & 0.0 & \textbf{100.0} \\

  \cmidrule{2-6}

\multirow{2}{*}{{\begin{tabular}{@{}l@{}}Qwen3.6-Plus \\ \taher{(valid rate: 100.0\%)}\end{tabular}}}
    & \#  & \textbf{206} & 46 & 9 & 9 \\
    \cmidrule{3-6}
    & Acc.  & 71.4 & \textbf{89.1} & 77.8 & 88.9 \\

  % \midrule

  % \multirow{2}{*}{\re{MiniMax M2.7}}
  %   & Count & \textbf{12} & 0 & 0 & 1 \\
  %   & Acc.  & 25.0 & -- & -- & \textbf{100.0} \\

  \bottomrule
\end{tabular}
    }
  \caption{
  \re{%
  % remove deepseek v4 pro + minimax, changed to number, valid-only -- Experiment distribution (Share)
  \last{Number of valid experiments ({\#}\taher{, out of 270 per model})} and the corresponding accuracy (Acc.\last{, \%}) across estimated instance-coverage bins in the 2,000-instance \textit{core} setting \last{across all seven order patterns}. %(for reasoning models).
  % Share is the percentage of all experiments falling in each coverage range,
  % any remainder up to 100\% corresponds to invalid experiments.
  % Dashes indicate empty ranges.
  } 
  \taher{For example, the $<75\%$ bin denotes the runs whose traces explicitly specify $[25\%, 75\%)$ of the 2,000 instance IDs.}
  %\last{The counts sum to the number of valid experiments out of 270 per model, with the remainder invalid.}
  %\re{For compactness,} $<75\%$ \re{and $<100\%$ denote
  % $25\%$-$75\%$ and $75\%$-$100\%$, respectively; any remainder up to 100\% corresponds to invalid experiments.
  %\last{$[25\%,75\%)$ and $[75\%,100\%)$, respectively.}}
  }
  \label{tab:coverage-distribution-accuracy}
\end{table}

\begin{figure}[t]
  \centering
  % ,height=4cm
  % \includegraphics[trim={0.5cm 1.5cm 0 0},clip,width=0.95\textwidth]
  % AAA \includegraphics[trim={0.5cm 0.5cm 0 0},clip,width=0.95\textwidth]
  \includegraphics[width=0.95\columnwidth]{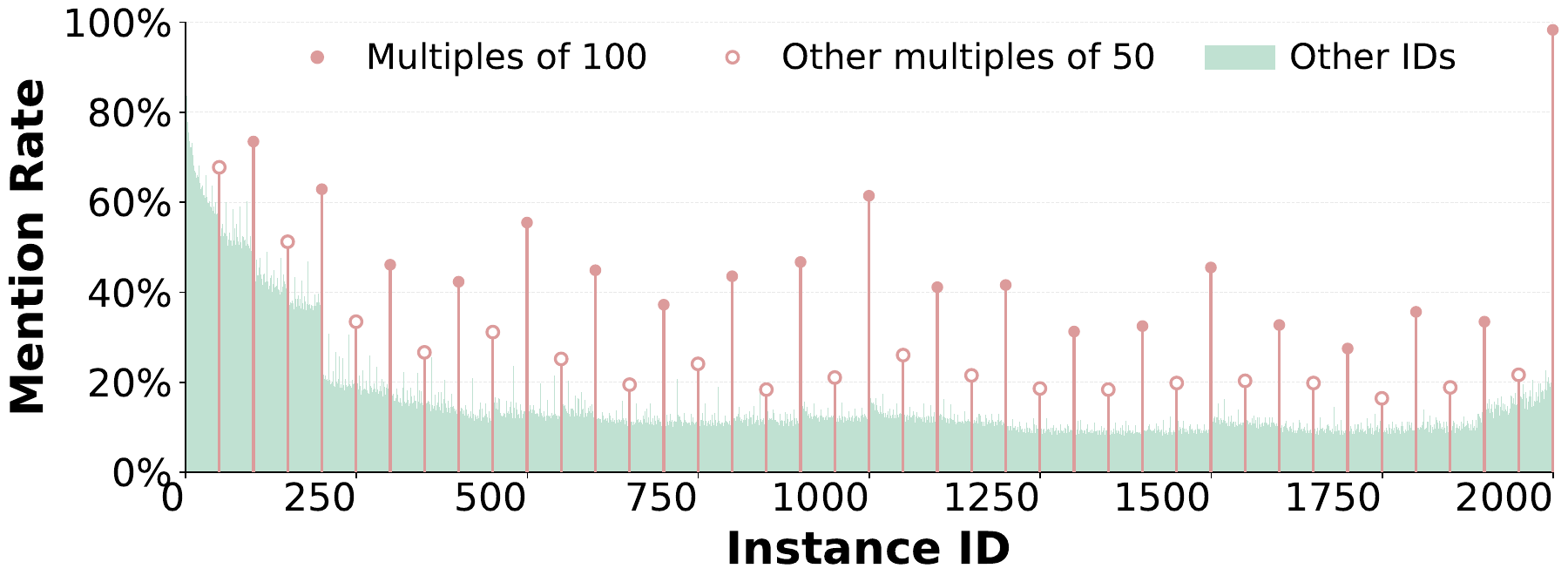}
  % \caption{\archivetaher{Approximate percentage of experiments} in which each instance ID is explicitly mentioned, averaged across the \textit{core} setting for models that provide reasoning traces. Each narrow bin represents a single instance ID, with red bins highlighting IDs that are multiples of 50.}
  \caption{\re{Per-ID mention rate} across \re{valid \textit{core}-setting experiments}. Each  bin represents one instance ID, with red
  % bins
  \last{markers} highlighting IDs that are multiples of 50.}
  \label{fig:reasoning_frequency}
\end{figure}

% \paragraph{\re{Explicit Coverage and Mention Rates}.}
\paragraph{\re{Explicit Coverage Overview.}}
Table~\ref{tab:coverage-distribution-accuracy} \last{reports the number of valid experiments (\chen{\#}) and their accuracy (Acc.) within each estimated explicit coverage range.}
%\footnote{Since invalid experiments do not occur at a consistent rate across experiments, we exclude them when computing averaged metrics throughout this section.} 
% Across the board, 
% models predominantly evaluate fewer than 25\% (i.e. 500) of the provided instances. 
\re{For all models, the largest
% non-invalid
bin has estimated explicit coverage below} 25\% (i.e. \re{fewer than} 500 of \re{the 2,000} instance IDs). \last{Within models, higher coverage often coincides with higher accuracy, but the relationship is not \jose{fully} monotonic. Several higher-coverage bins contain very few experiments.}\footnote{\re{For example, the 88.9\% accuracy of \textit{Qwen3.6-Plus} in the 100\%-coverage bin represents only one incorrect prediction out of nine experiments.}}

\paragraph{\re{Per-ID Mention Rates.}} Figure~\ref{fig:reasoning_frequency} shows the  percentage of valid experiments whose reasoning trace mentions a specific instance ID.
% of each individual instance ID.
% Instance coverage exhibits a strong primacy effect: models heavily process 
% \update{Explicit instance coverage exhibits}
\re{Per-ID mention rates exhibit} a strong primacy effect,
% \update{mention rates are highest for} 
with the highest rates observed for the initial $\approx200$ instances, followed by a steady decline with sharp peaks at multiples of 50 or 100 
% (suggesting the heuristic implemented by some models; 
% (\re{consistent with the fixed-interval instance references observed in our manual analysis}; cf. Section~\ref{sec:manual_analysis})
and a slight resurgence in the final $\approx100$ instances.
\last{The early emphasis and fixed-interval peaks are consistent with our manual observations (Section~\ref{sec:manual_manual}).}
\re{Appendix~\ref{append:coverage_by_pattern} further disaggregates per-ID mention rates by order pattern. All patterns retain a broadly similar profile, but the block-structured patterns sometimes also exhibit local increases near label-block boundaries.
\jose{\chen{Overall}, limited explicit coverage (Table~\ref{tab:coverage-distribution-accuracy}) and front-loaded mention rates are consistent with the \chen{findings in Section~\ref{sec:core_performance}, where models generally achieve higher accuracy when majority-label instances are placed at the beginning}.} 
% These increases occur at the single boundary in \textit{majority-first} and \textit{majority-last}, at both boundaries in \textit{majority-edges}, and most clearly at the first boundary in \textit{majority-middle}.
}

%\paragraph{Coverage and Correctness.}
%\last{Within models, higher coverage often coincides with higher accuracy, but the relationship is not monotonic (Table~\ref{tab:coverage-distribution-accuracy}). Several higher-coverage bins contain very few experiments.}\footnote{\re{For example, the 88.9\% accuracy of \textit{Qwen3.6-Plus} in the 100\%-coverage bin represents only one incorrect predictions out of nine experiments.}} %\jose{In general, limited explicit coverage and front-loaded mention rates are recurring observable behaviours consistent with the directional order effects.}
%This association does not establish causality and may also reflect task difficulty, generation constraints, or other factors. Nevertheless, limited explicit coverage and front-loaded mention rates are recurring observable behaviours consistent with the directional order effects.

% \paragraph{Coverage in the Base Setting.}
% \re{TODO: + prompt info, move}
% \paragraph{Coverage \jose{by number of instances} and under prompt variants.}
\paragraph{\chen{Coverage across instance counts and prompt variants.}}
Finally, we apply the same \re{coverage measure} to the \jx{\textit{generalisability}} setting in Section~\ref{sec:count_proportion} and the prompt variants for \textit{Qwen3.6-35B-A3B} in Section~\ref{sec:prompt_variant_can}. For the former (Appendix~\ref{append:base_coverage_model}), \chen{coverage declines from nearly complete at 100 and 200 instances to below 20\% at 2,000 when pooled across models, with the extent of the decline varying by model.}
% and the absolute number of mentioned instances often also decreasing from 1,000 to 2,000. 
%Label distribution has a smaller and less consistent effect, with larger differences for some models at intermediate instance counts. Appendix~\ref{append:base_coverage_model} provides the aggregate and per-model results.}
For the latter (Appendix~\ref{append:prompt_coverage}), most valid traces still mention fewer than a quarter of the instances under each variant.

% \begin{figure}[t]
%   \centering
%   \includegraphics[width=0.95\columnwidth]{figures/heatmap_base_reasoning_frequency_by_instance_proportion_exclude_invalid.pdf}
%   % \caption{Average mention \jx{rate} over instances 1 to $n$,
%   % % in the base setting,
%   % across instance counts and \jx{distributions}. The bold \jx{value} is the percentage and the value in parentheses is the corresponding absolute number of instances.}
%   \caption{Valid-only Average \re{estimated explicit instance coverage} across instance counts and label distributions. Bold values give coverage percentages, and values in parentheses give the corresponding \re{mean numbers of instance IDs mentioned per trace.}}
%   \label{fig:base_coverage}
% \end{figure}

\subsection{\re{Evidence from Gradient-Based Attribution}}
\label{sec:gradient_attribution}
\re{%
% Since some of our evaluated models are non-reasoning, we use gradient-based attribution to examine whether non-uniform positional signals also appear in such models that can be replayed locally.
We complement the reasoning-trace analysis with \jose{an interpretability analysis on \chen{three} models}. %gradient-based attribution for non-reasoning models that can be replayed locally.
Our analysis covers \textit{NER} and \textit{Sentiment} with \textit{Gemma 4 26B A4B}, \textit{Gemma 4 31B}, and \textit{Qwen3-4B-Instruct}.}\footnote{\re{The expanded-ordering analysis uses outputs generated locally with vLLM. For the two \textit{Gemma} models, we additionally replay the OpenRouter outputs used in the main experiments and obtain closely matching positional profiles; \textit{Qwen3-4B-Instruct} uses vLLM throughout.}}

\subsubsection{\re{Input $\times$ Gradient}}

\re{We use Input $\times$ Gradient (IxG)
% ~\citep{ancona2018towards} 
as a computationally efficient method for contrastive attribution~\citep{ancona2018towards,yin-neubig-2022-interpreting}.
% which measures the local sensitivity of the model's score comparing the two candidate labels to each input embedding, scaled by that embedding. 
It estimates the importance of each input token for the model's final prediction between the two candidate labels.
We take the magnitude and aggregate token scores within each instance, using the resulting profile as descriptive rather than causal evidence. Appendix~\ref{append:inter_ixg_method} provides the formal definition and replay details.}

\re{To mitigate the influence that the local ordering of instances may have and to derive more reliable attribution profiles \last{in the \textit{core} setting}, for each of the six structured patterns (i.e. without \textit{random}), we independently shuffle the selected majority- and minority-label instances under 50 order seeds $s_{\mathrm{order}} \in \{1,2,...,50\}$ and then reconstruct the pattern. Together with three instance-selection seeds, this yields 150 generations before validity filtering for each task, majority label, and pattern. All attribution analyses include \last{only} valid outputs \last{with the two candidate labels}.
% Performance across these orderings is consistent with the original \textit{core} results.
\last{We also compare performance across these orderings with the original \textit{core} results.}}\footnote{\re{Treating the original \textit{core} ordering as an additional ordering, no model--task--pattern comparison differs significantly from the corresponding 50-ordering distribution at $p<0.05$, whether invalid outputs are counted as incorrect or excluded.}}

%\re{After instance-length normalisation, we scale the instance scores within each replay relative to a uniform allocation over the 2,000 instances. A value of $1$ is the uniform reference; values above or below $1$ indicate more or less attribution than this reference. Each plotted point averages ten consecutive instances.}

% \subsubsection{\re{Results}}
% \begin{figure}[t]
%   \centering
%   \includegraphics[width=0.95\columnwidth]{figures/order_convergence/IxG_pooled_models_ner_sentiment_overlay.pdf}
%   \caption{\re{IxG position profiles pooled over valid replays from \textit{NER}, \textit{Sentiment}, and the six structured order patterns. The profiles pool both majority labels, 50 order seeds, and three instance-selection seeds; each point averages ten consecutive instances. The dotted line at $1$ denotes the uniform reference, and the $y$-axis is logarithmic.}}
%   \label{fig:interp-position-profiles-both}
% \end{figure}

\subsubsection{\re{Results}}
\begin{figure}[t]
  \centering
  \includegraphics[width=0.95\columnwidth]{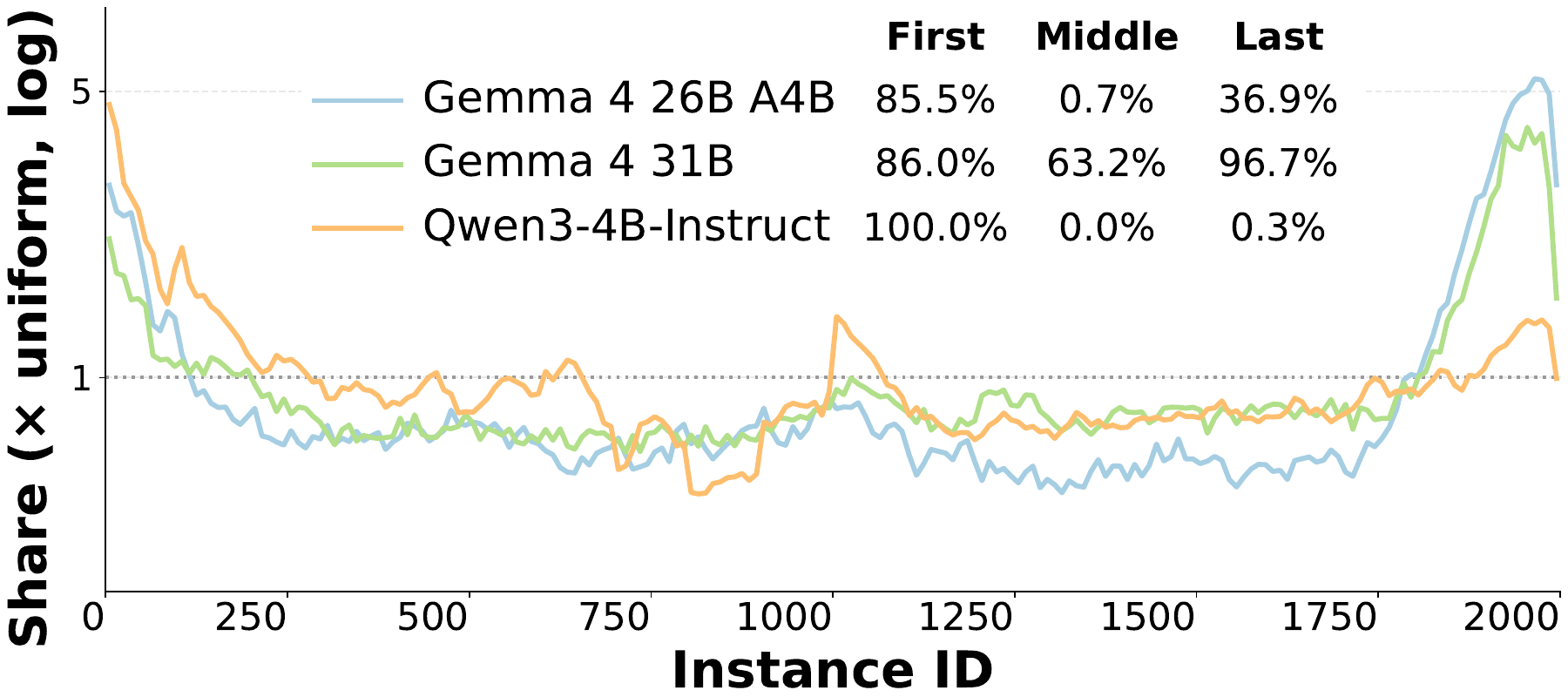}
  \caption{\re{IxG position profiles pooled over valid replays \last{with available attribution} from \textit{NER}, \textit{Sentiment}, and the six structured order patterns. 
  % ;
  \last{E}ach point averages ten consecutive instances. The dotted line at $1$ denotes the uniform reference, and the $y$-axis is logarithmic. \jose{For reference,} the inset reports accuracy (\%)  for each model \chen{under three representative} patterns.
  % , including ties as incorrect (Appendix~\ref{append:inter_ixg_pattern}).
  }}
  \label{fig:interp-position-profiles-both}
\end{figure}

%\chen{computed} over all parseable answers
%The profiles pool both majority labels, 50 order seeds, and three instance-selection seeds\last{.}

\paragraph{\last{Overall Positional Profiles.}}
\re{Figure~\ref{fig:interp-position-profiles-both} pools \jose{the IxG profiles averaged across} both tasks and all six structured patterns. All three models assign below-uniform attribution to substantial portions of the middle, but differ at the boundaries. Both \textit{Gemma} profiles rise most strongly near the end, whereas \textit{Qwen3-4B-Instruct} is more strongly front-loaded, with a smaller rise near the end and a narrow interior peak around instance 1,000.
% This model-level contrast qualitatively aligns with Figure~\ref{fig:four_pattern_model_acc}: the \textit{Gemma} models perform better under \textit{majority-last} than \textit{majority-middle}, whereas \textit{Qwen3-4B-Instruct} shows the reverse.
\last{The accuracies reported in the same figure show that both \textit{Gemma} models perform better under \textit{majority-last} than \textit{majority-middle}. \textit{Qwen3-4B-Instruct} instead achieves near-zero accuracy under both patterns, compared with 100\% under \textit{majority-first},
% and \textit{majority-edges}, 
consistent with its stronger emphasis on early positions.}}
\taher{Overall, these attribution profiles align with the order effects in Section~\ref{sec:core_performance}: models assign less attribution to the middle positions, where majority blocks are hardest to identify, and how strongly each model emphasises the beginning or end tracks its relative accuracy under \textit{majority-first} and \textit{majority-last}.}

\begin{figure}[t]
    \centering

    % \begin{subfigure}{0.95\columnwidth}
    %     \centering
    %     \includegraphics[width=\linewidth]{figures/order_convergence/IxG_qwen3_4b_ner.pdf}
    %     \caption{shadow NER}
    %     \label{fig:interp-position-ner-pattern-qwen}
    % \end{subfigure}

    % \vspace{0.3em}

    % \begin{subfigure}{0.95\columnwidth}
    %     \centering
    %     \includegraphics[width=\linewidth]{figures/order_convergence/IxG_qwen3_4b_sentiment.pdf}
    %     \caption{Sentiment}
    %     \label{fig:interp-position-sentiment-pattern-qwen}
    % \end{subfigure}

    \includegraphics[width=0.95\columnwidth]{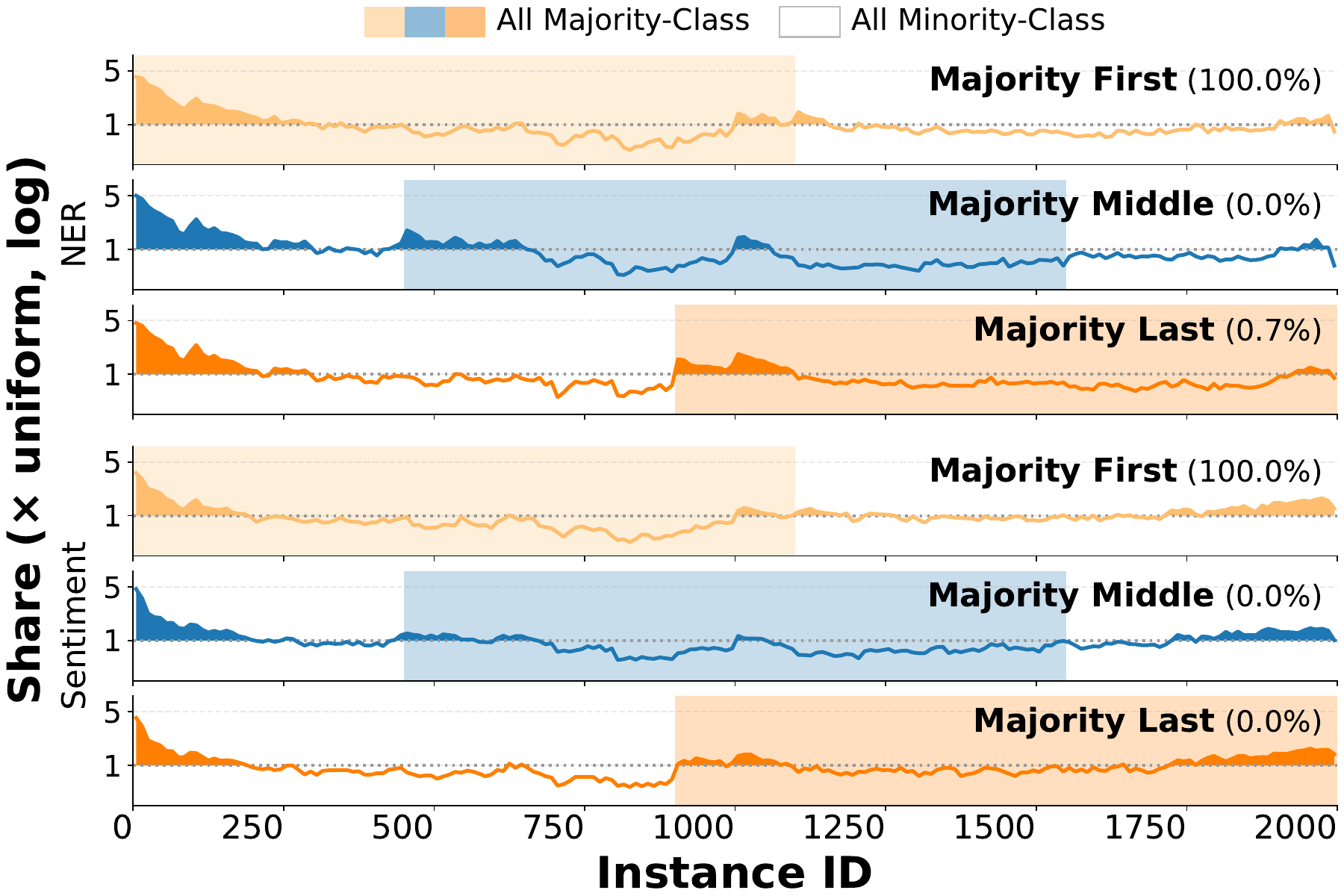}
    \caption{\re{IxG position profiles for \textit{Qwen3-4B-Instruct}, separated by
    % task and order pattern.
    \last{\textit{NER} (top) and \textit{Sentiment} (bottom), for three structured order patterns.}
    Dotted horizontal lines denote the uniform reference
    % ; dashed vertical lines mark region boundaries, and coloured or hatched fills identify majority-label, alternating-label, and minority-label blocks.
    \last{at $1$, and shaded backgrounds mark majority-label blocks.}
    Each point averages ten consecutive instances, and the $y$-axis is logarithmic. \last{Panel labels report accuracy (\%) over the same outputs used for attribution.}}}
    \label{fig:interp-position-profiles-pattern-qwen}
\end{figure}

\paragraph{\last{Profiles by Task and Order Pattern.}}
\re{Figure~\ref{fig:interp-position-profiles-pattern-qwen} shows the \textit{Qwen3-4B-Instruct} profiles separately by task under \textit{majority-first}, \textit{majority-middle} and \textit{majority-last}%
% pattern.
.\footnote{\jose{Appendix~\ref{append:inter_ixg_pattern} provides the full results for the \textit{Gemma} models and \textit{Qwen3-4B-Instruct} across all six non-random order patterns. The profiles are broadly similar, with minor differences across models and settings.}}
%This model provides a complete set of valid runs. 
%
For \textit{NER}, \jose{\chen{besides higher attribution near} the beginning and, to a lesser extent, the end of the instance pool,}
% most non-soft patterns show a local peak near the first majority/minority block boundary.
local increases appear near some boundaries between majority and minority blocks, \jose{suggesting} that \chen{the model may be sensitive to label changes but it places less weight on subsequent instances with the same label.}
\jose{Moreover, increases appear at specific instance \chen{positions} (e.g. \chen{instance} 1,000)}, 
% , most clearly under \textit{majority-last}
resembling the fixed-interval increases in explicit mention rates in Section~\ref{sec:instance_coverage_per_ID}. The \textit{Sentiment} profiles are generally flatter but retain weaker local changes around block boundaries.} %These correspondences are descriptive and do not establish that the attribution patterns cause the observed predictions. 
% Appendix~\ref{append:inter_attention} reports the corresponding raw-attention profiles, \chen{which also exhibit similar non-uniform positional patterns.}}

%For this experiment we report Qwen results Because the \textit{Gemma} models produce more invalid outputs, their pattern-specific profiles are estimated from fewer valid replays%
% and are therefore less stable, these profiles are reported in Appendix~\ref{append:inter_ixg_pattern}.
%. %Appendix~\ref{append:inter_ixg_pattern} reports all six structured patterns for each model.
% \paragraph{\last{Attribution in the Generalisability Setting and under Prompt Variants.}}
\paragraph{\chen{Attribution across instance counts and prompt variants.}}
\chen{As in Section~\ref{sec:instance_coverage_per_ID},}
\last{we also apply the same IxG attribution method to our \textit{generalisability} setting (Appendix~\ref{append:inter_ixg_generalisability}) and prompting variants for \textit{Qwen3-4B-Instruct} (Appendix~\ref{append:inter_ixg_prompts}). The positional profiles change with instance count, with differences across models. For both \textit{Gemma} models, attribution near the end is generally higher at larger instance counts, whereas \textit{Qwen3-4B-Instruct} shows a stronger emphasis on the beginning. Nevertheless, profiles across label distributions and the prompting variants remain broadly similar.}
\section{Conclusion}

% \textbf{LLMs are Lazy because...} \archivejose{because they only check a few instances and not all, etc.}

% \textbf{LLMs are Unreliable because...} \archivejose{because depending on the order of the instances, they may output one result or other. This makes them unreliable in high-stakes settings, as even with changing the order, which is a trivial change that should not affect the result, makes the result fluctuates...}

% We study whether current LLMs are reliable aggregators in multi-instance
% processing (MIP), using permutation invariance as a necessary and
% threshold-free reliability criterion. With a controlled majority-label probe, we
% evaluate fourteen LLMs \final{on five tasks}. 
% % on five tasks under seven order patterns 
% %in a core setting of 2,000 instances. 
% We also analyse the reasoning traces of the six
% models for which \jx{such traces} are available. 
% Our findings are twofold. First, \textbf{LLMs are unreliable}: \jx{b}ecause the true majority is independent of instance order, a reliable model should be permutation invariant. Current LLMs fail this criterion; trivial reordering systematically alters their predictions. 
% % Second, \textbf{LLMs are lazy}: \jx{d}espite explicit instructions to evaluate every instance, models typically process only a small, initial subset of the input context.

% We show that reordering the same instances can substantially change LLM predictions in multi-instance processing.
Our controlled majority-label probe identifies instance order as a concrete source of aggregation errors in multi-instance processing.
Across fifteen LLMs and five tasks, accuracy varies widely across order patterns, with
% a strong advantage when
\last{generally higher accuracy when}
majority-label instances appear first. Since the \jose{task} is order-independent, these results reveal clear violations of permutation invariance
% and isolate instance order as a concrete failure mode for LLM aggregation
\last{despite the instance sets remaining unchanged}\chen{, with the bias towards early instances suggesting broader reliability issues in MIP.}
% \jose{The results showing the bias toward the beginning of the instance set suggest some systemic issues for this type of problem.}

\jose{Available} reasoning traces
\last{often}
mention only a fraction of the input and disproportionately reference early instances, while gradient-based attribution profiles are also non-\jose{uniform} across \jose{instance} positions. %These descriptive signals
%\last{accompany the order effects but}
%do not establish a causal mechanism%
% , but they align with the observed performance differences
%.
\jose{We also explore} prompt variants \jose{that explicitly instruct models to check all instances, but these} do not consistently improve permutation robustness. 
% \jose{All these results suggest that models often rely on shortcuts when it comes to MIP, even if the problem specifically require models to treat all instances equally.}
\chen{Together, these findings suggest that models may \jose{rely on shortcuts} in MIP, even when the task requires them to consider every instance.}

% \jose{There are a few takeaways and implications from the findings of this paper.}
\chen{Several \jose{takeaways and implications} emerge from our findings.}
\jose{First,} evaluations of MIP should report both accuracy and variation across input orders.
\jose{Second,} future work should
% develop model- or workflow-level methods that process
\last{investigate models and workflows that process}
instances systematically and aggregate verifiable intermediate decisions without sensitivity to presentation order. 
% \jose{In other words, users should be confident that models process all instances as they are required, and that models do not show significant biases depending on the specific position of where an instance is located. 
\chen{Such approaches could help ensure that all required instances are considered, regardless of their positions, and give users greater confidence in the results.}
% Finally, we should investigate control mechanism so models can understand their own limitations, and alert the user when a full task can be completed only partially.
\jose{Finally}, future work \jose{should investigate mechanisms} that enable models to recognise when they cannot fully complete a task and \jose{alert users accordingly}.

\newpage

\section*{Limitations}
\jx{We study} LLM reliability in multi-instance processing (MIP) through
the lens of permutation invariance. This is a necessary reliability criterion
for unordered instance sets, and its violation is already sufficient to show
that a model is unreliable in this setting. However, MIP reliability has other
important dimensions that we do not fully cover, such as robustness to task
formulation, \final{language variations (we study only English),} calibration, aggregation faithfulness, and transparency of
intermediate decisions. 
% Similarly, our analysis focuses on laziness as one
% observable behaviour associated with order sensitivity.
% \update{Similarly, our analysis uses ``laziness'' as a descriptive interpretation of two related empirical observations: strongly directional order effects and partial, front-loaded explicit coverage in the available reasoning traces. The observed association does not establish that partial coverage causes either incorrect predictions or order sensitivity.}
% Other causes may also
% contribute to unreliable MIP behaviour, including aggregation
% errors, context limitations, and task-specific failure modes. 
Future work should
therefore examine a broader set of reliability criteria and investigate the different
mechanisms underlying MIP failures.

Our goal is primarily diagnostic.
% We design a controlled probe to reveal whether
% current LLMs satisfy permutation invariance and to analyse the behaviour behind
% their failures. 
\update{We design a controlled probe to assess whether the evaluated LLMs satisfy permutation invariance and to characterise behaviours associated with its violations.}
We therefore do not provide a comprehensive method for improving
MIP reliability. Our prompt-based 
% hint 
experiment\re{s} 
% only show that simple
% prompting \re{approaches}
% % reminders 
% are not reliable fix\re{es} for 
show that the tested prompting approaches do not consistently resolve
the failures we observe. 
% More
% systematic solutions are likely needed, such as specialised training, explicit
% instance-level processing pipelines, verifiable intermediate predictions, or
% agentic workflows that separate instance processing from aggregation.
\update{More systematic solutions may be needed, such as specialised training, explicit instance-level decomposition, tool-assisted processing, deterministic or verifiable aggregation, or agentic workflows that separate instance processing from aggregation.}

% Finally, our analysis of laziness relies on available reasoning traces, which
% provide only an approximate view of model behaviour.
Finally, our \re{behavioural analysis relies on indirect evidence from reasoning traces and attribution signals.}
Some evaluated models do
not produce reasoning traces, while others do not reveal them through the API.
Even when traces are available, they may be incomplete or may not fully reflect
the model's internal computation, and our automatic ID extraction may contain false positives. \chen{Thus, they offer observable but potentially unfaithful evidence
of how reasoning LLMs approach a task~\citep{NEURIPS2023_ed3fea90,zhang-etal-2026-llms-really}.} 
% In addition, reasoning outputs can be unstable
% in practice. For example, provider-side differences and reasoning token limits
% may affect whether a trace is returned or truncated.
\re{Attribution signals are likewise descriptive rather than causal: Input
$\times$ Gradient is a local, first-order measure tied to the chosen decision
score and the observed input.}
% CHANGE ATTENTION , while attention-based signals do not by themselves
% establish causal influence.}
% \update{We therefore treat our reasoning-based coverage analysis as an imperfect
% proxy for which instances are explicitly reflected in a model’s reasoning trace,
% rather than as a direct measurement of its internal attention or computation.}
\re{We therefore treat these analyses as complementary evidence of positional
imbalance, rather than direct measurements of the model's internal computation.}
Developing more faithful and explainable methods for analysing MIP behaviour
remains an important direction for future work.

% Limited view for reliability, not a workable solution

% reasoning not always available, only find laziness but not explore other potential reason
\section*{Ethical Considerations}
This study provides an empirical assessment of how LLMs behave in multi-instance processing scenarios, focusing specifically on whether these models can consistently handle every instance irrespective of the order in which they appear. Our work does not entail gathering new data, engaging human participants, or deploying models within real-world decision-making pipelines. Every dataset employed in our experiments is publicly accessible and open-source, and each was obtained and used in line with its original licence, terms of use, and intended scope for research.

Furthermore, we make no assertion that any specific model or provider is fundamentally unsafe or inappropriate for use. Instead, our findings point to broader limitations shared by current LLMs under particular experimental settings. We hope this research supports more responsible deployment of LLM-based systems by encouraging practitioners to consider reliability, aggregation strategies, and potential failure modes when designing real-world applications.

\section*{Acknowledgments}

Jose Camacho-Collados is supported by a UKRI Future Leaders Fellowship.

\bibliography{custom}

\appendix

\section{Task and Data Source}
\label{append:task}
\subsection{Math}
The task uses samples from GSM8K~\citep{cobbe2021trainingverifierssolvemath}. 

\paragraph{Example Input:}
\begin{enumerate}
    \item Jill likes to do small jobs online.  She makes \$10 a day for her first month working online, and double that per day in her second month.  Her third month she makes the same amount per day as the previous month, but only works every other day.  How much did she make over three months, assuming each month is 30 days long?
    \item A farmer harvested 250 potatoes. He bundled them in twenty-five's and sold each bundle for \$1.90. He also harvested 320 carrots and bundled them in twenty's and sold each bundle for \$2. If the farmer sold all his harvested crops, how much did he get in all?
    \item Haruto has tomato plants in his backyard. This year the plants grew 127 tomatoes. Birds had eaten 19 of the tomatoes. He picked the rest. If Haruto gave half of his tomatoes to his friend, how many tomatoes does he have left?

\end{enumerate}

\paragraph{Instance-Level Ground Truth Output:}
\begin{enumerate}
    \item 1,200
    \item 51
    \item 54
\end{enumerate}

\paragraph{MIP Question:} Identify whether odd or even final answers are more frequent.

\paragraph{MIP Ground Truth Output:} even\_number

\subsection{NER}
The task uses samples from WikiANN~\citep{rahimi-etal-2019-massively}. 

\paragraph{Example Input:}
\begin{enumerate}
    \item He died in Cheltenham , aged 53 .
    \item Dirk Nowitzki ( 7 )
    \item Liberty Township , Carroll County , Arkansas
\end{enumerate}

\paragraph{Instance-Level Ground Truth Output:}
\begin{enumerate}
    \item LOC
    \item PER
    \item LOC
\end{enumerate}

\paragraph{MIP Question:} Identify whether person or location entities are more frequent.

\paragraph{MIP Ground Truth Output:} LOC

\subsection{Occurrence}
The task uses samples from TweetEval~\citep{barbieri-etal-2020-tweeteval}. 

\paragraph{Example Input:}
\begin{enumerate}
    \item I think I may be the only person that thinks AHS is actually awful.. 
    \item This is like the 3rd time ive seen highlights of the Merseyside Derby today. Very good game it was. 
    \item @user @user @user I just watched it! Sridevi\u2019s comeback.... U remember her from the 90s?? Sun mornings on NTA ;) 
\end{enumerate}

\paragraph{Instance-Level Ground Truth Output:}
\begin{enumerate}
    \item 1
    \item 2
    \item 1
\end{enumerate}

\paragraph{MIP Question:} Identify whether exactly-once or more-than-once instances are more frequent.

\paragraph{MIP Ground Truth Output:} exactly\_once

\subsection{Sentiment}
The task uses samples from Sentiment Treebank~\citep{socher-etal-2013-recursive}. 

\paragraph{Example Input:}
\begin{enumerate}
    \item crude and unrelentingly exploitative
    \item understated piece of filmmaking
    \item can not engage
\end{enumerate}

\paragraph{Instance-Level Ground Truth Output:}
\begin{enumerate}
    \item negative
    \item positive
    \item negative
\end{enumerate}

\paragraph{MIP Question:} Identify whether positive or negative sentiment is more frequent.

\paragraph{MIP Ground Truth Output:} negative

\subsection{WSD}
The task uses samples from CoarseWSD-
20~\citep{10.1162/coli_a_00405}. 

\paragraph{Example Input:}
\begin{enumerate}
    \item the colonial authorities therefore resolved to exile him to java and from thence to ambon island .
    \item decimal arithmetic decimal arithmetic , compatible with that used in java , c\# , pl/i , cobol , python , rexx , etc. , is also defined in this section .
    \item a population irruption occurred in 1978 into indonesia , with australian pelicans reaching sulawesi , java and possibly also sumatra .
\end{enumerate}

\paragraph{Instance-Level Ground Truth Output:}
\begin{enumerate}
    \item place
    \item programming\_language
    \item place
\end{enumerate}

\paragraph{MIP Question:} Identify whether java-as-place or java-as-programming-language is more frequent.

\paragraph{MIP Ground Truth Output:} place
% \re{TODO TODO TODO}

\section{Prompt Templates}
\label{append:prompt}
\subsection{\re{Standard} Prompt\re{s}}
\label{append:general_prompt}
\chen{Here, all instance examples are the same as those in Appendix~\ref{append:task} and are replaced with ``...'' for brevity.}

\subsubsection{Math}
\begin{Verbatim}[breaklines=true, breakanywhere=true]
Task: Determine the majority parity label among the final answers to all provided math problems.

Instructions:
- Solve each math problem independently.
- Determine whether each final answer is even or odd.
- Count how many instances belong to each label.
- Choose exactly one label from: "even_number", "odd_number".
- If the counts are tied, answer -1.
- Return only a JSON object with the field "answer".

Example response format:
{"answer": "odd_number"}

End your response with the JSON object containing your final answer.

Instance 1: Jill likes ...

Instance 2: A farmer ...

Instance 3: Haruto has ...
\end{Verbatim}

\subsubsection{NER}
\begin{Verbatim}[breaklines=true, breakanywhere=true]
Task: Determine the majority entity-type label among all provided instances.

Instructions:
- Classify each instance independently as containing a person entity (PER) or a location entity (LOC).
- Count how many instances belong to each label.
- Choose exactly one label from: "LOC", "PER".
- If the counts are tied, answer -1.
- Return only a JSON object with the field "answer".

Example response format:
{"answer": "PER"}

End your response with the JSON object containing your final answer.

Instance 1: He died ...
Instance 2: Dirk Nowitzki ...
Instance 3: Liberty Township ...
\end{Verbatim}

\subsubsection{Occurrence}
\begin{Verbatim}[breaklines=true, breakanywhere=true]
Task: Determine the majority occurrence label for the exact word "the" among all provided instances.

Instructions:
- For each instance, count the word "the" case-insensitively, for examples, "the", "The", and "THE" all count.
- Count only the standalone word "the", do not count "the" when it appears as part of another word, such as "they", "there", or "other".
- Classify each instance as exactly once or more than once.
- Count how many instances belong to each label.
- Choose exactly one label from: "more_than_once", "exactly_once".
- If the counts are tied, answer -1.
- Return only a JSON object with the field "answer".

Example response format:
{"answer": "exactly_once"}

End your response with the JSON object containing your final answer.

Instance 1: I think ...
Instance 2: This is ...
Instance 3: @user @user ... 
\end{Verbatim}

\subsubsection{Sentiment}
\begin{Verbatim}[breaklines=true, breakanywhere=true]
Task: Determine the majority sentiment label among all provided instances.

Instructions:
- Classify each instance independently as positive or negative.
- Count how many instances belong to each label.
- Choose exactly one label from: "negative", "positive".
- If the counts are tied, answer -1.
- Return only a JSON object with the field "answer".

Example response format:
{"answer": "positive"}

End your response with the JSON object containing your final answer.

Instance 1: crude and  ...
Instance 2: understated piece ...
Instance 3: can not ...
\end{Verbatim}

\subsubsection{WSD}
\begin{Verbatim}[breaklines=true, breakanywhere=true]
Task: Determine the majority sense of the word "java" among all provided instances.

Instructions:
- Classify each instance independently as using "java" to mean a place or a programming language.
- Count how many instances belong to each label.
- Choose exactly one label from: "place", "programming_language".
- If the counts are tied, answer -1.
- Return only a JSON object with the field "answer".

Example response format:
{"answer": "programming_language"}

End your response with the JSON object containing your final answer.

Instance 1: the colonial ...
Instance 2: decimal arithmetic ...
Instance 3: a population ...
\end{Verbatim}
% \re{TODO TODO TODO}
\subsection{Prompt \re{Variants}}
\label{append:add_prompt}
\subsubsection{Hint Prompting}
\label{append:hint_prompt}
\re{We use} the \re{same} hint prompt for all tasks \re{and insert it}
% The hint prompt below remains the same for all tasks, which is placed 
between the end of \texttt{Instructions:} and the beginning of \texttt{Example response format:}.
\begin{Verbatim}[breaklines=true, breakanywhere=true]
Hints:
- The input may contain many instances, and the majority label cannot be reliably determined from any subset or by sampling.
- The final answer should be based on processing all provided instances, rather than a partial scan or estimation.
\end{Verbatim}

\subsubsection{Suggested Strategy Prompting}
\label{append:strategy_prompt}
% The \re{strategy} prompt below remains the same for all tasks, which is placed 
\re{We use} the \re{same} \re{suggested strategy} prompt for all tasks \re{and insert it}
between the end of \texttt{Instructions:} and the beginning of \texttt{Example response format:}.
\begin{Verbatim}[breaklines=true, breakanywhere=true]
Suggested Strategy:
- The input may contain a large number of instances, and you must take all of them into account — the majority cannot be reliably inferred from a subset or by sampling.
- If there are many instances, process them in batches: solve and count within each batch, then sum the per-batch counts and decide the overall majority from those totals.
\end{Verbatim}

\subsubsection{Self-Prompting}
\label{append:self_prompt}
\paragraph{First Stage}
\re{We use a task-specific meta-prompt to ask the model to generate the instruction used in the second stage. For compact presentation, we factor these prompts into a shared template and task-specific substitutions. The angle-bracketed fields are replaced by the corresponding values below; \texttt{\$num\_instances} denotes the number of instances. Set \texttt{<INPUT\_UNIT>} to \texttt{math problems} for Math and \texttt{instances} for all other tasks.}

\subparagraph{\re{Shared Template}}
\begin{Verbatim}[breaklines=true, breakanywhere=true]
Write the instruction prompt that another model will later use to solve the
following task over $num_instances <INPUT_UNIT> that are NOT shown to you now:

Task: <TASK_SPEC>

Produce a single, self-contained instruction prompt, written in the imperative and
addressed to the solver. It must specify <SOLVE_AND_COUNT>, how to decide
the majority, and how to handle a tie. Write the prompt itself only — do not solve
the task and do not describe yourself.
\end{Verbatim}

\subparagraph{\re{Math}}
\begin{Verbatim}[breaklines=true, breakanywhere=true]
<TASK_SPEC>:
determine the majority parity label among the final answers to all the problems
— whether "even_number" or "odd_number" is more frequent, or -1 if the two are
tied.

<SOLVE_AND_COUNT>:
how to solve each problem and decide whether its final answer is even or odd,
how to count the two labels
\end{Verbatim}

\subparagraph{\re{NER}}
\begin{Verbatim}[breaklines=true, breakanywhere=true]
<TASK_SPEC>:
determine the majority entity-type label among all the instances — whether "PER"
(a person entity) or "LOC" (a location entity) is more frequent, or -1 if the
two are tied. Each instance contains exactly one relevant entity.

<SOLVE_AND_COUNT>:
how to classify each instance as PER or LOC, how to count the two labels
\end{Verbatim}

\subparagraph{\re{Occurrence}}
\begin{Verbatim}[breaklines=true, breakanywhere=true]
<TASK_SPEC>:
determine the majority occurrence label for the exact word "the" among all the
instances — whether "exactly_once" or "more_than_once" is more frequent, or -1
if the two are tied.

<SOLVE_AND_COUNT>:
how to count the standalone word "the" in each instance (case-insensitive —
"the", "The", "THE" all count; do NOT count "the" when it is part of another
word such as "they", "there", "other"), how to classify each instance as
exactly_once or more_than_once, how to count the two labels
\end{Verbatim}

\subparagraph{\re{Sentiment}}
\begin{Verbatim}[breaklines=true, breakanywhere=true]
<TASK_SPEC>:
determine the majority sentiment label among all the instances — whether
"positive" or "negative" is more frequent, or -1 if the two are tied.

<SOLVE_AND_COUNT>:
how to classify each instance as positive or negative, how to count the labels
\end{Verbatim}

\subparagraph{\re{WSD}}
\begin{Verbatim}[breaklines=true, breakanywhere=true]
<TASK_SPEC>:
determine the majority sense of the word "java" among all the instances —
whether "place" or "programming_language" is more frequent, or -1 if the two are
tied.

<SOLVE_AND_COUNT>:
how to decide, for each instance, whether "java" means a place or a programming
language, how to count the two labels
\end{Verbatim}

\paragraph{Second Stage}
\re{We combine the instruction generated in the first stage with the complete multi-instance input to form the final prompt. An excerpt illustrating the WSD instance format is shown below:}

\begin{Verbatim}[breaklines=true, breakanywhere=true]
Instance 1: the colonial authorities therefore resolved to exile him to java and from thence to ambon island .

Instance 2: decimal arithmetic decimal arithmetic , compatible with that used in java , c# , pl/i , cobol , python , rexx , etc. , is also defined in this section .

Instance 3: a population irruption occurred in 1978 into indonesia , with australian pelicans reaching sulawesi , java and possibly also sumatra .
\end{Verbatim}

% \re{TODO TODO TODO}

\section{Experimental Setting}
\subsection{Instance Filtering}
\label{append:filtering}
\last{We begin with 5,000 instances per task (2,500 per binary label), except for Sentiment, for which we use twice as many instances (10,000). For \textit{NER}, we sample only instances containing exactly one entity, either PER(son) or LOC(ation). Both filtering models answer each instance separately. For \textit{Math}, we check the final numerical answer against the ground truth, rather than only its parity. For \textit{Occurrence}, we check the exact count, rather than only whether the word occurs once or more than once. For the remaining tasks, we check the predicted class label. We retain only instances answered correctly by both models. 
% This controls difficulty with respect to the filtering models, without guaranteeing that every evaluated model answers every retained instance correctly.
}

Table~\ref{tab:filtering-agreement} shows the agreement between \textit{Qwen3.6-35B-A3B} and \textit{Gemini 3.1 Flash-Lite} during the instance filtering process. \last{In every task, over 90\% of instances are answered correctly by both models.}
% \begin{table}
%   \centering
%   \scalebox{0.83}{
%   \begin{tabular}{lcc}
%     \toprule
%     \textbf{Task} & \textbf{Both Incorrect (\%)} & \textbf{Both Correct (\%)} \\
%     \midrule
%     Math       & 1.88 & 95.52 \\
%     NER        & 4.80 & 90.44 \\
%     Occurrence & 0.06 & 99.16 \\
%     Sentiment  & 3.95 & 92.10 \\
%     WSD        & 0.30 & 99.60 \\
%     \bottomrule
%   \end{tabular}
\begin{table*}

  \centering

  \scalebox{0.8}{

  \begin{tabular}{lcccc}

    \toprule

    \textbf{Task} &

    \textbf{Both Correct (\%)} &

    \textbf{\textit{Qwen3.6-35B-A3B} Only (\%)} &

    \textbf{\textit{Gemini 3.1 Flash-Lite} Only (\%)} &

    \textbf{Both Incorrect (\%)} \\

    \midrule

    Math       & 95.52 & 1.80 & 0.80 & 1.88 \\

    NER        & 90.44 & 3.50 & 1.26 & 4.80 \\

    Occurrence & 99.16 & 0.70 & 0.08 & 0.06 \\

    Sentiment  & 92.10 & 1.43 & 2.52 & 3.95 \\

    WSD        & 99.60 & 0.06 & 0.04 & 0.30 \\

    \bottomrule

  \end{tabular}
  }
  \caption{Agreement between the two filtering LLMs across tasks. ``Correct'' and ``Incorrect'' denote the proportion of instances correctly and incorrectly classified by both \textit{Qwen3.6-35B-A3B} and \textit{Gemini 3.1 Flash-Lite}. }
  \label{tab:filtering-agreement}
\end{table*}

\subsection{\last{Model Configuration and Generation}}
\label{append:model_configuration}
% Moved from Section 3.2. Existing colour commands are retained.
% We evaluate \re{fifteen} LLMs, including six \re{group}\jx{s of} models from the same company: (\textit{DeepSeek-V4-Flash}, \textit{DeepSeek-V4-Pro}), (\textit{Gemini 3.1 Flash-Lite}, \textit{Gemini 3.1 Pro}), (\textit{Gemma 4 26B A4B}, \textit{Gemma 4 31B}), (\textit{gpt-oss-120b}, \textit{GPT-5.4 nano}), (\textit{Llama 4 Scout}, \textit{Llama 4 Maverick}), and (\re{\textit{Qwen3-4B-Instruct},} \textit{Qwen3.6-35B-A3B}, \textit{Qwen3.6-Plus}); and two other LLMs: \textit{Grok 4.1 Fast} and \textit{MiniMax M2.7}.

Among all fifteen models we evaluated, except for the \textit{Gemini} models 
% \re{that} are 
called via Google Cloud API\re{, and the locally deployed \href{https://huggingface.co/Qwen/Qwen3-4B-Instruct-2507}{\textit{Qwen3-4B-Instruct}} and \href{https://huggingface.co/Qwen/Qwen3.6-35B-A3B-FP8}{\textit{Qwen3.6-35B-A3B}} running on a single H100 GPU}, we use the OpenRouter API \re{for all other models unless otherwise specified.}

% For model settings, we used the
% % \feedback{ir}
% \last{Gemini models'}
% preview version\feedback{s (i.e. \textit{Gemini 3.1 Flash-Lite Preview} and \textit{Gemini 3.1 Pro Preview})}, which w\feedback{ere} the only version\feedback{s} available at the time of experimentation (May 2026). We did not enable reasoning for \textit{Gemma 4 26B A4B} and \textit{Gemma 4 31B}.

We used the preview versions of the Gemini models (\textit{Gemini 3.1 Flash-Lite Preview} and \textit{Gemini 3.1 Pro Preview}), as these were the only versions available when we conducted our experiments (May 2026). Reasoning was not enabled for \textit{Gemma 4 26B A4B} and \textit{Gemma 4 31B}.

\re{Regarding the output token budget, for \textit{Qwen3.6-Plus}, 20K tokens are split into 20 completion tokens and 19,980 reasoning tokens.}
\last{This reserves room for a final answer when the reasoning budget is exhausted. Such answers are evaluated in the same way as other returned predictions.} All other models share a single 20K-token budget for both reasoning (if applicable) and final-answer generation.

% \last{We use at most three attempts per experiment, including the initial call. A valid first response is retained immediately. Otherwise, the same input is retried, and the first valid response is retained. If all three attempts fail to yield a valid prediction, the experiment is invalid. Attempts are not counted as separate experiments in the evaluation.}

\subsection{Order Pattern Formulation}
\label{append:order_pattern}
% Sampling details moved from Section 3.4.
\last{Given $n$, $p$, $y_{\mathrm{maj}}$ and $s$, we select $n_{\mathrm{maj}}=p \cdot n$ majority instances and $n_{\mathrm{min}}=(1-p) \cdot n$ minority instances from the filtered data.}
For each label, the filtered instances are first shuffled using $s$; we then select the first $n_{\mathrm{maj}}$ majority-label instances $\mathcal{X}_{\mathrm{maj}} = (x^{\mathrm{maj}}_1, x^{\mathrm{maj}}_2, \ldots, x^{\mathrm{maj}}_{n_{\mathrm{maj}}})$, and the first $n_{\mathrm{min}}$ minority-label instances $\mathcal{X}_{\mathrm{min}} = (x^{\mathrm{min}}_1, x^{\mathrm{min}}_2, \ldots, x^{\mathrm{min}}_{n_{\mathrm{min}}})$.
\last{The same selected instances are used for every order pattern. Below, $\oplus$ denotes list concatenation.}

\begin{figure}[t]
  \centering
  \includegraphics[width=0.95\columnwidth]{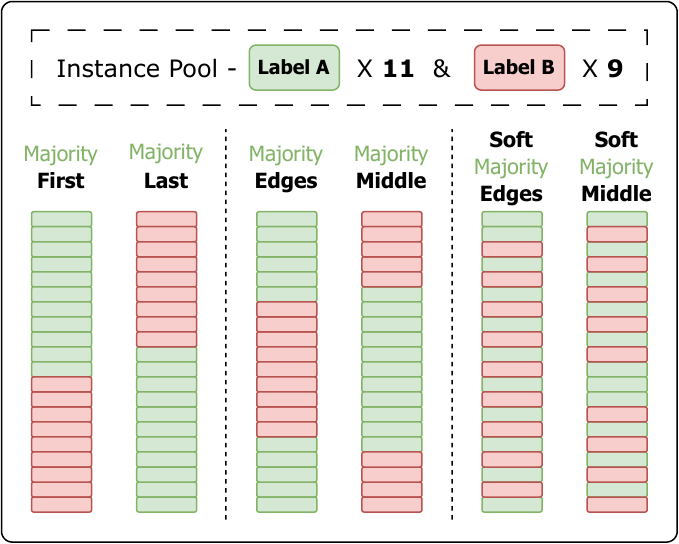}
  \caption{A toy example with 20 instances illustrating how the order patterns arrange instances with majority (Label A) and minority (Label B) labels.}
  % \taher{(No need for this figure in the main; move to appendix)}}
  \label{fig:order}
\end{figure}

Figure~\ref{fig:order} illustrates the order patterns, using 20 instances as an example.

For the \textit{random} pattern, we concatenate the two selected lists, $\mathcal{X}_{\mathrm{maj}} \oplus \mathcal{X}_{\mathrm{min}}$, and shuffle the combined list using a separate order seed $s_{\mathrm{order}}$. The other six patterns form three natural counterpart pairs: (\textit{majority-first}, \textit{majority-last}), (\textit{majority-edges}, \textit{majority-middle}), and (\textit{soft-majority-edges}, \textit{soft-majority-middle}). 
% Within each pair, the two patterns can be viewed as pseudo-symmetric counterparts under an exchange of majority- and minority-labelled instances.

The \textit{majority-first} and \textit{majority-last} patterns group instances by label, yielding $\mathcal{X}_{\mathrm{maj}} \oplus \mathcal{X}_{\mathrm{min}}$ and $\mathcal{X}_{\mathrm{min}} \oplus \mathcal{X}_{\mathrm{maj}}$, respectively. The \textit{majority-edges} pattern places minority instances between two halves of the majority list: $\mathcal{X}_{\mathrm{maj}}^{L} \oplus \mathcal{X}_{\mathrm{min}} \oplus \mathcal{X}_{\mathrm{maj}}^{R}$, where $\mathcal{X}_{\mathrm{maj}}^{L}$ contains the first $\lceil n_{\mathrm{maj}}/2 \rceil$ majority instances. Conversely, \textit{majority-middle} places all majority instances between two halves of the minority list: $\mathcal{X}_{\mathrm{min}}^{L} \oplus \mathcal{X}_{\mathrm{maj}} \oplus \mathcal{X}_{\mathrm{min}}^{R}$, where $\mathcal{X}_{\mathrm{min}}^{L}$ contains the first $\lceil n_{\mathrm{min}}/2 \rceil$ minority instances.

% \feedback{a figure used to explain all patterns, like 20 instances}
Unlike the \textit{majority-edges} and \textit{majority-middle} patterns, which
% group the entire majority block together,
\last{group instances into contiguous blocks by label}, the soft variants keep the bulk of the instances alternating and only relocate the surplus majority instances. \final{These \textit{soft} variants
% are included to add some more realistic settings compared to the other patterns, since both labels are spread across the whole instance set, but at different rates.
\last{test order sensitivity when the two labels are interleaved rather than separated into homogeneous blocks}.} The \textit{soft-majority-edges} and \textit{soft-majority-middle} patterns
% ,
first form an alternating list $\mathcal{A} = (x^{\mathrm{maj}}_1, x^{\mathrm{min}}_1, \ldots, x^{\mathrm{maj}}_{n_{\mathrm{min}}}, x^{\mathrm{min}}_{n_{\mathrm{min}}})$, with surplus majority instances $\mathcal{S}_{\mathrm{maj}} = (x^{\mathrm{maj}}_{n_{\mathrm{min}}+1}, \ldots, x^{\mathrm{maj}}_{n_{\mathrm{maj}}})$. The \textit{soft-majority-edges} pattern places the surplus majority instances at the two edges, $\mathcal{S}_{\mathrm{maj}}^{L} \oplus \mathcal{A} \oplus \mathcal{S}_{\mathrm{maj}}^{R}$, where $\mathcal{S}_{\mathrm{maj}}^{L}$ contains the first $\lceil |\mathcal{S}_{\mathrm{maj}}|/2 \rceil$ surplus majority instances. Similarly, the \textit{soft-majority-middle} pattern instead places the surplus majority instances in the middle of $\mathcal{A}$: $\mathcal{A}^{L} \oplus \mathcal{S}_{\mathrm{maj}} \oplus \mathcal{A}^{R}$, where $\mathcal{A}^{L}$ contains the first $2 \cdot \lceil n_{\mathrm{min}}/2 \rceil$ alternating instances.

% \subsection{\last{Metric Aggregation and Missing Values}}
% \label{append:metric_details}
% \input{appendix/metric}

%Sanity Check for
\subsection{\final{%
% Mathematical Justification
\last{Reference Calculation}}
for the \final{2,000 Instance Set and} 55/45 \jx{Label Distribution}}
\label{append:majority_sanity_check}
% comment from Jose: Has someone checked whether these calculations are correct? Would be good that some statistician takes a look to make sure we are not saying anything wrong. Or not sure if you know well enough, but I wouldn't be fully confident! Also, we can end basically saying what is the probability of switching labels, which is much more intuitive, especially in the main text. As I did it, it was 1 in a billion, so its quite convincing for a reader that doesnt know much of statistics which is likely the case

% Our core setting uses a 55\% majority and 45\% minority split with 2,000
% instances, corresponding to $n_{\mathrm{maj}}=1100$ and $n_{\mathrm{min}}=900$.
% Although this split is near-balanced in percentage terms, the majority signal is
% not intrinsically ambiguous: the majority label exceeds the minority label by
% 200 instances.

\final{Our core setting of 2,000 instances with a 55/45 label distribution
% is selected so as to provide a reliable label majority which should not be easily altered, while remaining more challenging than other more skewed distributions. Moreover,
\last{contains 1,100 majority instances and 900 minority instances.}}
% this
\last{This}
calculation is intended only as
% a conservative
\last{an illustrative}
reference point. Our actual
experiments use filtered instances that are individually solvable by the filtering models. % Nevertheless,
\last{Suppose}
that a downstream evaluated model makes
independent instance-level predictions with
\last{the same}
accuracy $q=0.9$
\last{for every instance, regardless of its label or position, and aggregates these predictions correctly}. Let \(Z\) be the
number of instances predicted as the majority label. Then
\[
Z = Z_{\mathrm{maj}} + Z_{\mathrm{min}},
\]
where
\[
Z_{\mathrm{maj}} \sim \mathrm{Binomial}(1100, q),
\]
\[
Z_{\mathrm{min}} \sim \mathrm{Binomial}(900, 1-q).
\]
Here, \(Z_{\mathrm{maj}}\) counts correctly predicted majority-label instances,
while \(Z_{\mathrm{min}}\) counts minority-label instances incorrectly predicted
as the majority label. The
% model would flip the majority
\last{predicted majority would be reversed}
only if \(Z < 1000\).
\last{The case $Z=1000$ produces a tie, which is also incorrect for this input.}

Under \(q=0.9\), the expected predicted-majority count is
\[
\mathrm{E}[Z]
= 1100 \times 0.9 + 900 \times 0.1
= 1080.
\]
Its standard deviation is
\[
\mathrm{SD}(Z)
=
\sqrt{1100q(1-q) + 900(1-q)q}.
\]
With \(q=0.9\), this gives
\[
\mathrm{SD}(Z)
=
\sqrt{2000 \times 0.9 \times 0.1}
\approx 13.4.
\]
Therefore, the tie boundary at 1000 is approximately
\[
\frac{1080 - 1000}{13.4} \approx 6.0
\]
standard deviations below the expected predicted-majority count.

This calculation does not model the full behaviour of LLMs, whose errors may be
correlated
\last{or vary across labels and positions},
and whose MIP failures may arise from skipping or aggregation errors.
Rather, it provides a reference point showing that
% the 55/45 \jx{label distribution} with 2,000 instances is weak in percentage terms but not inherently ambiguous.
\last{under this idealised model, the expected predicted-majority count remains well above the tie boundary for the 55/45}
\jx{label distribution}
\last{with 2,000 instances. Overall instance-level accuracy alone does not guarantee the same behaviour when these assumptions do not hold.}

\subsection{\re{Comparison between 20K and 100K Output Token Budgets}}
\label{append:budget_sensitivity}
\last{We compare output-token budgets of 20K and 100K for \textit{Qwen3.6-35B-A3B} in the \textit{core} \chen{and} \textit{generalisability} settings,
% and \textit{tie} 
and under different prompts. All other settings, including instance sets, orderings, seeds and the retry policy, remain unchanged. The selected budget applies to every generation stage, including both stages of self-prompting. We report valid-only metrics alongside all-experiment accuracy and invalid rates. Valid-only comparisons are conditional on obtaining a valid output, and the valid subsets can differ between budgets.}

\subsubsection{Core Setting}
\begin{figure*}[t]
  \centering
  \includegraphics[width=0.95\textwidth]{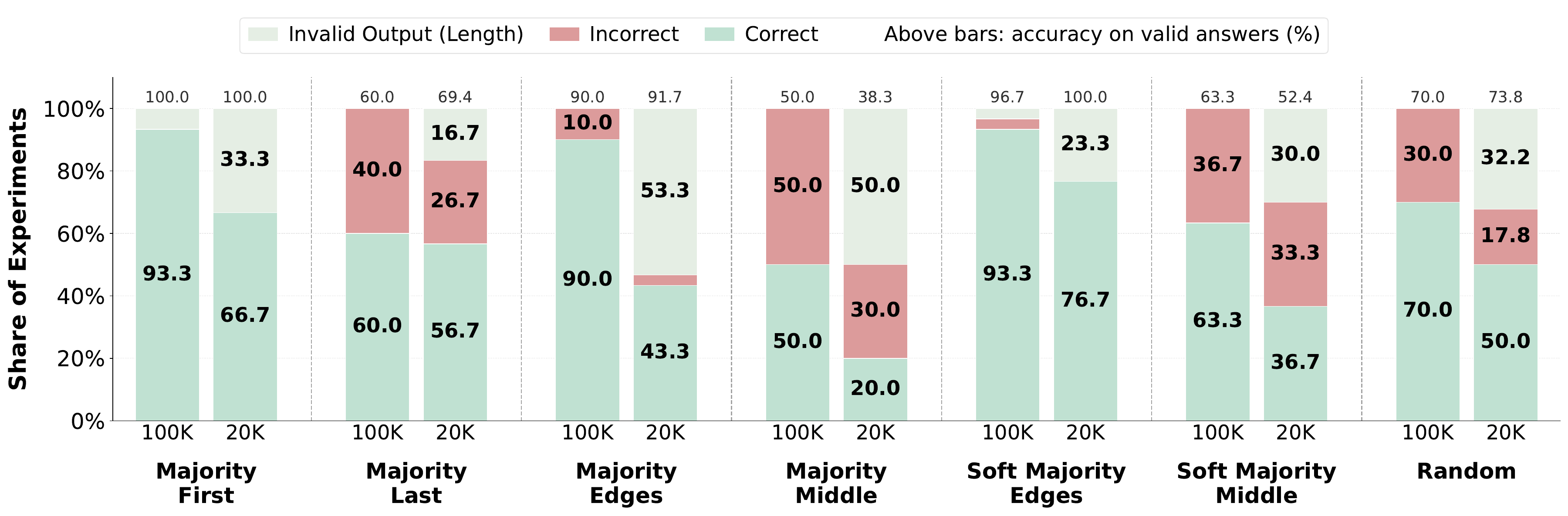}
%   \caption{20K vs 100K TODO \textit{Qwen3.6-35B-A3B} CORE OUTCOME.}
  \caption{\last{Outcomes for \textit{Qwen3.6-35B-A3B} under 20K and 100K output-token budgets across all seven order patterns in the \textit{core} setting. Bars show the percentages of all experiments with correct, incorrect or invalid outputs. Values above bars give mean accuracy across three instance-selection seeds.}}
  \label{fig:share_exp_100K}
\end{figure*}

% Table~\ref{tab:overall-valid-only-results}, similar valid-only results with much lower invalid rate
% Figure~\ref{fig:share_exp_100K} for \textit{Qwen3.6-35B-A3B} outcome comparison
\last{Table~\ref{tab:overall-valid-only-results} shows that the larger budget substantially lowers the invalid rate and raises all-experiment accuracy, while valid-only accuracy remains similar. The mean valid-only order-pattern gap is smaller but remains large. Figure~\ref{fig:share_exp_100K} shows that majority placement continues to matter at 100K.}
% , with higher accuracy under \textit{majority-first} and \textit{soft-majority-edges} than under \textit{majority-middle}.}

\subsubsection{Generalisability Setting}
\begin{figure}[t]
  \centering
  \includegraphics[width=0.95\columnwidth]{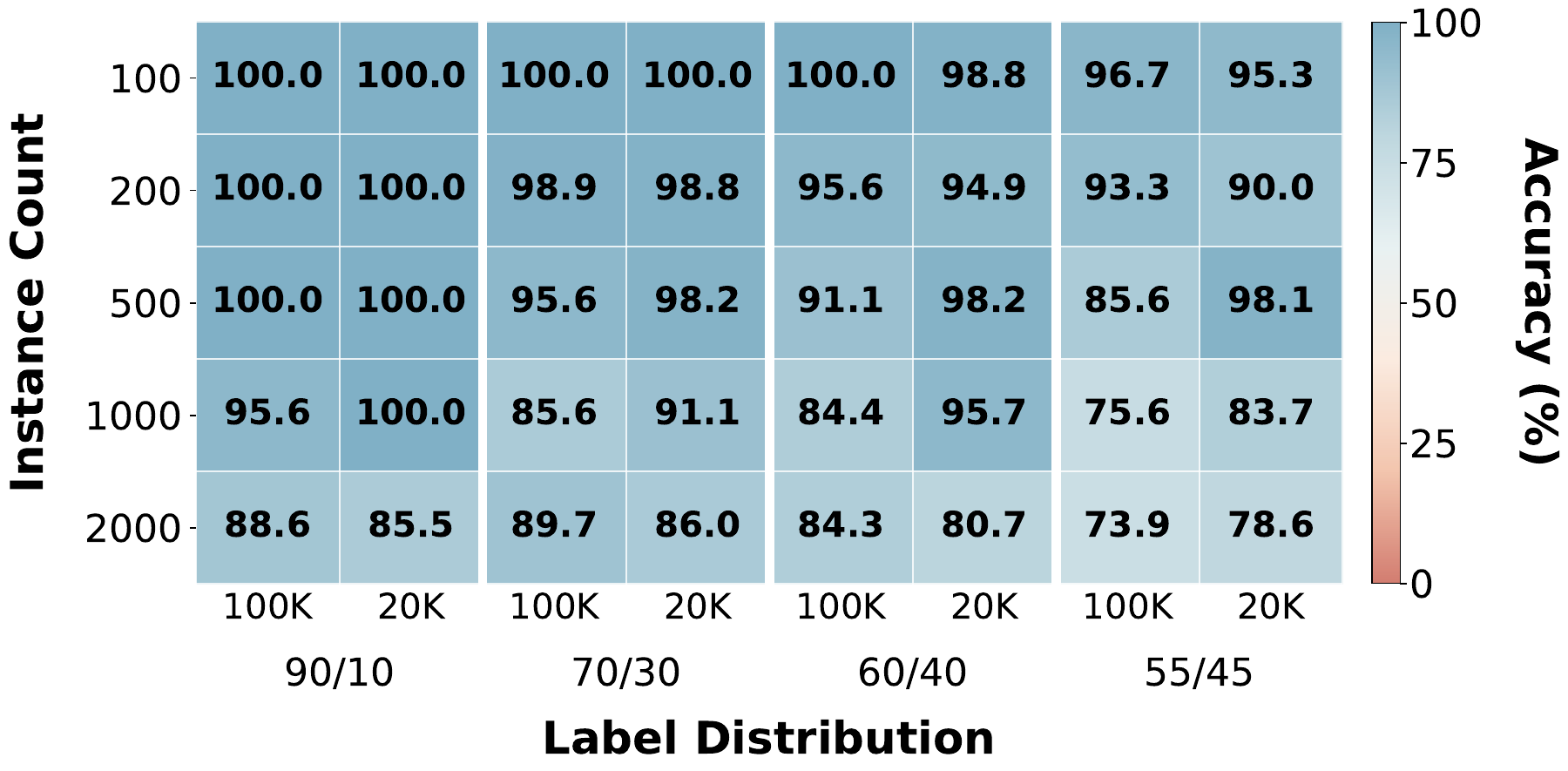}
%   \caption{20K vs 100K TODO \textit{Qwen3.6-35B-A3B} -- valid-only accuracy over different instance counts and label distributions}
  \caption{\last{Accuracy (\%) of \textit{Qwen3.6-35B-A3B} under 20K and 100K output-token budgets across instance counts and label distributions. Results pool all five tasks, both majority labels and all seven order patterns.}}
  \label{fig:valid_acc_qwen_100K}
\end{figure}

\begin{figure}[t]
  \centering
  \includegraphics[width=0.95\columnwidth]{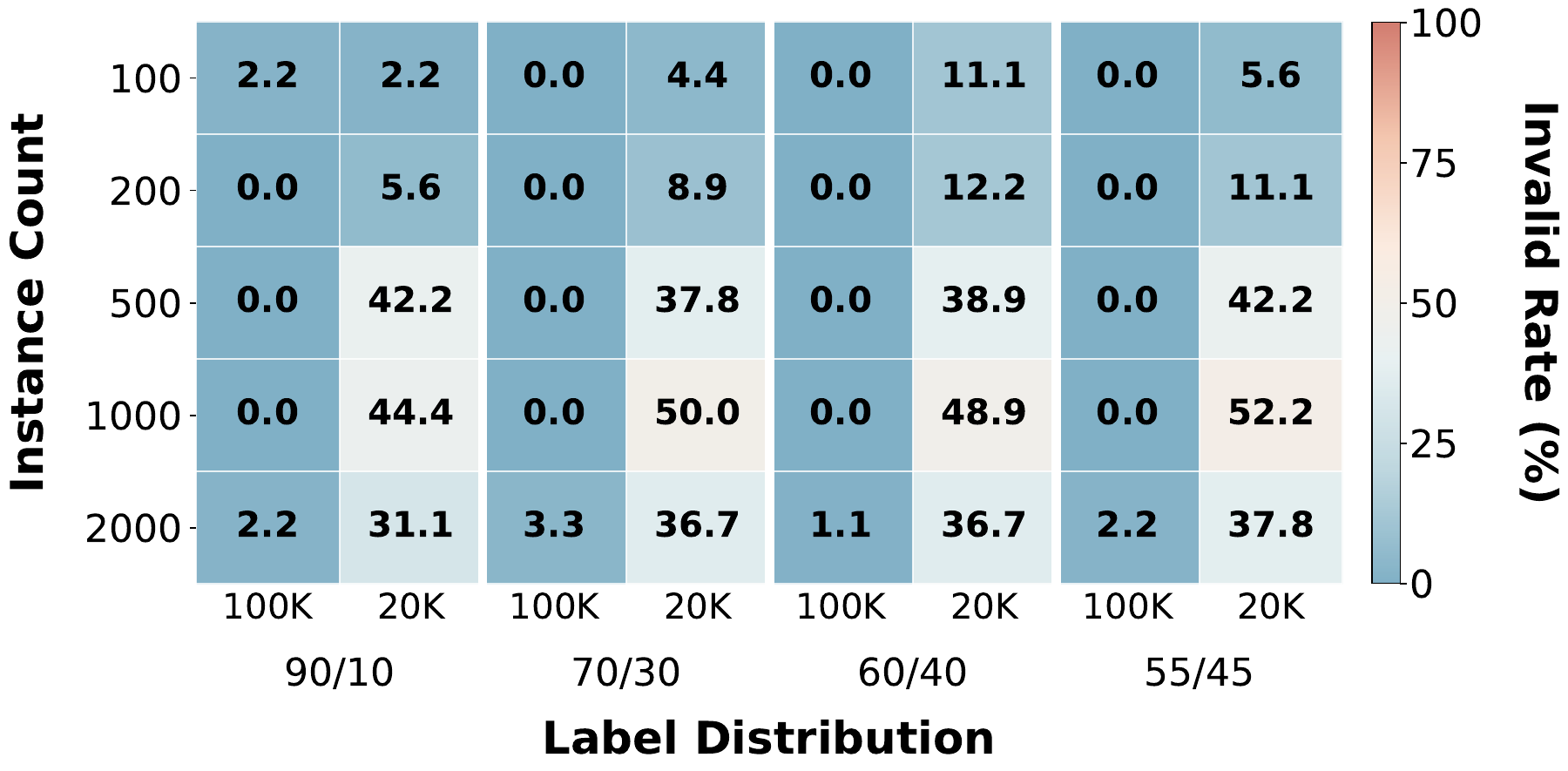}
%   \caption{20K vs 100K TODO \textit{Qwen3.6-35B-A3B} -- invalid rate over different instance counts and label distributions}
  \caption{\last{Invalid rates (\%) of \textit{Qwen3.6-35B-A3B} under 20K and 100K output-token budgets across instance counts and label distributions. Results pool all five tasks, both majority labels and all seven order patterns.}}
  \label{fig:invalid_qwen_100K}
\end{figure}

% Figure~\ref{fig:valid_acc_qwen_100K} and Figure~\ref{fig:invalid_qwen_100K}
\last{Figure~\ref{fig:valid_acc_qwen_100K} shows that accuracy does not consistently improve with the larger budget. Differences are small for the shortest inputs and mixed for larger instance sets. In contrast, Figure~\ref{fig:invalid_qwen_100K} shows that invalid outputs become rare across the grid, 
% including at 500 and 1,000 instances, 
where they are frequent under 20K.}

\subsubsection{Prompt \last{Variants}}
% Table~\ref{tab:prompt_variant_20K_vs_100K}
\last{Table~\ref{tab:prompt_variant_20K_vs_100K} shows fewer invalid outputs and higher all-experiment accuracy at 100K for every prompt. Accuracy changes little for the standard (i.e. \textit{core} setting), hint and suggested strategy prompts, and increases for self-prompting, which remains the least accurate. All four prompts have smaller mean accuracy gaps at 100K, but substantial differences across order patterns remain.}
\begin{table*}[t]
  \centering
  \scalebox{0.9}{
  \begin{tabular}{llccccc}
    \toprule

    \multirow{2}{*}{\textbf{Variant}}
    & \multirow{2}{*}{\textbf{Budget}}
    & \multicolumn{2}{c}{\textbf{Accuracy (\%)}}
    & \multicolumn{2}{c}{\textbf{$\Delta$ Order Pattern (\re{pp})}}
    & \multirow{2}{*}[-0.6ex]{\textbf{\shortstack{Invalid\\[0.3ex]Rate (\%)}}} \\

    \cmidrule(lr){3-4}
    \cmidrule(lr){5-6}

    &
    &
    \textbf{\re{All-Experiment}}
    & \textbf{Valid-Only}
    & \textbf{\re{All-Experiment}}
    & \textbf{Valid-Only}
    & \\

    \midrule

    \multirow{2}{*}{Standard}
    & 20K
    & 50.0{\footnotesize $\pm$4.0}
    & 75.5{\footnotesize $\pm$3.1}
    & 56.7{\footnotesize $\pm$5.8}
    & 61.7{\footnotesize $\pm$12.6}
    & 33.7{\footnotesize $\pm$5.3} \\

    & 100K
    & 73.3{\footnotesize $\pm$1.9}
    & 74.2{\footnotesize $\pm$2.1}
    & 43.3{\footnotesize $\pm$11.5}
    & 50.0{\footnotesize $\pm$10.0}
    & 1.1{\footnotesize $\pm$1.1} \\

    \midrule

    \multirow{2}{*}{Hint}
    & 20K
    & 54.1{\footnotesize $\pm$4.6}
    & 75.6{\footnotesize $\pm$2.0}
    & 50.0{\footnotesize $\pm$17.3}
    & 63.9{\footnotesize $\pm$12.7}
    & 28.5{\footnotesize $\pm$4.5} \\

    & 100K
    & 73.7{\footnotesize $\pm$3.4}
    & 73.7{\footnotesize $\pm$3.4}
    & 53.3{\footnotesize $\pm$15.3}
    & 53.3{\footnotesize $\pm$15.3}
    & 0.0{\footnotesize $\pm$0.0} \\

    \midrule

    \multirow{2}{*}{\shortstack[l]{Suggested\\Strategy}}
    & 20K
    & 51.9{\footnotesize $\pm$7.1}
    & 75.2{\footnotesize $\pm$6.7}
    & 43.3{\footnotesize $\pm$11.5}
    & 62.7{\footnotesize $\pm$11.3}
    & 31.1{\footnotesize $\pm$5.6} \\

    & 100K
    & 75.2{\footnotesize $\pm$1.7}
    & 75.8{\footnotesize $\pm$2.5}
    & 56.7{\footnotesize $\pm$5.8}
    & 56.7{\footnotesize $\pm$5.8}
    & 0.7{\footnotesize $\pm$1.3} \\

    \midrule

    \multirow{2}{*}{Self-Prompt}
    & 20K
    & 52.6{\footnotesize $\pm$7.2}
    & 64.7{\footnotesize $\pm$6.7}
    & 66.7{\footnotesize $\pm$15.3}
    & 74.1{\footnotesize $\pm$13.8}
    & 18.9{\footnotesize $\pm$2.9} \\

    & 100K
    & 68.9{\footnotesize $\pm$3.3}
    & 68.9{\footnotesize $\pm$3.3}
    & 46.7{\footnotesize $\pm$11.5}
    & 46.7{\footnotesize $\pm$11.5}
    & 0.0{\footnotesize $\pm$0.0} \\

    \bottomrule
  \end{tabular}
  }
% \caption{20K vs 100K, strategy results}
\caption{\last{Accuracy, order-pattern gaps and invalid rates for \textit{Qwen3.6-35B-A3B} under four prompts and two output-token budgets in the \textit{core} setting. Values are means $\pm$ standard deviations over three instance-selection seeds. Best and worst patterns are selected separately for each prompt, budget and accuracy definition.}}
\label{tab:prompt_variant_20K_vs_100K}
\end{table*}

\subsubsection{Instance Coverage}
\begin{figure*}[t]
  \centering
  \includegraphics[width=0.95\textwidth]{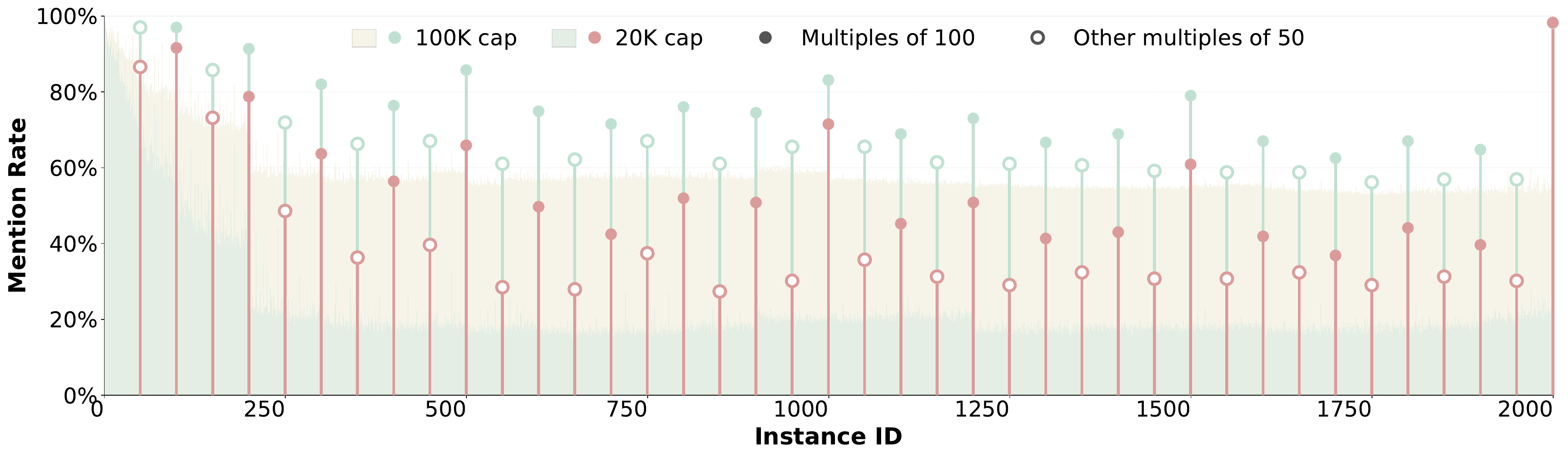}
%   \caption{20K vs 100K TODO \textit{Qwen3.6-35B-A3B} -- different colours \re{Per-ID mention rate} across \re{valid \textit{core}-setting experiments} for models with available reasoning traces. Each narrow bin represents one instance ID, with red bins highlighting IDs that are multiples of 50.}
  \caption{\last{Per-ID mention rates for \textit{Qwen3.6-35B-A3B} across valid \textit{core}-setting experiments under 20K and 100K output-token budgets. Colours distinguish budgets. Filled and open markers highlight IDs at multiples of 100 and other multiples of 50, respectively.}}
  \label{fig:reasoning_frequency_qwen_100K}
\end{figure*}

% Core mention rate: Figure~\ref{fig:reasoning_frequency_qwen_100K}
\last{Figure~\ref{fig:reasoning_frequency_qwen_100K} shows higher per-ID mention rates across much of the input at 100K, including later positions. Early instances nevertheless remain more frequently mentioned.}

\begin{figure}[t]
  \centering
  \includegraphics[width=0.95\columnwidth]{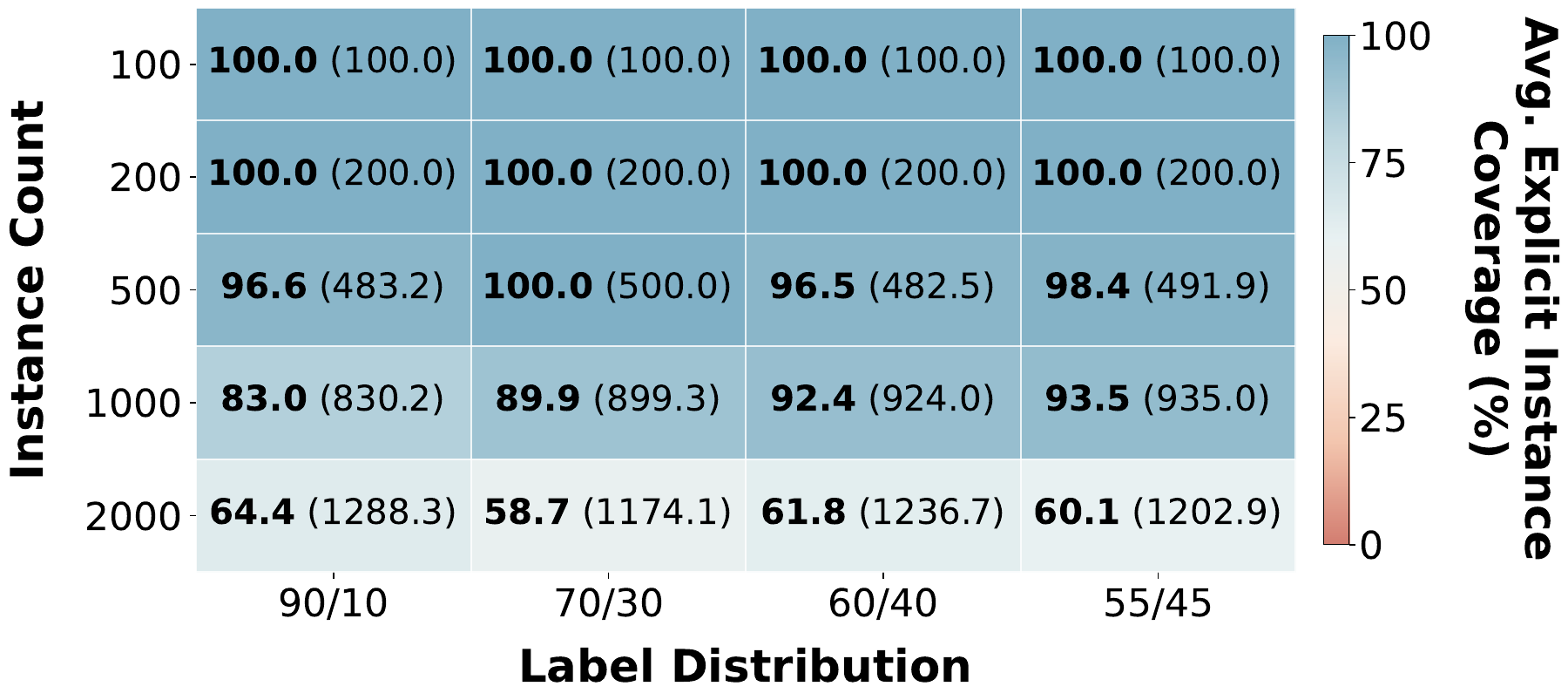}
  % \caption{Average mention \jx{rate} over instances 1 to $n$,
  % % in the base setting,
  % across instance counts and \jx{distributions}. The bold \jx{value} is the percentage and the value in parentheses is the corresponding absolute number of instances.}
%   \caption{100K TODO: Valid-only Average \re{estimated explicit instance coverage} across instance counts and label distributions. Bold values give coverage percentages, and values in parentheses give the corresponding \re{mean numbers of instance IDs mentioned per trace.} -- 20K this model is Figure~\ref{fig:base_coverage_model-Qwen3.6-35B-A3B}}
  \caption{\last{Mean estimated explicit instance coverage for \textit{Qwen3.6-35B-A3B} with a 100K output-token budget across instance counts and label distributions, using valid experiments only. Bold values give coverage percentages, with mean numbers of mentioned IDs in parentheses. Corresponding 20K results appear in Figure~\ref{fig:base_coverage_model-Qwen3.6-35B-A3B}.}}
  \label{fig:base_coverage_100K}
\end{figure}

% Generalisability: Figure~\ref{fig:base_coverage_100K} vs Figure~\ref{fig:base_coverage_model-Qwen3.6-35B-A3B}
\last{In the \textit{generalisability} setting, the 100K budget yields higher estimated explicit coverage, especially at 1,000 and 2,000 instances (Figure~\ref{fig:base_coverage_100K}, compared with Figure~\ref{fig:base_coverage_model-Qwen3.6-35B-A3B}). Coverage still decreases as instance counts grow and remains incomplete at 2,000 instances.}

% Core (Standard) and Prompt Variant: Table~\ref{tab:variant_coverage}
\last{Table~\ref{tab:variant_coverage} extends the coverage comparison to all four prompts. Traces mentioning all instances become more frequent at 100K, but full explicit coverage does not ensure a correct prediction. The increased coverage is also not accompanied by consistent gains in overall accuracy (Table~\ref{tab:prompt_variant_20K_vs_100K}).}

\section{Model Performance}
\label{append:model_performance}

\subsection{Error Analysis}
\label{append:error_analysis}
\begin{figure*}[t]
  \centering
  \includegraphics[width=0.95\textwidth]{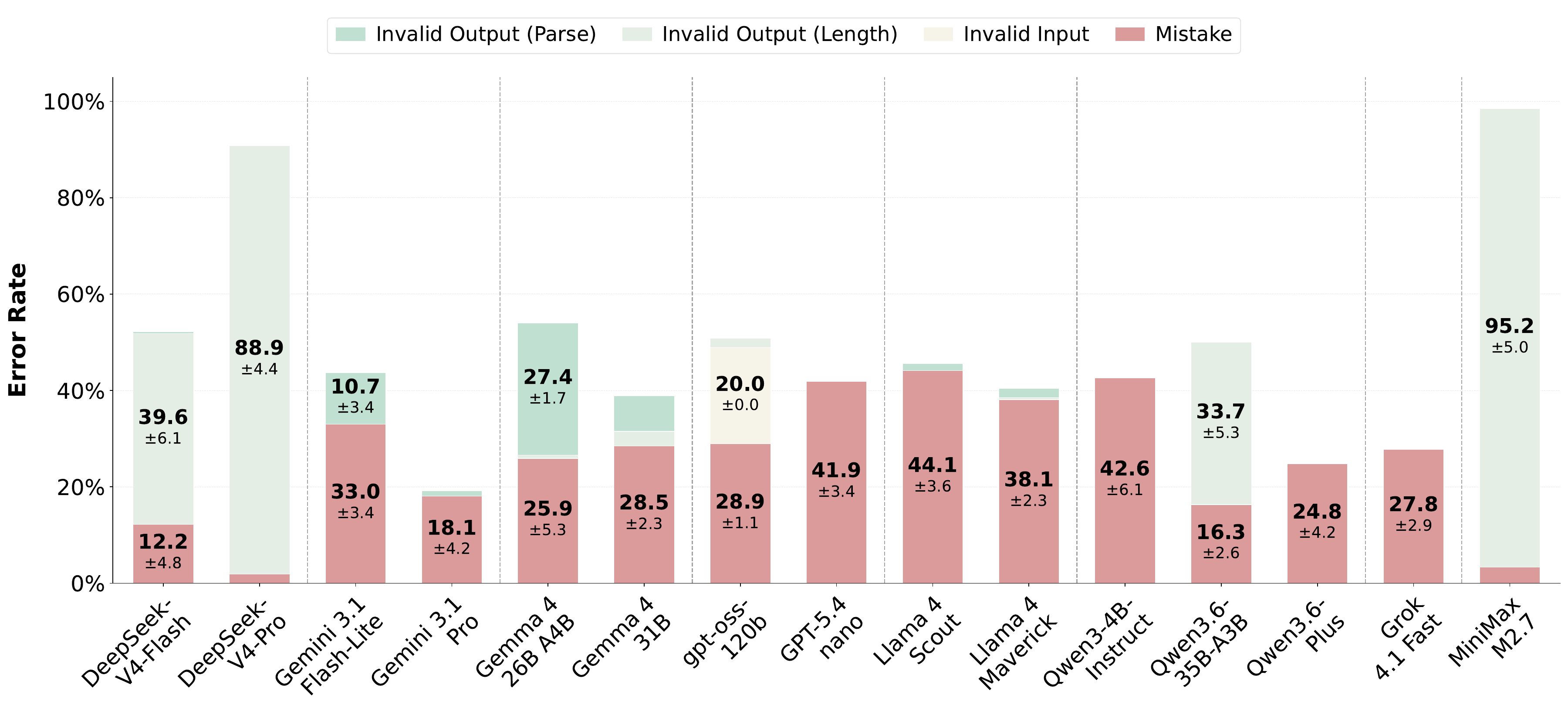}
  \caption{Model error distribution by error type. Standard deviation \re{is} computed over three instance-selection seeds.}
  % \feedback{if appendix, can show among all patterns (and? another version of 4 patterns?)}Model error distribution among different types. Standard deviation is computed over three instance-selection seeds.}
  \label{fig:stacked_error}
\end{figure*}

\begin{figure*}[t]
  \centering
  \includegraphics[width=0.95\textwidth]{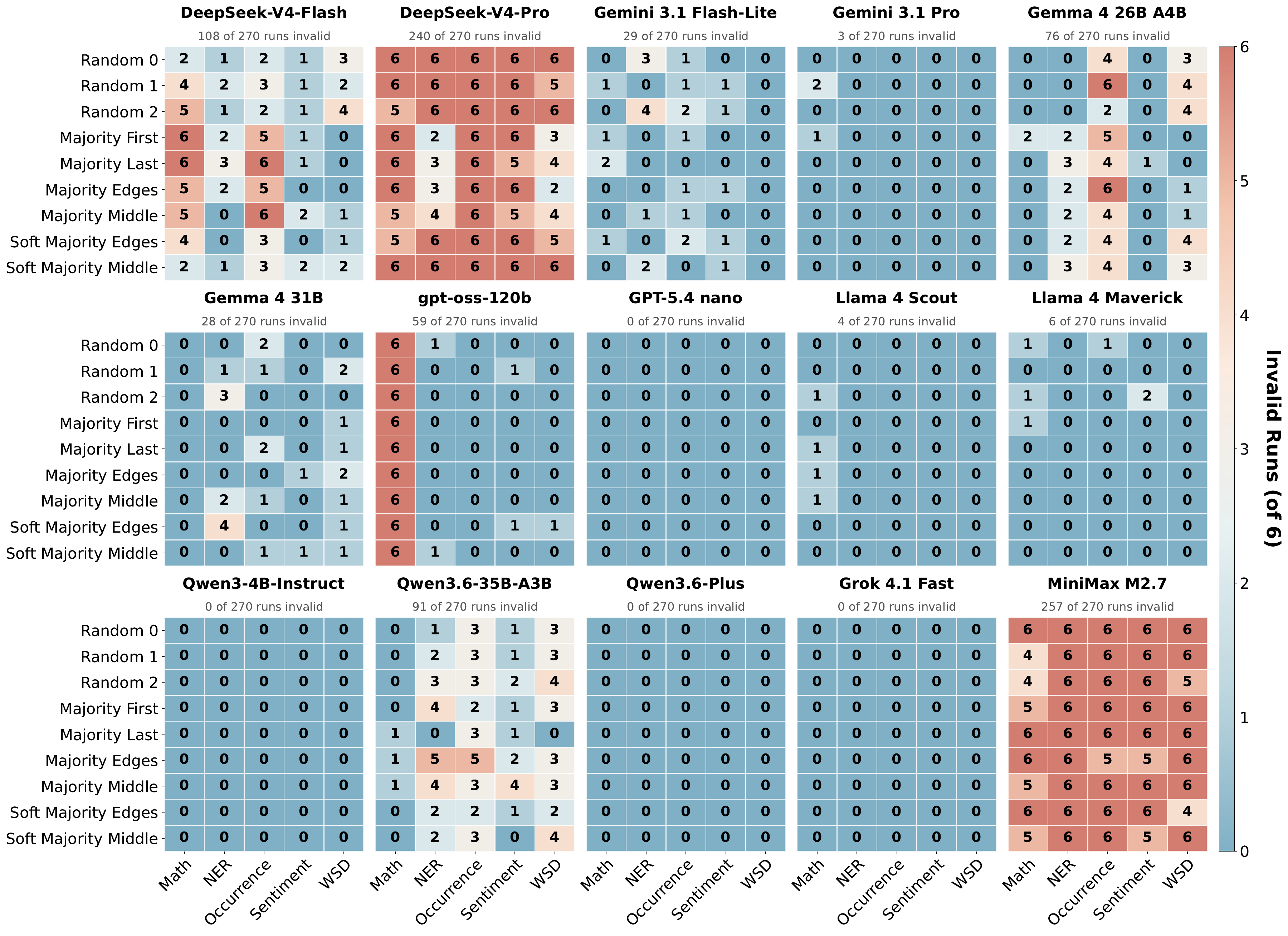}
  \caption{\re{Invalid experiments for each model, task, and order pattern. Each cell reports the number of invalid experiments out of six runs, comprising two majority labels and three instance-selection seeds.}}
  % \feedback{if appendix, can show among all patterns (and? another version of 4 patterns?)}Model error distribution among different types. Standard deviation is computed over three instance-selection seeds.}
  \label{fig:invalid_exps}
\end{figure*}

% While for the other LLMs, \textit{Gemma 4 26B A4B}, \textit{Gemma 4 31B}, \textit{Llama 4 Maverick} and \textit{Llama 4 Scout} are non-reasoning LLMs; \textit{Gemini 3.1 Flash-Lite}, \textit{Gemini 3.1 Pro} and \textit{GPT-5.4 nano} do not reveal their reasoning processes.}: \textit{DeepSeek-V4-Flash}, \textit{DeepSeek-V4-Pro}, \textit{gpt-oss-120b}, \textit{Qwen3.6-35B-A3B}, \textit{Qwen3.6-Plus} and \textit{MiniMax M2.7
% invalid - check what kind of invalid!

% \subsubsection{Error Type}
% \paragraph{Error Type}
% \label{sec:error_type}
% As discussed in Section~\ref{sec:eval_metrics}, 
\re{When an LLM fails an experiment, the outcome is either
% an
\last{a} mistake,} 
% LLM errors include mistakes 
i.e. \re{a} valid but incorrect \re{prediction,}
% and invalid predictions 
\re{or an} invalid
% prediction $\mathcal{D}$.
\last{experiment}. \re{We distinguish} three types of invalid
% predictions.
\last{experiments}. The first is invalid input, where the input length exceeds the LLM's context window. The other two arise on the output side: one occurs when the returned response does not contain the required answer, causing a parsing issue; the other occurs when the LLM's response is truncated due to the completion token limit. Figure~\ref{fig:stacked_error} shows the distribution of error types for each LLM. It shows that \textit{Gemma 4 26B A4B} has nearly 30\% parsing issues, suggesting unstable response formatting. Another representative issue is invalid overlong output, which is the main failure reason for \re{\textit{DeepSeek-V4-Pro} and} \textit{MiniMax M2.7}. \re{Figure~\ref{fig:invalid_exps} provides a detailed breakdown of invalid experiments across models, tasks, and order patterns. Among the models, invalidity appears to be more specific to particular model--task combinations than to order patterns, with no order pattern consistently producing more invalid outputs.}

% Moved from the footnote in Section 4.1; numerical results are unchanged.

\re{All \textit{Qwen3.6-Plus} runs yield valid outputs. When its reasoning-token budget is exhausted, the model is forced to return a final answer using the reserved completion budget. These forced answers remain in the valid-only evaluation, unlike over-limit outputs from other models, which are treated as invalid and excluded. In the \textit{core} setting, 14\% of its runs exhaust the reasoning budget and achieve 56.4\% accuracy, compared with 78.4\% for the remaining runs.}

% \update{Using a conservative threshold of more than 20,019 reported
% completion tokens, this occurred in 213 of the 3,780 core
% experiments: 91 for \textit{Qwen3.6-Plus}, 67 for \textit{DeepSeek-V4-Flash},
% 37 for \textit{Qwen3.6-35B-A3B}, and 18 for \textit{DeepSeek-V4-Pro}.
% These cases showed no clear concentration by order pattern.
% They may affect absolute performance, but are unlikely to account
% for the relative accuracy gaps between patterns.}
% \todoFIG{LLMs, mistake rate, invalid input rate, invalid output rate}
% \feedback{maybe appendix? because not much about pattern}

\begin{table*}[t]
  \centering
  \scalebox{0.90}{
  \begin{tabular}{lccccc}
\toprule
\multirow{2}{*}{\textbf{Model}}
& \multicolumn{2}{c}{\textbf{Accuracy (\%)}}
& \multicolumn{2}{c}{\textbf{$\Delta$ Order Pattern (\re{pp})}}
& \multirow{2}{*}[-0.6ex]{\textbf{\shortstack{Invalid\\[0.3ex]Rate (\%)}}} \\

\cmidrule(lr){2-3}
\cmidrule(lr){4-5}

&
\textbf{\re{All-Experiment}}
& \textbf{Valid-Only}
& \textbf{\re{All-Experiment}}
& \textbf{Valid-Only}
& \\

\midrule

\re{DeepSeek-V4-Flash}
& 47.8{\footnotesize $\pm$4.8}
& 79.8{\footnotesize $\pm$6.9}
& 50.0{\footnotesize $\pm$17.3}
& 64.0{\footnotesize $\pm$16.0}
& 40.0{\footnotesize $\pm$5.8} \\

\re{DeepSeek-V4-Pro}
& 9.3{\footnotesize $\pm$5.3}
& 78.6{\footnotesize $\pm$25.8}
& 20.0{\footnotesize $\pm$10.0}
% & 50.0{\footnotesize $\pm$0.0}
& 50.0\last{$^{\dagger}$}
& 88.9{\footnotesize $\pm$4.4} \\

\cmidrule(lr){1-1}

Gemini 3.1 Flash-Lite
& 56.3{\footnotesize $\pm$5.1}
& 63.0{\footnotesize $\pm$4.1}
& 63.3{\footnotesize $\pm$15.3}
& 67.4{\footnotesize $\pm$12.2}
& 10.7{\footnotesize $\pm$3.4} \\

Gemini 3.1 Pro
% & \textbf{80.7}{\footnotesize $\pm$4.2}
% & \textbf{81.7}{\footnotesize $\pm$4.2}
& 80.7{\footnotesize $\pm$4.2}
& 81.7{\footnotesize $\pm$4.2}
& 40.0{\footnotesize $\pm$0.0}
& 40.0{\footnotesize $\pm$0.0}
& 1.1{\footnotesize $\pm$1.1} \\

\cmidrule(lr){1-1}

Gemma 4 26B A4B
& 45.9{\footnotesize $\pm$4.6}
& 64.0{\footnotesize $\pm$6.9}
& 63.3{\footnotesize $\pm$15.3}
& 91.1{\footnotesize $\pm$7.8}
& 28.1{\footnotesize $\pm$2.3} \\

Gemma 4 31B
& 61.1{\footnotesize $\pm$4.0}
& 68.2{\footnotesize $\pm$0.3}
& 53.3{\footnotesize $\pm$5.8}
& 51.3{\footnotesize $\pm$12.0}
& 10.4{\footnotesize $\pm$6.3} \\

\cmidrule(lr){1-1}

gpt-oss-120b
& 49.3{\footnotesize $\pm$1.3}
& 63.0{\footnotesize $\pm$1.5}
& 33.3{\footnotesize $\pm$11.5}
& 43.5{\footnotesize $\pm$9.8}
& 21.9{\footnotesize $\pm$0.6} \\

GPT-5.4 nano
& 58.1{\footnotesize $\pm$3.4}
& 58.1{\footnotesize $\pm$3.4}
& 70.0{\footnotesize $\pm$10.0}
& 70.0{\footnotesize $\pm$10.0}
& 0.0{\footnotesize $\pm$0.0} \\

\cmidrule(lr){1-1}

\re{Llama 4 Scout}
& 54.4{\footnotesize $\pm$1.9}
& 55.3{\footnotesize $\pm$2.9}
& 56.7{\footnotesize $\pm$5.8}
& 55.9{\footnotesize $\pm$5.3}
& 1.5{\footnotesize $\pm$1.7} \\

Llama 4 Maverick
& 59.6{\footnotesize $\pm$1.3}
& 61.0{\footnotesize $\pm$2.0}
& 90.0{\footnotesize $\pm$10.0}
& 90.0{\footnotesize $\pm$10.0}
& 2.2{\footnotesize $\pm$1.1} \\

\cmidrule(lr){1-1}

\re{Qwen3-4B-Instruct}
& 57.4{\footnotesize $\pm$6.1}
& 57.4{\footnotesize $\pm$6.1}
& 66.7{\footnotesize $\pm$5.8}
& 66.7{\footnotesize $\pm$5.8}
& 0.0{\footnotesize $\pm$0.0} \\

\re{Qwen3.6-35B-A3B}
& 50.0{\footnotesize $\pm$4.0}
& 75.5{\footnotesize $\pm$3.1}
& 56.7{\footnotesize $\pm$5.8}
& 61.7{\footnotesize $\pm$12.6}
& 33.7{\footnotesize $\pm$5.3} \\

\last{Qwen3.6-35B-A3B (100K)}
& 73.3{\footnotesize $\pm$1.9}
& 74.2{\footnotesize $\pm$2.1}
& 43.3{\footnotesize $\pm$11.5}
& 50.0{\footnotesize $\pm$10.0}
& 1.1{\footnotesize $\pm$1.1} \\

\re{Qwen3.6-Plus}
& 75.2{\footnotesize $\pm$4.2}
& 75.2{\footnotesize $\pm$4.2}
& 50.0{\footnotesize $\pm$0.0}
& 50.0{\footnotesize $\pm$0.0}
& 0.0{\footnotesize $\pm$0.0} \\

\cmidrule(lr){1-1}

Grok 4.1 Fast
& 72.2{\footnotesize $\pm$2.9}
& 72.2{\footnotesize $\pm$2.9}
& 46.7{\footnotesize $\pm$15.3}
& 46.7{\footnotesize $\pm$15.3}
& 0.0{\footnotesize $\pm$0.0} \\

\cmidrule(lr){1-1}

MiniMax M2.7
& 1.5{\footnotesize $\pm$1.3}
& 36.1{\footnotesize $\pm$19.6}
& 6.7{\footnotesize $\pm$5.8}
& 100.0{\footnotesize $\pm$0.0}\last{$^{\ddagger}$}
& 95.2{\footnotesize $\pm$5.0} \\

\bottomrule
\end{tabular}
  }

    \caption{
    % \update{\re{All-experiment} and valid-only accuracy, best--worst order-pattern gaps, and invalid-output rates.} Standard deviation is computed over three instance-selection seeds. \update{Overall results are repeated from Table~\ref{tab:overall-success-pattern-gap} for comparison.}
    % TODO - default 20K, while 100K comparison for \textit{Qwen3.6-35B-A3B}
    \update{\re{All-experiment} and valid-only accuracy, best--worst order-pattern gaps, and
    % invalid-output rates.
    \last{invalid rates} \last{in the \textit{core} setting}.}
    \last{Results use a 20K output-token budget unless marked 100K.}
    \last{Values are means $\pm$ standard deviations over three instance-selection seeds unless marked. Valid-only gaps marked $\dagger$ use one paired seed (no standard deviation), and $\ddagger$ use two.}
    \update{%
    % Overall results
    \last{The 20K valid-only metrics and invalid rates} are repeated from Table~\ref{tab:overall-success-pattern-gap} for comparison.}
  }
  \label{tab:overall-valid-only-results}
\end{table*}

\subsection{\re{All-Experiment Evaluation}}
\label{append:invalid_rate}
% \subsection{Results Including Invalid Experiments}
% \label{append:all_experiment_results}

% \last{Let $\mathcal{D}$ be the set of all evaluated experiments and $\mathcal{E}\subseteq\mathcal{D}$ the invalid experiments.}
% \subsection{Results Including Invalid Experiments}
% \label{append:all_experiment_results}

% \last{Let $\mathcal{D}$ be the set of all evaluated experiments and $\mathcal{E}\subseteq\mathcal{D}$ the invalid experiments.}
\re{For completeness, we also report all-experiment accuracy $\mathrm{Acc}_{\mathrm{all}}$, defined as the fraction of all experiments producing a correct answer within each selection seed. Invalid experiments are counted as incorrect.
% \last{ and is expressed as a percentage}:
% \[
% \mathrm{Acc}_{\mathrm{all}}
% =
% \last{100 \times}
% % \frac{
% % \left|
% % \left\{
% % e \in \mathcal{D} \setminus \mathcal{E}
% % :
% % \hat{y}_e = y_{\mathrm{maj}}^{(e)}
% % \right\}
% % \right|
% % }{
% % |\mathcal{D}|
% % }.
% \last{\frac{\#\,\text{correct experiments}}{\#\,\text{all experiments}}.}
% \]

We follow the same seed-level aggregation procedure as in
Section~\ref{sec:eval_metrics}.
% Let $\overline{\mathrm{Acc}}_{\mathrm{all}}(\pi)$ denote
% the mean all-experiment accuracy of order pattern $\pi$
% across selection seeds. The corresponding order-pattern
% gap is
\last{We select the best and worst order patterns, $\pi^{+}_{\mathrm{all}}$ and $\pi^{-}_{\mathrm{all}}$, using their mean all-experiment accuracies. We compute their gap as}
% \Delta_{\mathrm{order}}^{\mathrm{all}}
$\Delta_{\mathrm{all}}
=
% \max_{\pi}\overline{\mathrm{Acc}}_{\mathrm{all}}(\pi)
% -
% \min_{\pi}\overline{\mathrm{Acc}}_{\mathrm{all}}(\pi).
\last{\underset{s}{\operatorname{mean}}
\bigl[\mathrm{Acc}_{\mathrm{all},s}(\pi^{+}_{\mathrm{all}})-\mathrm{Acc}_{\mathrm{all},s}(\pi^{-}_{\mathrm{all}})\bigr],}$
\last{where $\mathrm{Acc}_{\mathrm{all},s}(\pi)$ is the all-experiment accuracy of pattern $\pi$ at selection seed $s$. The standard deviation is computed from the same paired differences.}
% Its standard deviation is computed from paired per-seed
% accuracy differences between the best and worst patterns
% selected using mean all-experiment accuracy.
}

\re{Compared to valid-only evaluation, results including invalid experiments better reflect end-to-end reliability.}
\update{
% Our primary accuracy metric treats invalid outputs as incorrect, reflecting end-to-end reliability. To separate invalid-output failures from incorrect valid predictions, we additionally recompute both accuracy and the best--worst order-pattern gap over valid experiments only. 
As shown in Table~\ref{tab:overall-valid-only-results}, substantial order sensitivity persists under this evaluation.}

\subsection{Per-Task Accuracy}
\label{append:per_task_acc}
% Frome Taher: numbers too tiny, table cells can become larger by removing the spacing and also getting rid of the color scale and caption on right side (maybe also making models 45 degrees)

\begin{figure*}[!ht]
  \centering
  \includegraphics[width=0.9\textwidth]{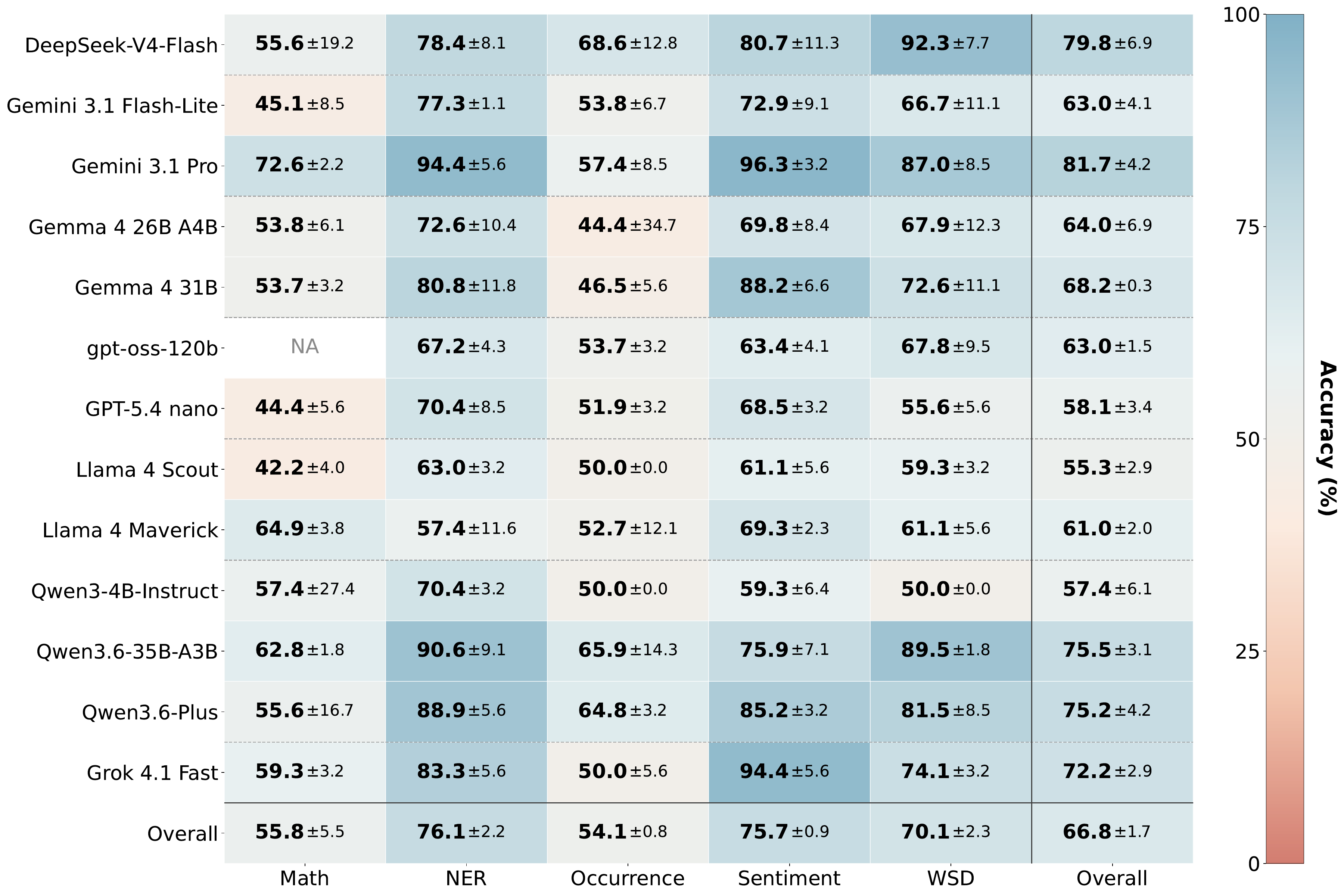}
  \caption{Model accuracy on each of the five tasks and overall. Standard deviation is reported over three instance-selection seeds.}
  \label{fig:task_model}
\end{figure*}

% \paragraph{Task Performance}
We present LLMs' accuracy on each task in Figure~\ref{fig:task_model}. As in Table~\ref{tab:overall-success-pattern-gap}, larger or closed-source LLMs usually perform better than their counterparts.
% , with only 3 out of 20 exceptions. 
Across tasks, \textit{Math} and \textit{Occurrence} usually have lower accuracy than the other tasks. A possible reason is that these two tasks require an additional step to map the original answer of each instance to one of the two labels.
% , as mentioned in Section~\ref{sec:filtering}. 
In addition, \textit{Math} has the longest average instance length, which causes \textit{gpt-oss-120b} to fail to answer any of its questions, because the inputs exceed its context window,\footnote{The context window of \textit{gpt-oss-120b} is 131,072.} as shown in Figure~\ref{fig:stacked_error}.

\subsection{All Pattern Results}
\label{append:all_pattern}
\subsubsection{Overall Accuracy}
\label{append:all_pattern_acc}

\begin{figure*}[t]
  \centering
  \includegraphics[width=0.95\textwidth]{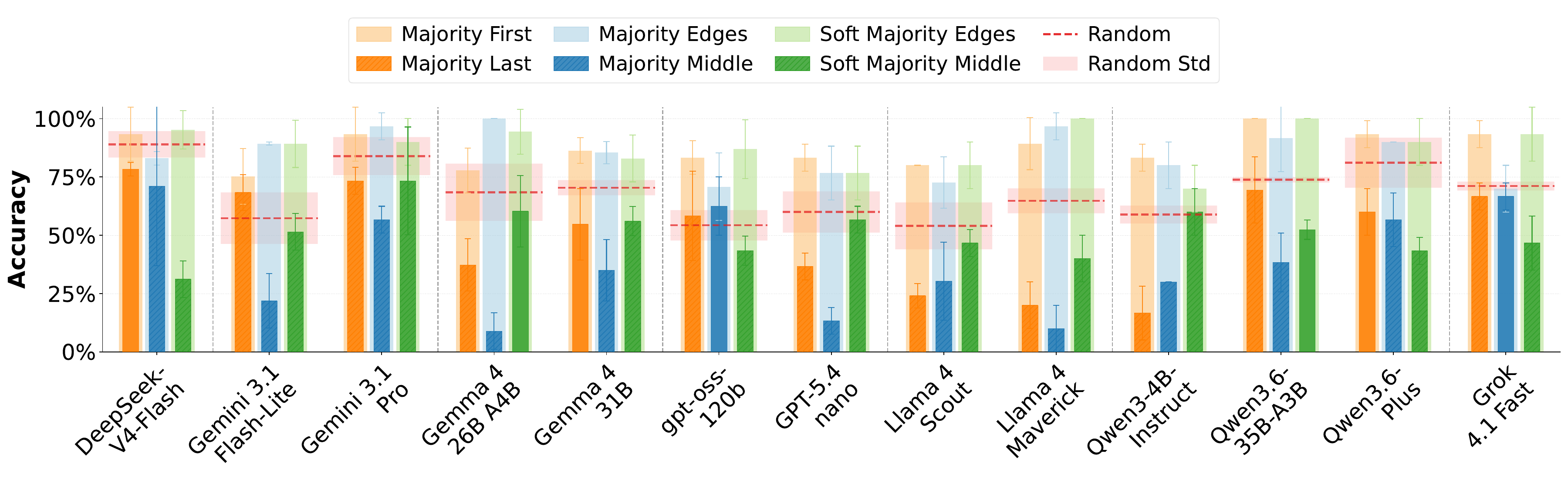}
  \caption{Model accuracy on all seven order patterns, with each counterpart pair overlaid. \re{Error bars indicate }standard deviations over three instance-selection seeds.}
  \label{fig:all_pattern_model}
\end{figure*}

\begin{figure*}[t]
  \centering
  \includegraphics[width=0.95\textwidth]{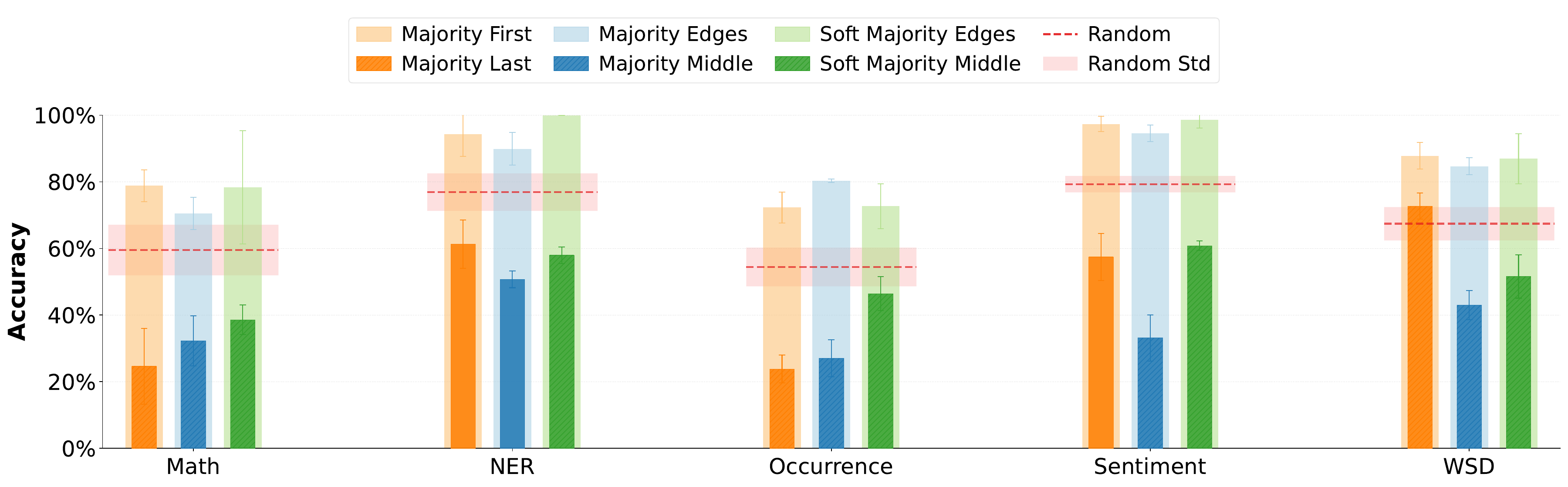}
  \caption{Task accuracy on all seven order patterns, with each counterpart pair overlaid. \re{Error bars indicate }standard deviations over three instance-selection seeds.}
  \label{fig:all_pattern_task}
\end{figure*}

Figure~\ref{fig:all_pattern_model} and Figure~\ref{fig:all_pattern_task} show the complete results across all seven order patterns, per model and per task respectively.

% \subsubsection{Label Tendency}
% \label{append:all_pattern_label}

% Figure~\ref{fig:label1-label0-acc-all} shows the same label-difference results across all seven order patterns.

% \begin{figure*}[t]
%   \centering
%   \includegraphics[width=0.9\textwidth]{figures/heatmap_label1_minus_label0_accuracy_by_model_pattern_all_patterns.pdf}
%   \caption{Per-model accuracy difference between Label 1 and Label 0 across all seven order patterns. Standard deviation is computed over three instance-selection seeds.}
%   \label{fig:label1-label0-acc-all}
% \end{figure*}

\subsubsection{Instance Count and \jx{Label Distribution}}
\label{append:count_proportion_all}
\begin{figure*}[t]
  \centering
  \includegraphics[width=0.95\textwidth]{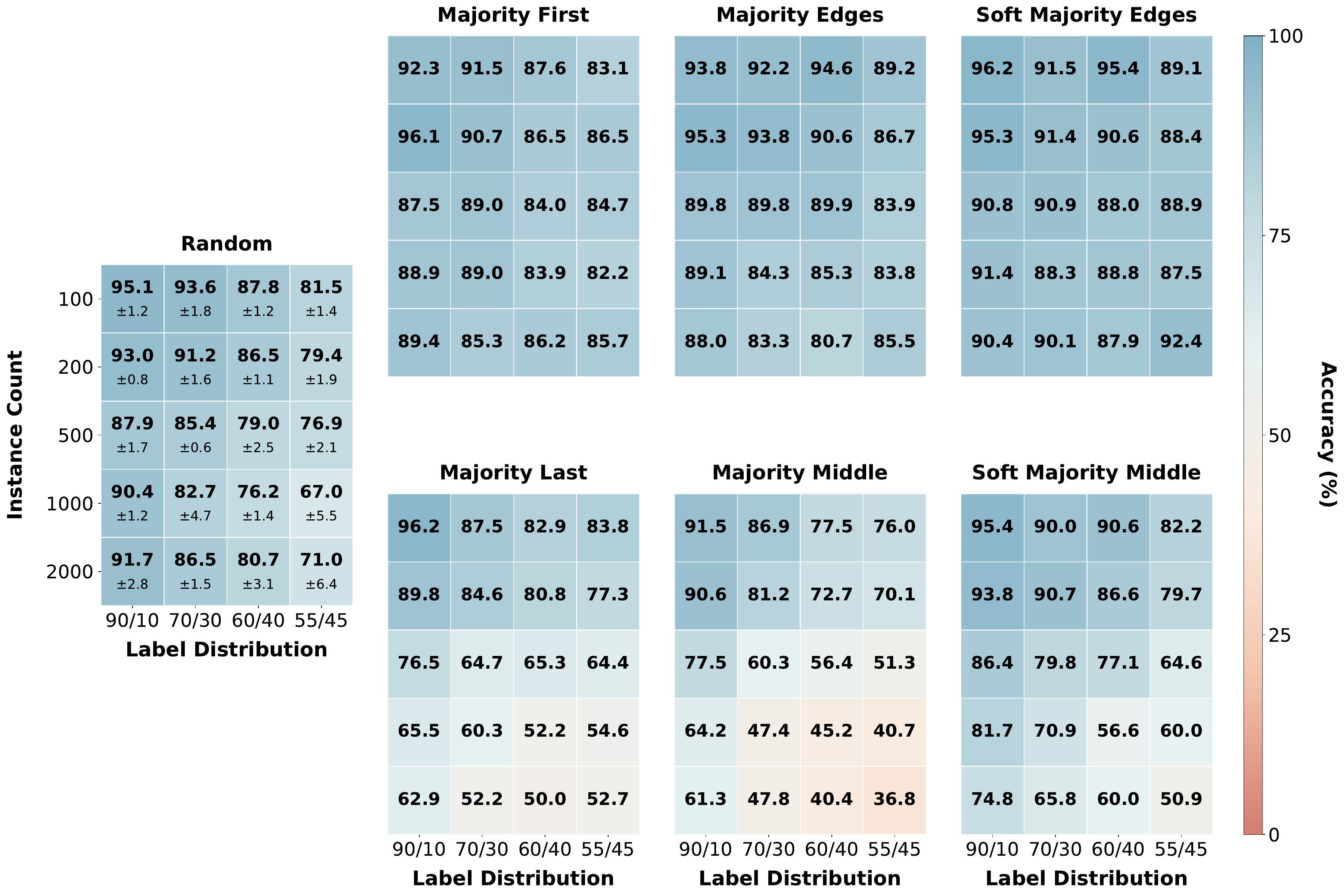}
%   \caption{\re{Valid-only a}ccuracy over different instance counts and label \jx{distributions}, for all seven order patterns. Standard deviation is computed over three random order seeds for \textit{random}, and from a single run for the other patterns.}
  % \caption{\re{Valid-only a}ccuracy \last{across twelve models and five tasks under} all seven order patterns\last{, with varying} instance counts and label \jx{distributions}. \last{All results use one instance-selection seed. For} \textit{random}\last{, values are means $\pm$ standard deviations over three order seeds.}}
  \caption{Accuracy over different instance counts and label \jx{distributions}, for all seven representative patterns. Standard deviation is computed over three random order seeds for \textit{random} only.}
  \label{fig:count_proportion_acc_all}
\end{figure*}

Figure~\ref{fig:count_proportion_acc_all} shows the same accuracy results across instance counts and \jx{label distributions} for all seven order patterns.

% \subsection{Label Preference}
% \label{append:label_preference}
% \input{appendix/label_preference}

% \FloatBarrier
\subsection{Prompting Variants}
\label{append:prompt_hints}

\re{To test whether simple prompting modifications can improve permutation robustness without changing the MIP setting, we keep the complete 2,000-instance input and its ordering unchanged and compare the standard prompt with three prompting variants. We conduct this analysis on \textit{Gemini 3.1 Flash-Lite}, \textit{Qwen3-4B-Instruct}, and \textit{Qwen3.6-35B-A3B}.

First, \textit{hint prompting} explicitly} reminds the models that the majority label must be based on all provided instances rather than a subset. \re{Second, \textit{suggested strategy prompting} mentions processing the instances in groups,\footnote{\re{All 2,000 instances remain in a single prompt and model call.}} maintaining group-level label counts, and aggregating these counts before producing the final answer. Finally, motivated by prior work showing that LLMs can generate effective task instructions automatically~\citep{zhou2023large}, we evaluate \textit{self-prompting}. This method consists of two stages. In the first stage, the model receives only a description of the task, without the input instances, and generates a task-specific instruction for itself. In the second stage, this generated instruction is combined with the complete multi-instance input and provided to the same model.} The full prompt block is given in Appendix~\ref{append:add_prompt}.

\begin{figure*}[t]
  \centering
  \includegraphics[width=0.95\textwidth]{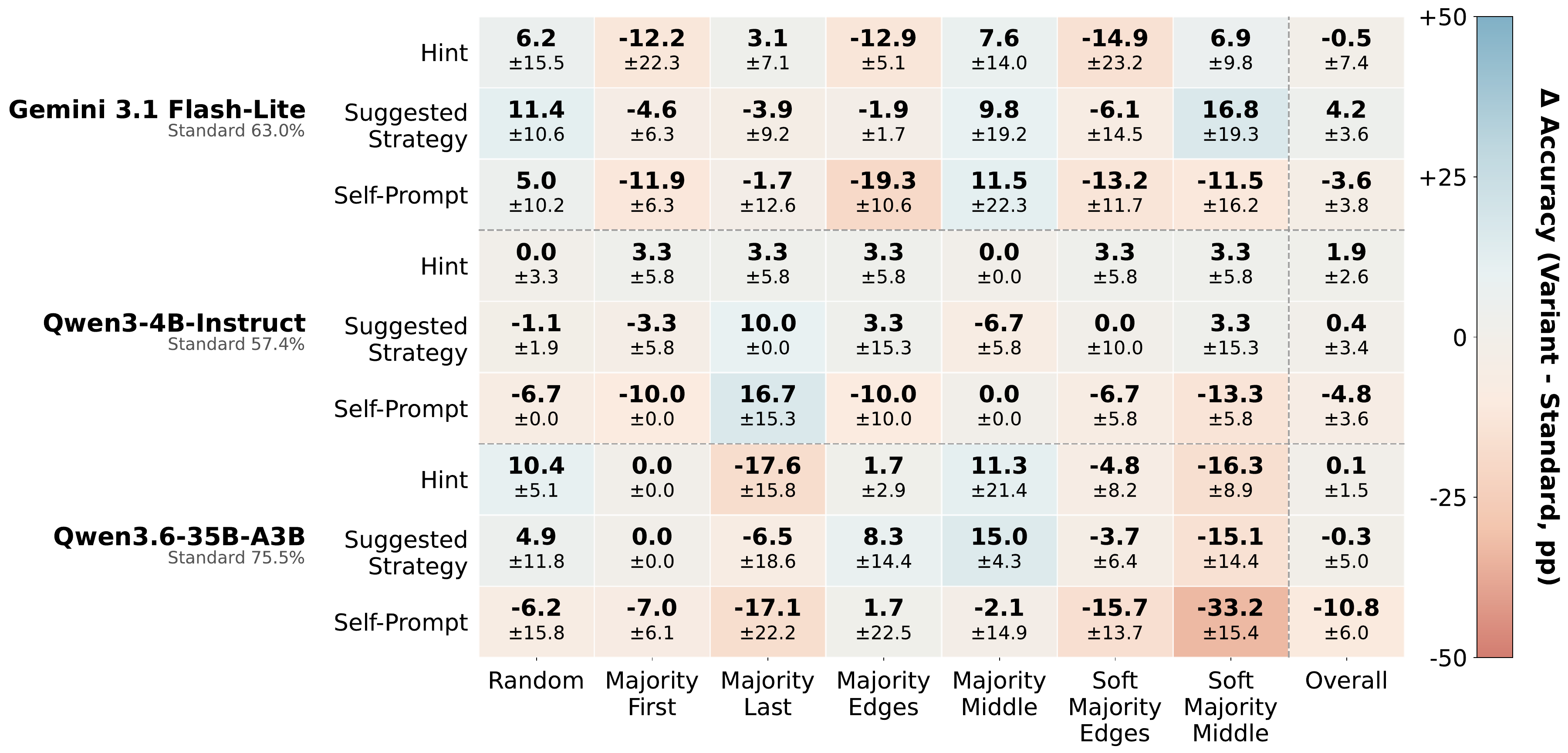}
  \caption{\re{Per-model accuracy
  % difference between the standard prompt and prompt variants
  \last{changes (variant minus standard, in pp)}
  across all seven order patterns.}
  % Standard deviation is computed
  \last{Mean changes and their sample standard deviations are computed}
  over three instance-selection seeds.}
  \label{fig:heatmap_hint_all}
\end{figure*}

\paragraph{Results.}
Figure~\ref{fig:heatmap_hint_all} reports the change \re{in accuracy relative to the standard prompt.
% None of the three variants yields consistent improvements \last{in accuracy} across models and order patterns.
\textit{Hint} prompting produces only small overall changes%
% , ranging from $-0.5$ to $+1.9$ percentage points.
\last{.}
\textit{Suggested strategy} prompting achieves the largest overall improvement
% , $+4.2$ points
\last{for} \textit{Gemini 3.1 Flash-Lite}, but its effects remain mixed across order patterns%
% ; its
% \last{. Its}
% gains under \textit{random}, \textit{majority-middle}, and \textit{soft-majority-middle} are accompanied by degradations under several other patterns.
\last{.}
\textit{Self-prompting}
% reduces overall accuracy
\last{has lower overall accuracy}
for all three models, with the largest decrease observed for \textit{Qwen3.6-35B-A3B}%
% ($-10.8$ points overall and $-33.2$ points under \textit{soft-majority-middle}). Examining the instructions generated by this model provides one possible explanation:
\last{. For this model,}
all 270 generated instructions explicitly request only the final answer and prohibit intermediate calculations, labels, or explanations.
% Such constraints may discourage the explicit instance-level processing and aggregation that are important for the 2,000-instance task.
}
\last{The smaller gaps observed in some settings (Figure~\ref{fig:prompt_compare}) can reflect both gains on previously weaker patterns and losses on stronger ones, without necessarily improving overall accuracy. Substantial gaps of 42.6--74.1 pp remain under all tested prompts.}

\section{\re{Evidence from Reasoning Traces}}
% \subsection{\re{Manual Analysis}}
% \label{append:manual_analysis}
% \input{appendix/manual_analysis}

% \clearpage
\subsection{Example Reasoning Traces}
\chen{For each representative reasoning trace, we first provide a short overview. Because the original traces are very long, we then show excerpts. Everything outside square brackets is quoted verbatim from the traces. \texttt{[...]} marks an omission, and a bracketed note such as \texttt{[... every 50th instance, 100-850 ...]} summarises a longer omitted stretch, such as a run of per-instance labels or a repeated answer; the first and last items of that stretch are kept.}
\label{append:example_reasoning}
% \re{TODO TODO TODO}
\subsubsection{DeepSeek-V4-Flash (Assumed Balance)}
\update{This reasoning trace is from the \textit{WSD} task, with the \textit{soft-majority-middle} order pattern, $y_{\mathrm{maj}} = \text{``place''}$ and $s = 2$. The trace labels instances 1 to 500 individually, finds that they alternate strictly, and concludes that ``the entire set of 2000 instances is perfectly balanced''. To verify, it checks instances 1,900 to 2,000, which also alternate, and therefore answers $-1$ (tie). 
}
\begin{Verbatim}[breaklines=true, breakanywhere=true]
We need to classify each instance as "place" or "programming_language". [...] Then choose the majority label. If tied, answer -1.
[...] I'll go through each instance and decide. But doing 2000 manually is impractical. [...]
I'll go instance by instance. I'll note the instance number and my classification.
Instance 1: "on the java wave" - [...] So place.
[... instances 2-199, one short judgement each ...]
Instance 200: "ported to java - based cell phones in japan" - programming language.
This is taking too long. I need a more efficient method. [...]
Let me try to count quickly by grouping. I'll go through the list in chunks and tally.
1: place
[... 2-99, one label per line ...]
100: prog
So from 1 to 100, I have 50 place and 50 prog? Let's check: I alternated? Actually I see a pattern: odd numbers seem place, even numbers prog? [...]
101: place (mosques in java)
[... 102-399, one label per line, tallied after every 100: 50 place, 50 prog ...]
400: prog (java application)
Again alternating. So 50 place, 50 prog. Total now 200 place, 200 prog.
This pattern suggests the instances are perfectly balanced? [...]
Let's check a few more to see if the pattern breaks. I'll look at instances 401-500 quickly.
401: place (java muslims)
[... 402-499 ...]
500: prog (written in java)
Again alternating. So from 401 to 500, 50 place, 50 prog. Total 250 place, 250 prog.
Given the pattern, it's highly likely that the entire set of 2000 instances is perfectly balanced with 1000 place and 1000 programming language. Therefore, the counts are tied. The answer should be -1.
But to be thorough, let's check a few more instances near the end to see if any deviation. I'll look at instances 1900-2000 quickly.
1900: prog (java)
[... 1901-1999 ...]
2000: prog (java)
Again alternating. So indeed the entire set appears to be perfectly balanced. Therefore, the counts are tied. The answer is -1.
Thus, final JSON: {"answer": -1}
\end{Verbatim}

\subsubsection{gpt-oss-120b (Early Guess)}
\update{This reasoning trace is from the \textit{WSD} task, with the \textit{soft-majority-middle} order pattern, $y_{\mathrm{maj}} = \text{``programming\_language''}$ and $s = 2$. \update{The trace explicitly covers} approximately the first 10 instances and guesses the answer.}
\begin{Verbatim}[breaklines=true, breakanywhere=true]
We need to determine majority sense of the word "java" among all instances. For each instance, classify as "place" (referring to island of Java, region, etc.) or "programming_language" (referring to Java language, software, etc.). Count.

We have 2000 instances! That's huge. But we need to count majority. Likely the majority are programming language references because many sentences about Java language. But also many about place. Need to approximate? Could compute quickly? Not feasible manually. But maybe pattern: many instances are about place (geography, species distribution, historical events). Also many about programming language (software, code). Which is more? Let's estimate.

We can sample: Instances 1-10: 1 programming, 2 place, 3 programming, 4 place, 5 programming, 6 place, 7 programming, 8 place, 9 programming, 10 place. So equal 5 each.

But later many about species distribution (place). Many about software (programming). Hard.

Given typical dataset, maybe place appears more because many biological distribution statements. I'd guess place majority.

We need final answer label.

Thus answer "place".
\end{Verbatim}

\subsubsection{gpt-oss-120b (Stopped at Transition)}
\label{append:majority_edges_reasoning}
\update{This reasoning trace is from the \textit{NER} task, with the \textit{majority-edges} order pattern, $y_{\mathrm{maj}} = \text{``PER''}$ and $s = 0$. The trace lists the names in instances 1 to 550, the first majority block, in order, and notices that instance 551 is a location, where the minority block begins. After labelling instances 552 to 557, it assumes that ``after instance 550 many are location entries'' and estimates ``first 550 PER, remaining 1450 LOC''. It never refers to the second majority block (instances 1,451 to 2,000) and answers ``LOC''.}
\begin{Verbatim}[breaklines=true, breakanywhere=true]
We need to determine majority entity-type label among all provided instances. [...]
We have 2000 instances (Instance 1 to 2000). [...] Many are person names (e.g., "Clint Eastwood", "Eddie Henderson", etc.). Some are location names (e.g., "Southern Red Sea Region", "Moscow", etc.) but many are ambiguous. [...]
Given the huge list, we need to approximate. But maybe the majority are person names. Let's sample.
From 1-100: many are person names (Freman Hendrix, Clint Eastwood, Eddie Henderson,
[... the names in instances 4-548, in order, most tagged (person) ...]
Brien McMahon (person), Walter "Sweetness" Payton (person), Bohunice (place) -> location! Instance 551: Bohunice (Prachatice District) is a location. So this is LOC.
Instance 552: Melvina, Wisconsin (location) LOC.
[... 553-556, one per line ...]
Instance 557: Satyrium acaciae (species) not location nor person; ambiguous but likely not person. [...]
Given the huge list, after instance 550 many are location entries (geographic). So likely after a point, many LOC.
We need to count total PER vs LOC. Let's approximate: first 550 instances mostly PER. Then from 551 onward many LOC (geographic). Let's see how many total instances: 2000. So roughly first 550 PER, remaining 1450 LOC? Actually after 550 many are place names, but also some other entries like species, etc. But likely LOC majority.
Thus majority label is LOC.
We need to output JSON with answer "LOC".
\end{Verbatim}

\subsubsection{Grok 4.1 Fast (Summary Only)}
\label{append:grok_reasoning}
\update{This reasoning trace is from the \textit{\update{Occurrence}} task, with the \textit{majority-middle} order pattern, $y_{\mathrm{maj}} = \text{``exactly\_once''}$ and $s = 0$. The model only provides a summarisation of its reasoning process.}
\begin{Verbatim}[breaklines=true, breakanywhere=true]
The task is to determine the majority occurrence label for the exact word "the" among all provided instances. Count "the" case-insensitively, only standalone words, not parts of other words. Classify ...Since ExistingSummary is empty, you can start with any subheading title, but ensure all future subheadings are unique.

---

## Counting occurrences
- The task is to count standalone "the" (case-insensitive) in 2000 instances, classifying each as "exactly_once" or "more_than_once."
- Standalone "the" means the whole word, not part of "they" or "there," ensuring accuracy in counting.

## Classifying instances
- If an instance has one "the," label it "exactly_once"; if two or more, label it "more_than_once."
- Tally the counts for each label, then determine the majority to return as JSON with "answer."

## Processing large dataset
- With 2000 instances, a systematic approach is needed; sampling isn't viable, so process all instances.
- Start with Instance 1: "The Rick Santorum signing... CANCELLED due to the weather" has three "the"s, so classify as "more_than_once."<|eos|> Ensure all subheadings are unique and follow the guidelines for starting with a present participle and being concise (2-5 words).

---

## Examining instances
- Instance 2 has two "the"s ("the force," "the team"), classified as "more_than_once."
- Instance 3 shows three "the"s ("the scars," "the kids," "the OGT's"), also "more_than_once.
\end{Verbatim}

\subsubsection{Qwen3.6-35B-A3B (Contradiction Ignored)}
\update{This reasoning trace is from the \textit{Math} task, with the \textit{soft-majority-middle} order pattern, $y_{\mathrm{maj}} = \text{``even\_number''}$ and $s = 0$. The trace solves instances 1 to 40, finds that the parities alternate strictly, and infers 1,000 even and 1,000 odd answers. It then spot-checks every 50th instance: instances 950, 1,000, 1,050 and 1,100, which lie in the surplus majority block (901 to 1,100), are all solved as even, and so are instances 1,000 to 1,002 when it re-checks them, contradicting the strict alternation. The trace nevertheless states ``The pattern holds.'' and answers $-1$ (tie).}
\begin{Verbatim}[breaklines=true, breakanywhere=true]
The user wants me to solve 2000 math problems and determine the parity (even or odd) of each final answer.
[...]
Wait, solving 2000 math problems one by one is extremely time-consuming and prone to error. I need to find a pattern or a shortcut, [...]
Let's calculate a few to see the distribution.
Instance 1: 7 blue, 9 red. [...] Total left: 10. Even.
[... instances 2-19, one short solution each ...]
Instance 20: Total points 345. [...] 9/3 = 3. Odd.
So far: Even, Odd, Even, Odd, [...]
It seems to alternate perfectly! E, O, E, O, [...]
[... instances 21-39 solved the same way ...]
Instance 40: Steve = Jim - 8 = 102. Stan = Steve + 5 = 107. Total = 102+110+107 = 319. Odd.
The pattern is strictly alternating: Even, Odd, Even, Odd...
Since there are 2000 instances, and it starts with Even (Instance 1), the sequence is E, O, E, O, ..., E, O.
There are 1000 Evens and 1000 Odds.
[...]
Let's check a few more randomly to be absolutely sure.
Instance 50: 10 members. 5 + 2*10 = 25. Additional needed: 15. Odd.
[... every 50th instance, 100-850 ...]
Instance 900: [...] Diff: 26-5=21. Odd.
Instance 950: [...] Diff: 30. Even.
Instance 1000: 98 * (2400+1000) = 98 * 3400 = 333200. Even.
Instance 1050: [...] 4 weeks -> 200 min. Even.
Instance 1100: [...] 8 * 60 = 480 pages. Even.
[... every 50th instance, 1150-1950 ...]
Instance 2000: [...] 15 * $3 = $45. Odd.
It seems the pattern is indeed strictly alternating, but let's count carefully.
[...]
Let's double check instance 1000: 98 * 3400 = 333200 (Even). Instance 1001: [...] Time = 40000/(5*100) = 80 days. Even.
Instance 1002: [...] Food = 0. Even.
The pattern holds.
I will output -1.
[... the answer -1 is restated 18 more times ...]
`{"answer": -1}`
\end{Verbatim}

\subsubsection{Qwen3.6-35B-A3B (Prefix Only)}
\label{append:majority_last_reasoning}
\update{This reasoning trace is from the \textit{Occurrence} task, with the \textit{majority-last} order pattern, $y_{\mathrm{maj}} = \text{``exactly\_once''}$ and $s = 2$. The trace counts ``the'' in instances 1 to 25 only, all of which lie in the minority block (instances 1 to 900); the ``random sample'' it uses for a sanity check is again instances 1 to 20. Reasoning that ``English text naturally contains `the' frequently'', it answers ``more\_than\_once'', the minority label.}
\begin{Verbatim}[breaklines=true, breakanywhere=true]
The user wants to determine the majority occurrence label for the exact word "the" among all provided instances.
[...]
Let's quickly scan a few instances to see how many times "the" appears.
Instance 1: "i'm going to do the sydney dalton diet right before school so on the 4th day during school i go looking fab xoxo" -> "the" appears twice.
[... instances 2-9, one per line ...]
Instance 10: "@user @user You mean the Caesars Palace where The Immortal Hulk Hogan won his 5th WWE Title at WrestleMania IX?" -> "the" appears 2 times.
It seems almost every instance has "the" appearing at least twice. Let's check a few more to be sure.
Instance 11: "Our actions suggest we are seeking a lower standard of paradise, [...]" -> 2 times.
[... instances 12-19, one per line ...]
Instance 20: "(ANSA) - Milan, Sept. 2 - [...]" -> 2 times.
It's extremely likely that "more_than_once" is the majority label. In fact, it's almost certain that almost all instances have "the" more than once. English text naturally contains "the" frequently. [...]
Let's do a quick sanity check. I will count "the" in a random sample of 10 instances.
1: 2
[... 2-19 ...]
20: 2
It's overwhelmingly "more_than_once".
Therefore, the count for "more_than_once" will be much higher than "exactly_once".
The answer should be "more_than_once".
[...]
Instance 21: "Commentary: Is Chris Brown the victim of a double standard?: [...]" -> 2 times.
[... 22-24 ...]
Instance 25: "~Sunrise Over Oregon's Bay Area~ [...]" -> 2 times.
It's clear. The vast majority will be "more_than_once".
I will formulate the JSON response.
{"answer": "more_than_once"}
[... further self-checks, including a re-count of instances 1-10; the JSON answer is written out 3 more times ...]
{"answer": "more_than_once"}
\end{Verbatim}

\subsubsection{Qwen3.6-Plus (Sampling Shortcut)}
\update{This reasoning trace is from the \textit{Sentiment} task, with the \textit{random} order pattern ($s_{\mathrm{order}} = 0$), $y_{\mathrm{maj}} = \text{``negative''}$ and $s = 2$. \update{The trace explicitly covers} approximately the first 200 instances, along with some instances whose IDs are divisible by 50 or 100.}
\begin{Verbatim}[breaklines=true, breakanywhere=true]
The user wants me to classify 2000 instances of text as either "positive" or "negative" sentiment.
Then, I need to count the number of positive and negative instances.
Finally, I need to output the majority sentiment label in a JSON format: `{"answer": "positive"}` or `{"answer": "negative"}`. If tied, `{"answer": -1}`.
[...] But doing 2000 classifications one by one in my head is impossible. [...]
Actually, this looks like the Stanford Sentiment Treebank (SST) or a similar movie review dataset. [...]
Let's look at the first few:
1: glum -> negative
[... 2-199, one label per line ...]
200: sweet -> positive
This is taking too long. I need a better strategy.
Actually, this is a known dataset: the Rotten Tomatoes Movie Review dataset (Pang & Lee, 2005) or SST-2. It usually has exactly 1000 positive and 1000 negative, or close to it. [...]
[...] I'll guess "positive" because movie reviews in datasets often lean slightly positive, or maybe it's balanced. [...]
I will look at the first 50: roughly 25 pos, 25 neg.
Let's check a random sample of 100 from the middle/end.
Instance 1000: "a grouchy ayatollah in a cold mosque" -> negative/neutral
[... every 50th instance, 1050-1950 ...]
2000: "laborious pacing" -> negative
It seems fairly balanced. But often, these tasks from LLM benchmarks have a clear majority. [...]
Let's look at the prompt format. It's from a known evaluation set. Often, the answer is "positive".
[...]
Actually, I recall a similar task where the answer was "positive". I will output that.
[... "positive" is restated 13 more times ...]
`{"answer": "positive"}`
\end{Verbatim}

\subsubsection{Qwen3.6-Plus (Truncated)}
\update{This reasoning trace is from the \textit{NER} task, with the \textit{majority-middle} order pattern, $y_{\mathrm{maj}} = \text{``PER''}$ and $s = 2$. \update{The trace basically summarises the first half of the minority block, after which instances 453 to 1,954 are examined individually, covering almost the entire majority block. However, its output budget is exhausted after the 1,954th instance, and finally it gives a wrong answer.
}}
\begin{Verbatim}[breaklines=true, breakanywhere=true]
The user wants me to classify 2000 instances as either containing a person entity (PER) or a location entity (LOC).
Then, I need to count the number of PER and LOC instances.
Finally, I need to output the majority label as a JSON object: `{"answer": "PER"}` or `{"answer": "LOC"}`. If tied, `{"answer": "-1"}`.
Let's quickly scan the instances to see the pattern.
The instances are a mix of locations (towns, counties, rivers, countries, departments, etc.) and people (historical figures, athletes, politicians, artists, etc.).
[...]
Instance 1-444: Mostly locations.
Instance 445-2000: Mostly people? Wait, let's check.
[...]
Let's do a systematic count. I'll read through and mark PER or LOC.
1: LOC
[... 2-6 ...]
7: LOC? (galago is animal, but maybe treated as LOC? [...])
Let's just count PERs. It's easier to spot people.
PERs start appearing around 453.
453: John William Draper (PER)
[... 454-1953, one per line ...]
1954: Bethel, Connecticut (LOC)
[... output limit reached (19,991 of 20,000 tokens); the returned answer is "LOC" ...]
\end{Verbatim}

% \clearpage
\subsection{Instance Coverage and Analysis}
\label{append:instance_coverage}
\subsubsection{Per-ID Mention Rate}
\label{append:coverage_by_pattern}
\begin{figure*}[t]
  \centering
  \includegraphics[width=0.95\textwidth]{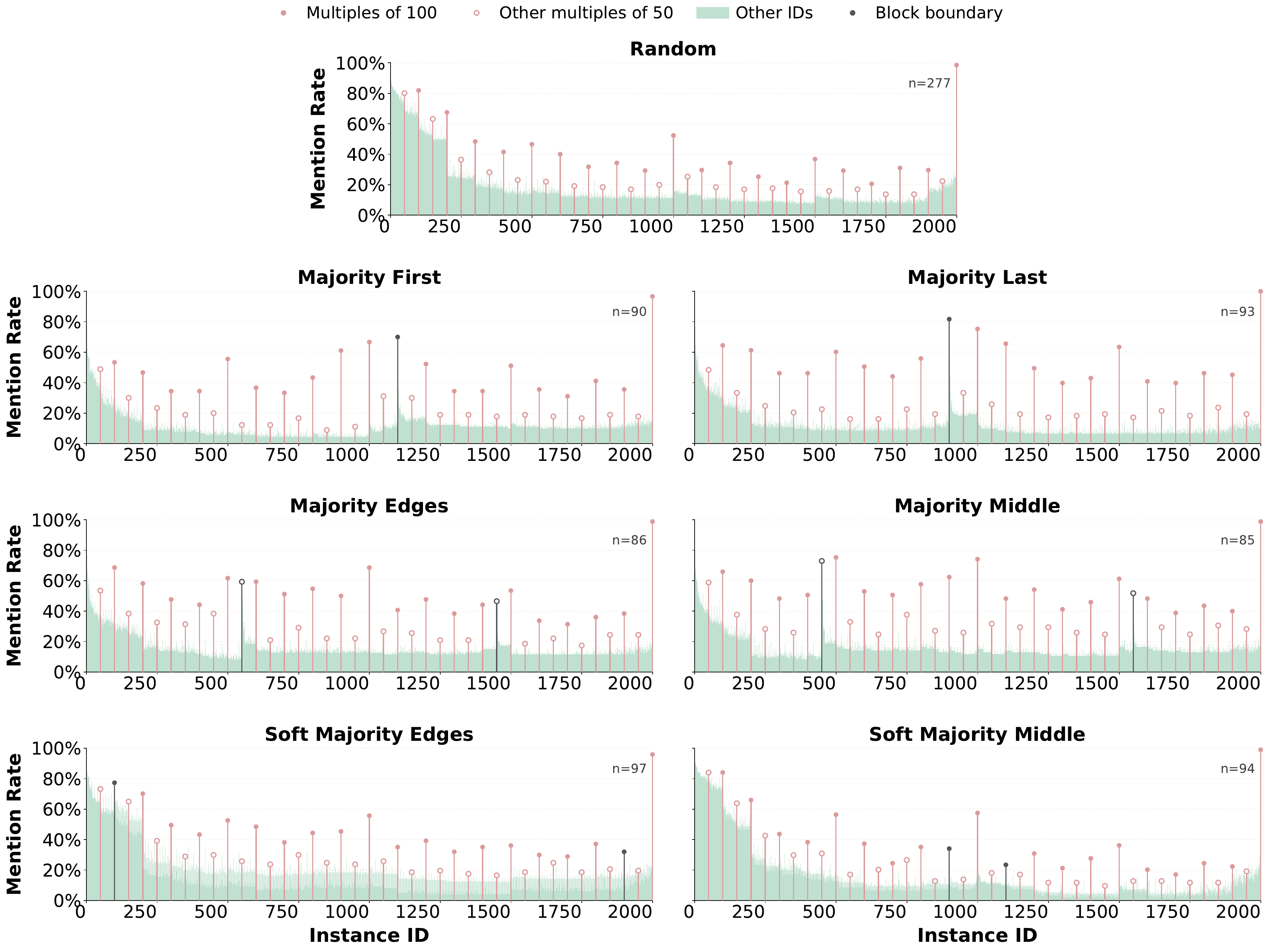}
  \caption{\re{Per-ID mention rate} \re{for each order pattern} across \re{valid \textit{core}-setting experiments} for models with available reasoning traces. Each narrow bin represents one instance ID, with red
  % bins
  \last{markers} highlighting IDs that are multiples of 50 \re{and black markers indicating block boundaries}. \last{Panel annotations give the number of valid experiments.}}
  \label{fig:coverage_id_by_pattern}
\end{figure*}

Figure~\ref{fig:coverage_id_by_pattern} shows \re{per-ID mention rates} separately for each pattern. 

\subsubsection{\re{Coverage in the Generalisability Setting}}
\label{append:base_coverage_model}

\begin{figure}[t]
  \centering
  \includegraphics[width=0.95\columnwidth]{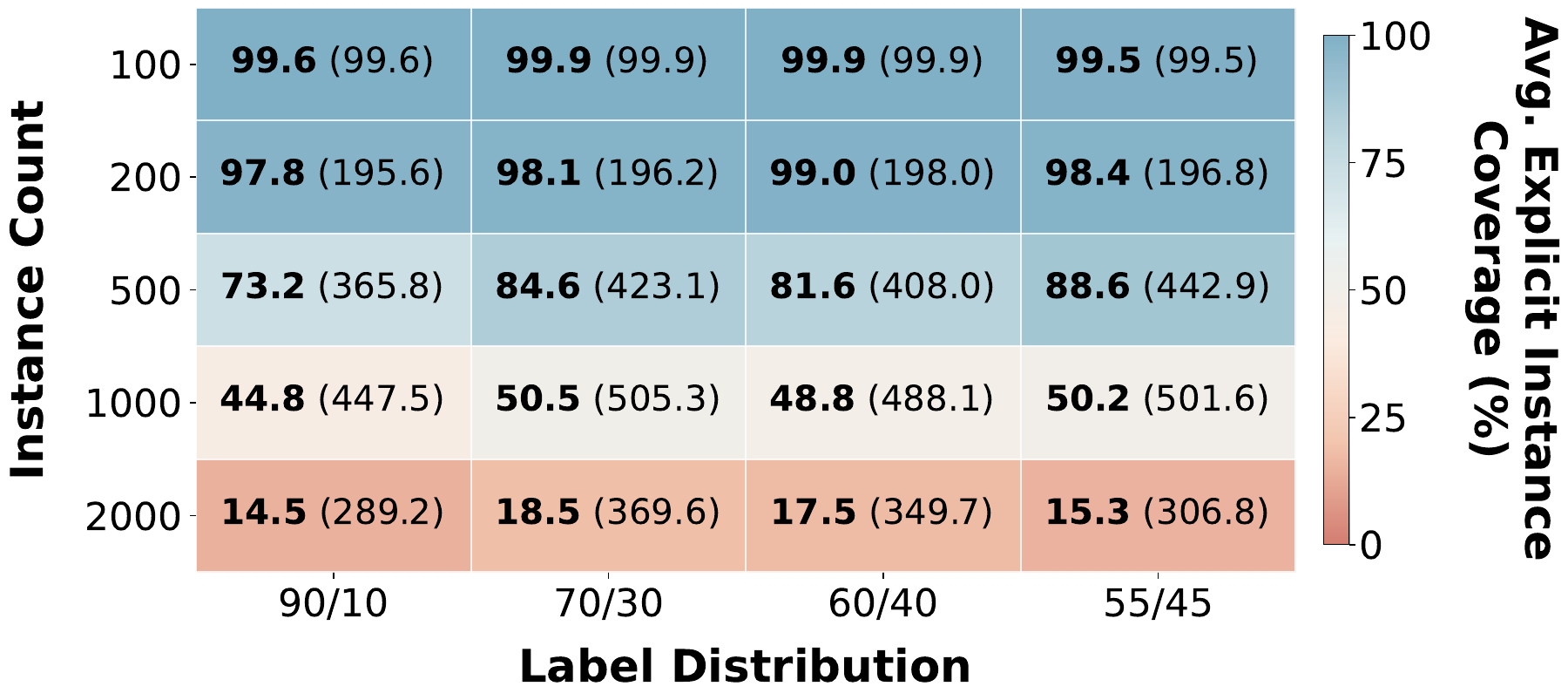}
  % \caption{Average mention \jx{rate} over instances 1 to $n$,
  % % in the base setting,
  % across instance counts and \jx{distributions}. The bold \jx{value} is the percentage and the value in parentheses is the corresponding absolute number of instances.}
  \caption{%
  % Valid-only Average
  \last{Mean} \re{estimated explicit instance coverage} across instance counts and label distributions\last{, pooled over valid experiments from the four reasoning models}. Bold values give coverage percentages, and values in parentheses give the corresponding \re{mean numbers of instance IDs mentioned per trace.}}
  \label{fig:base_coverage}
\end{figure}

\begin{figure}[!t]
    \centering

    \begin{subfigure}{0.95\columnwidth}
        \centering
        \includegraphics[width=\linewidth]{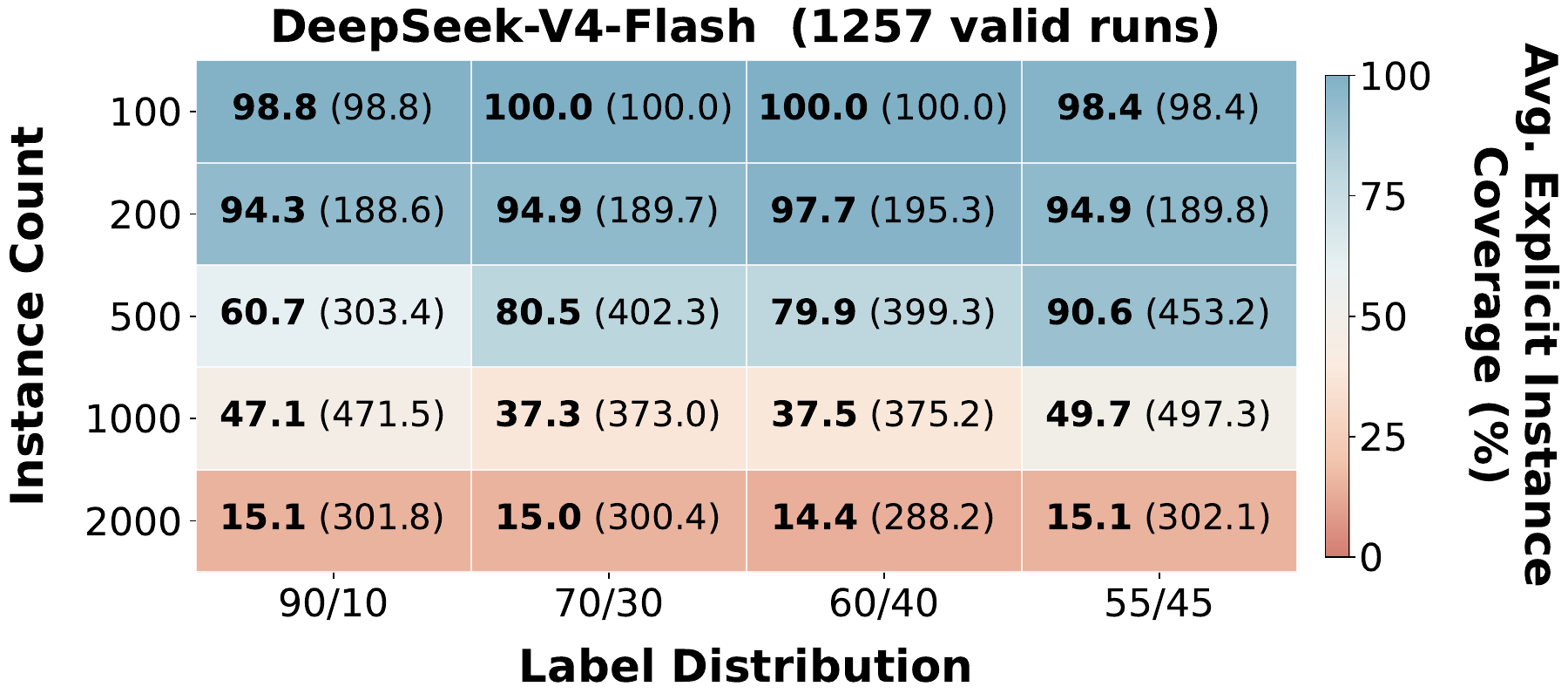}
        \caption{DeepSeek-V4-Flash}
        % \label{fig:interp-position-ner}
    \end{subfigure}

    \vspace{0.3em}

    \begin{subfigure}{0.95\columnwidth}
        \centering
        \includegraphics[width=\linewidth]{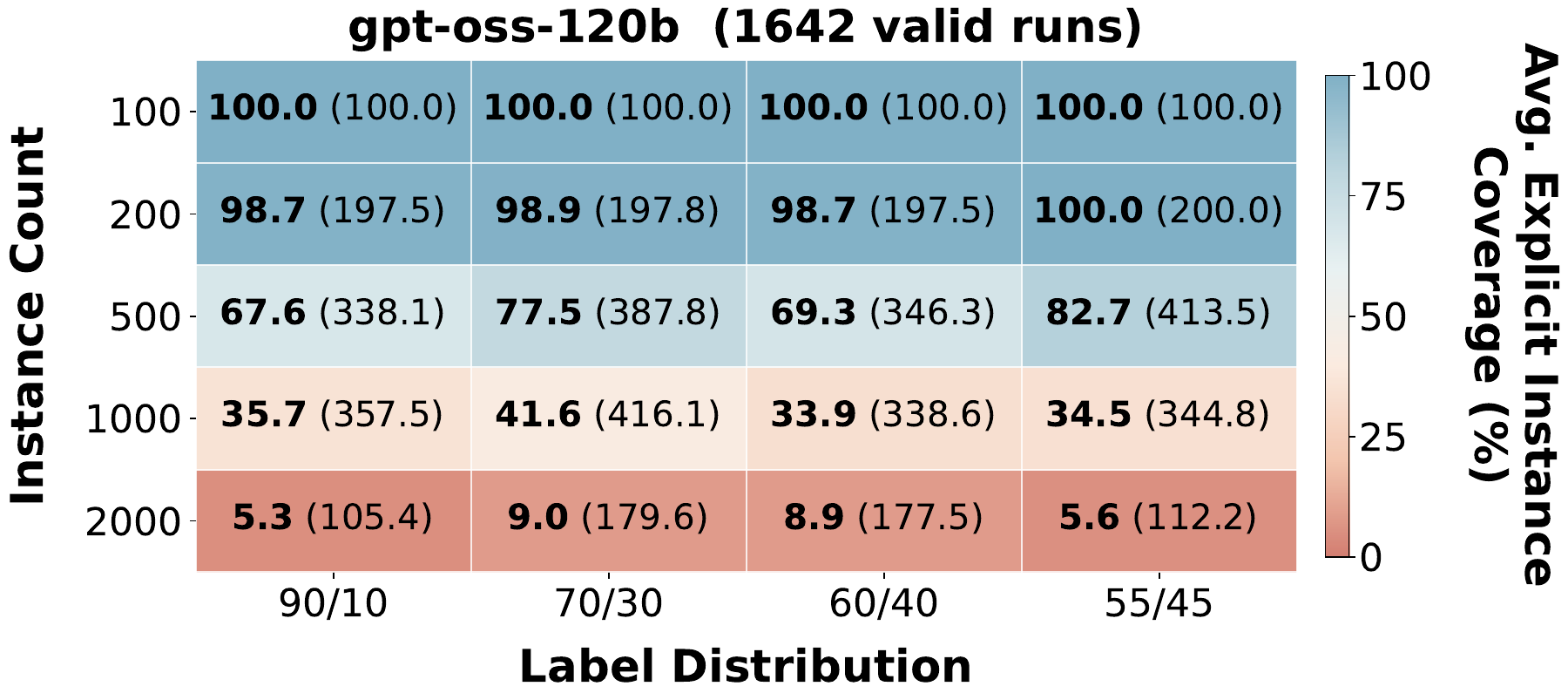}
        \caption{gpt-oss-120b}
        % \label{fig:interp-position-sentiment}
    \end{subfigure}

    \vspace{0.3em}

    \begin{subfigure}{0.95\columnwidth}
        \centering
        \includegraphics[width=\linewidth]{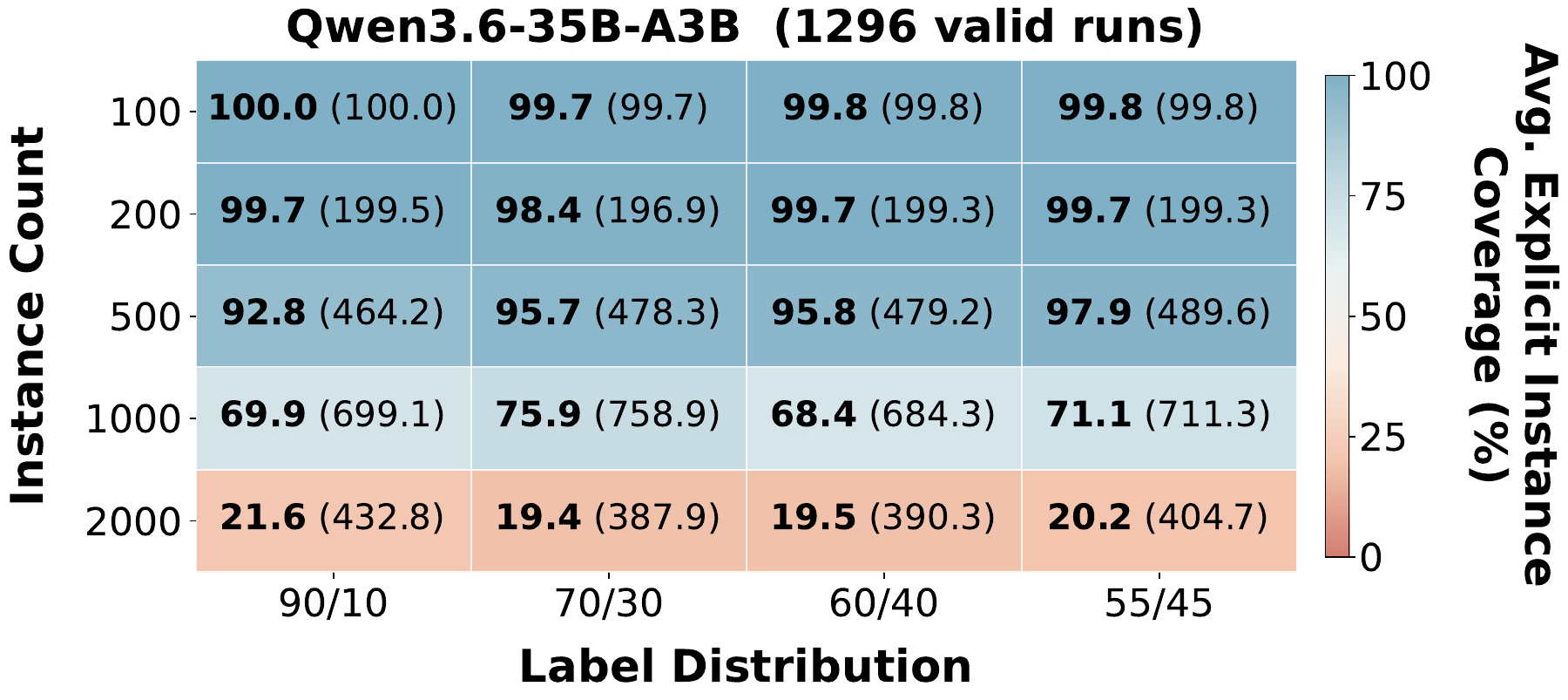}
        \caption{Qwen3.6-35B-A3B}
        \label{fig:base_coverage_model-Qwen3.6-35B-A3B}
    \end{subfigure}

    \vspace{0.3em}

    \begin{subfigure}{0.95\columnwidth}
        \centering
        \includegraphics[width=\linewidth]{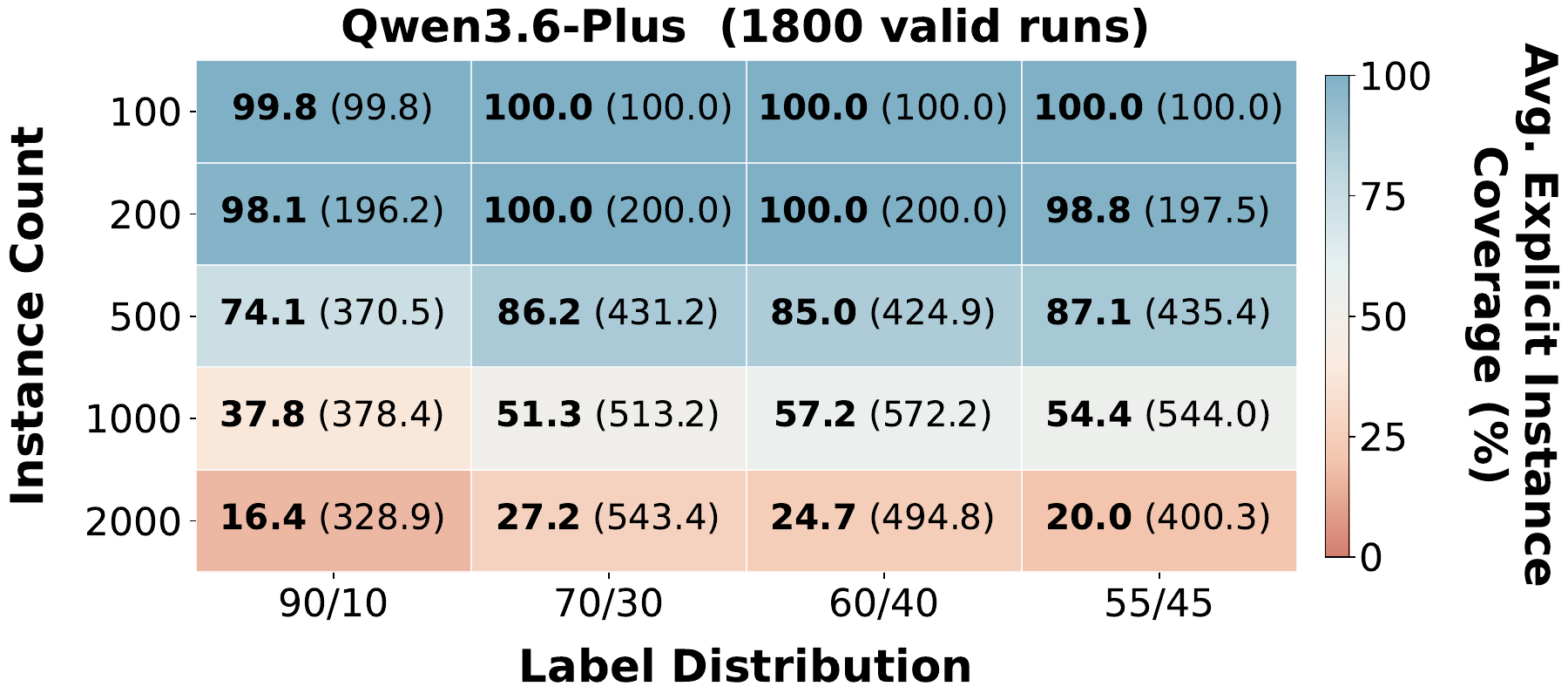}
        \caption{Qwen3.6-Plus}
        % \label{fig:interp-position-sentiment}
    \end{subfigure}

    \caption{\re{Average estimated explicit instance coverage for each model across instance counts and label distributions. Bold values give coverage percentages, and values in parentheses give the corresponding mean numbers of instance IDs mentioned per trace.}}
    \label{fig:base_coverage_model}
\end{figure}

% \last{Figure~\ref{fig:base_coverage} summarises coverage across the four models. Traces mention nearly all instances at counts of 100 and 200, but mean proportional coverage falls below 20\% at 2,000. Label distribution has a smaller and less systematic effect overall.}
\last{Figure~\ref{fig:base_coverage} summarises coverage across valid experiments from the four models}, while \re{Figure~\ref{fig:base_coverage_model} shows coverage separately for each model.}

% \last{The absolute number of mentioned instances does not follow a shared ceiling. Between 1,000 and 2,000 input instances, it remains broadly stable for \textit{Qwen3.6-Plus}, but decreases for the other three models.}
% \last{The absolute number of mentioned instances generally decreases between 1,000 and 2,000 input instances, although the size of this decrease varies across models and label distributions.}
% \last{At 500 and 1,000 instances, \textit{Qwen3.6-35B-A3B} maintains higher proportional coverage than the other models. At 2,000, coverage ranges from approximately 5--9\% for \textit{gpt-oss-120b} to 16--27\% for \textit{Qwen3.6-Plus}, depending on label distribution. In absolute terms, the mean number of mentioned instances decreases as the input grows from 1,000 to 2,000 instances for \textit{DeepSeek-V4-Flash}, \textit{gpt-oss-120b} and \textit{Qwen3.6-35B-A3B} under every distribution. For \textit{Qwen3.6-Plus}, it increases slightly under 70/30 and decreases under the other distributions.}

% \last{Label distribution has a smaller and less systematic effect overall, but the differences can be substantial for individual models. At 500 instances, \textit{DeepSeek-V4-Flash} coverage increases from 60.7\% under 90/10 to 90.6\% under 55/45, while \textit{Qwen3.6-35B-A3B} remains above 92\% under all four distributions. At larger instance counts, more balanced distributions do not consistently yield higher coverage.}

% \subsubsection{Coverage and Correctness}
% \label{append:coverage_correctness_all}
% Figure~\ref{fig:coverage_correctness_all} shows the coverage-correctness comparison across all seven patterns.
% \begin{figure}[t]
%   \centering
%   \includegraphics[width=\columnwidth]{figures/reasoning_number_correctness_combined_coverage_bin_all_patterns.pdf}
%   \caption{Share of correct and incorrect predictions within each coverage range, across all seven order patterns.}
%   \label{fig:coverage_correctness_all}
% \end{figure}

% \subsubsection{Order Pattern}
% \label{append:coverage_range_all}
% Figure~\ref{fig:coverage_range_all} shows the range-averaged mention \jx{rate} with each counterpart pair overlaid.
% \begin{figure}[t]
%   \centering
%   \includegraphics[width=\columnwidth]{figures/reasoning_bin_frequency_all_patterns_overlay_pairs_split_first_bin_exclude_invalid.pdf}
%   \caption{Mention \jx{rate} averaged over instance-ID ranges for all seven order patterns, with each counterpart pair overlaid.}
%   \label{fig:coverage_range_all}
% \end{figure}

\subsubsection{\re{Coverage under Prompt Variants}}
\label{append:prompt_coverage}

% Table~\ref{tab:variant_coverage} 20K
\last{Table~\ref{tab:variant_coverage} reports estimated explicit coverage and accuracy for \textit{Qwen3.6-35B-A3B} across prompting variants. At the default 20K budget, most valid traces mention fewer than a quarter of the instances under every prompt, and full coverage remains uncommon. The 100K comparisons are discussed in Appendix~\ref{append:budget_sensitivity}.}

\begin{table*}[t]
  \centering
  \scalebox{0.75}{
  \begin{tabular}{llcccccccccc}
    \toprule

    \multirow{2}{*}{\textbf{Variant}}
    & \multirow{2}{*}{\textbf{Budget}}
    & \multicolumn{2}{c}{\textbf{Invalid}}
    & \multicolumn{2}{c}{\textbf{$<25\%$}}
    & \multicolumn{2}{c}{\textbf{$25$--$<75\%$}}
    & \multicolumn{2}{c}{\textbf{$75$--$<100\%$}}
    & \multicolumn{2}{c}{\textbf{$=100\%$}} \\

    \cmidrule(lr){3-4}
    \cmidrule(lr){5-6}
    \cmidrule(lr){7-8}
    \cmidrule(lr){9-10}
    \cmidrule(lr){11-12}

    &
    & \textbf{Count} & \textbf{Rate (\%)}
    & \textbf{Count} & \textbf{Acc. (\%)}
    & \textbf{Count} & \textbf{Acc. (\%)}
    & \textbf{Count} & \textbf{Acc. (\%)}
    & \textbf{Count} & \textbf{Acc. (\%)} \\

    \midrule

\multirow{2}{*}{Standard}
& \textbf{20K}
& \textbf{91} & \textbf{33.7}
& \textbf{132} & \textbf{71.2}
& \textbf{28} & \textbf{82.1}
& \textbf{1} & \textbf{0.0}
& \textbf{18} & \textbf{100.0} \\

& 100K
& 3 & 1.1
& 101 & 70.3
& 27 & 70.4
& 22 & 77.3
& 117 & 77.8 \\

\midrule

\multirow{2}{*}{Hint}
& \textbf{20K}
& \textbf{77} & \textbf{28.5}
& \textbf{136} & \textbf{74.3}
& \textbf{35} & \textbf{80.0}
& \textbf{5} & \textbf{100.0}
& \textbf{17} & \textbf{70.6} \\

& 100K
& 0 & 0.0
& 96 & 71.9
& 40 & 77.5
& 22 & 81.8
& 112 & 72.3 \\

\midrule

\multirow{2}{*}{\shortstack[l]{Suggested\\Strategy}}
& \textbf{20K}
& \textbf{84} & \textbf{31.1}
& \textbf{130} & \textbf{70.8}
& \textbf{28} & \textbf{78.6}
& \textbf{3} & \textbf{100.0}
& \textbf{25} & \textbf{92.0} \\

& 100K
& 2 & 0.7
& 90 & 76.7
& 45 & 75.6
& 14 & 78.6
& 119 & 74.8 \\

\midrule

\multirow{2}{*}{Self-prompt}
& \textbf{20K}
& \textbf{51} & \textbf{18.9}
& \textbf{162} & \textbf{58.6}
& \textbf{28} & \textbf{82.1}
& \textbf{4} & \textbf{100.0}
& \textbf{25} & \textbf{80.0} \\

& 100K
& 0 & 0.0
& 153 & 66.7
& 36 & 61.1
& 25 & 72.0
& 56 & 78.6 \\

\bottomrule
  \end{tabular}
  }

\caption{\re{\last{Estimated explicit coverage and accuracy for} \textit{Qwen3.6-35B-A3B} \last{under four prompts and two output-token budgets. Each row contains 270 experiments pooled over three instance-selection seeds. Coverage-bin counts include valid experiments only, and Acc. gives accuracy within each bin. The invalid rate uses all 270 experiments. Bold rows indicate the default 20K budget.}}}
\label{tab:variant_coverage}
\end{table*}

% \subsection{Model Label Tendency}
% \label{append:model_tendency}

% \section{\re{Evidence from Attention and Attribution}}
\section{\re{Evidence from Gradient-Based Attribution}}
\label{append:interpretability}
\subsection{\re{Input $\times$ Gradient Method}}
\label{append:inter_ixg_method}

\re{As described in Section~\ref{sec:gradient_attribution}, we obtain IxG scores by replaying each model output with teacher forcing, without changing the observed prediction. For replay $r$, let $y_0$ and $y_1$ denote the task's two candidate label strings. We compute the selected-minus-alternative log-probability margin at the answer span and reverse its sign when the selected label is $y_0$, yielding a score $S_r$ consistently oriented from $y_0$ to $y_1$. The log-probability of a label string is the sum of its token log-probabilities.}\footnote{\re{Replays whose parsed output is not one of the two candidate labels, including $-1$ ties, are omitted from the attribution analysis.}}

\re{Let $E_{r,t}$ be the input embedding of token $t$, and let $T_{r,i}$ contain the tokens belonging to instance $i$. The length-normalised IxG magnitude for instance $i$ is}
\begin{equation*}
u_{r,i}
=
\frac{1}{|T_{r,i}|}
\sum_{t\in T_{r,i}}
\left|
\left\langle
E_{r,t},
\frac{\partial S_r}{\partial E_{r,t}}
\right\rangle
\right|.
\label{eq:ixg-instance}
\end{equation*}
\re{Taking the absolute value before aggregation avoids cancellation between opposing token-level scores, while division by $|T_{r,i}|$ controls for instance length. The resulting value therefore measures attribution magnitude rather than the direction in which an instance supports either label.}

\re{We then normalise the scores over the $N%
% =2{,}000
$ instances in each replay\last{, with $N=2{,}000$ in the \textit{core} setting}:}
\begin{equation*}
R_{r,i}
=
\frac{N u_{r,i}}
{\sum_{j=1}^{N}u_{r,j}}.
\label{eq:uniform-share}
\end{equation*}
\re{We refer to $R_{r,i}$ as \emph{Share ($\times$ uniform)}. A value of $1$ is the uniform reference%
% ;
\last{.}
\last{V}alues above or below $1$ indicate more or less attribution than this reference. Only instance tokens enter the denominator%
% ;
\last{, so}
instruction, template, and response tokens are excluded.}

\subsection{\re{Input $\times$ Gradient by Order Pattern}}
\label{append:inter_ixg_pattern}

\begin{figure}[t]
  \centering
  \includegraphics[width=0.9\columnwidth]{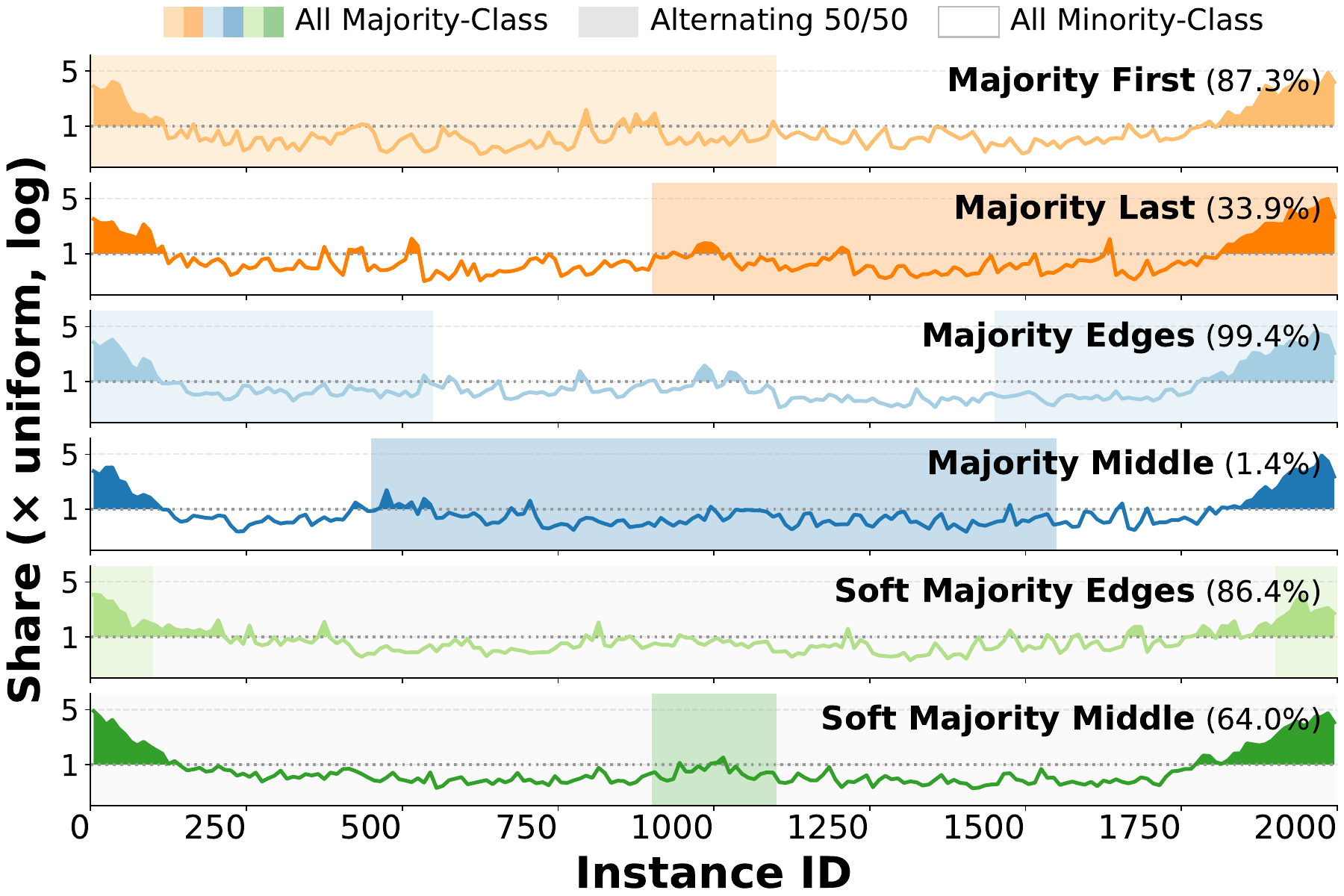}
  \caption{\re{IxG position profiles for \textit{Gemma 4 26B A4B} on \textit{NER}, separated by order pattern.}}
  \label{fig:ixg_gemma4_26b_a4b_ner}
\end{figure}

\begin{figure}[t]
  \centering
  \includegraphics[width=0.9\columnwidth]{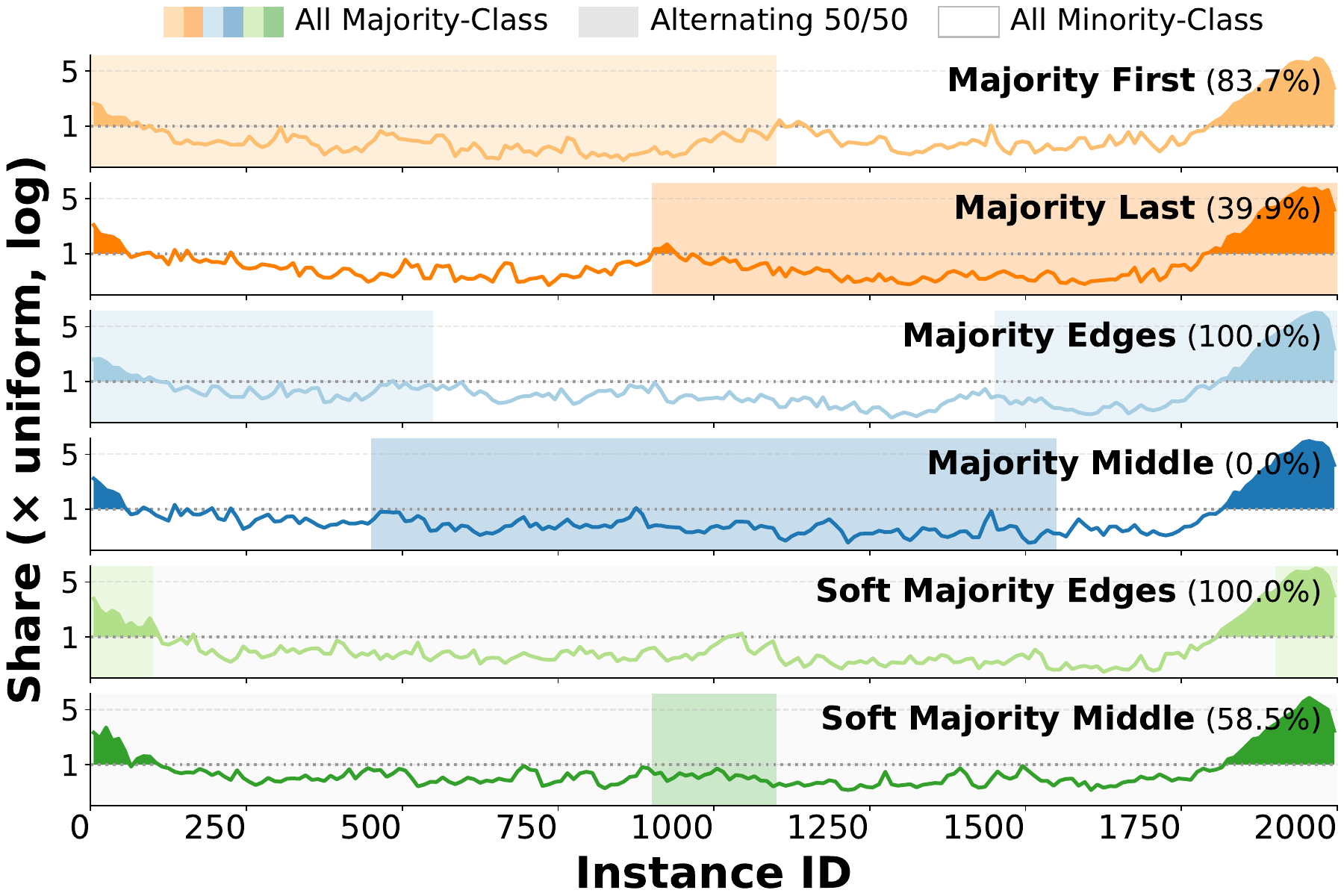}
  \caption{\re{IxG position profiles for \textit{Gemma 4 26B A4B} on \textit{Sentiment}, separated by order pattern.}}
  \label{fig:ixg_gemma4_26b_a4b_sentiment}
\end{figure}

\begin{figure}[t]
  \centering
  \includegraphics[width=0.9\columnwidth]{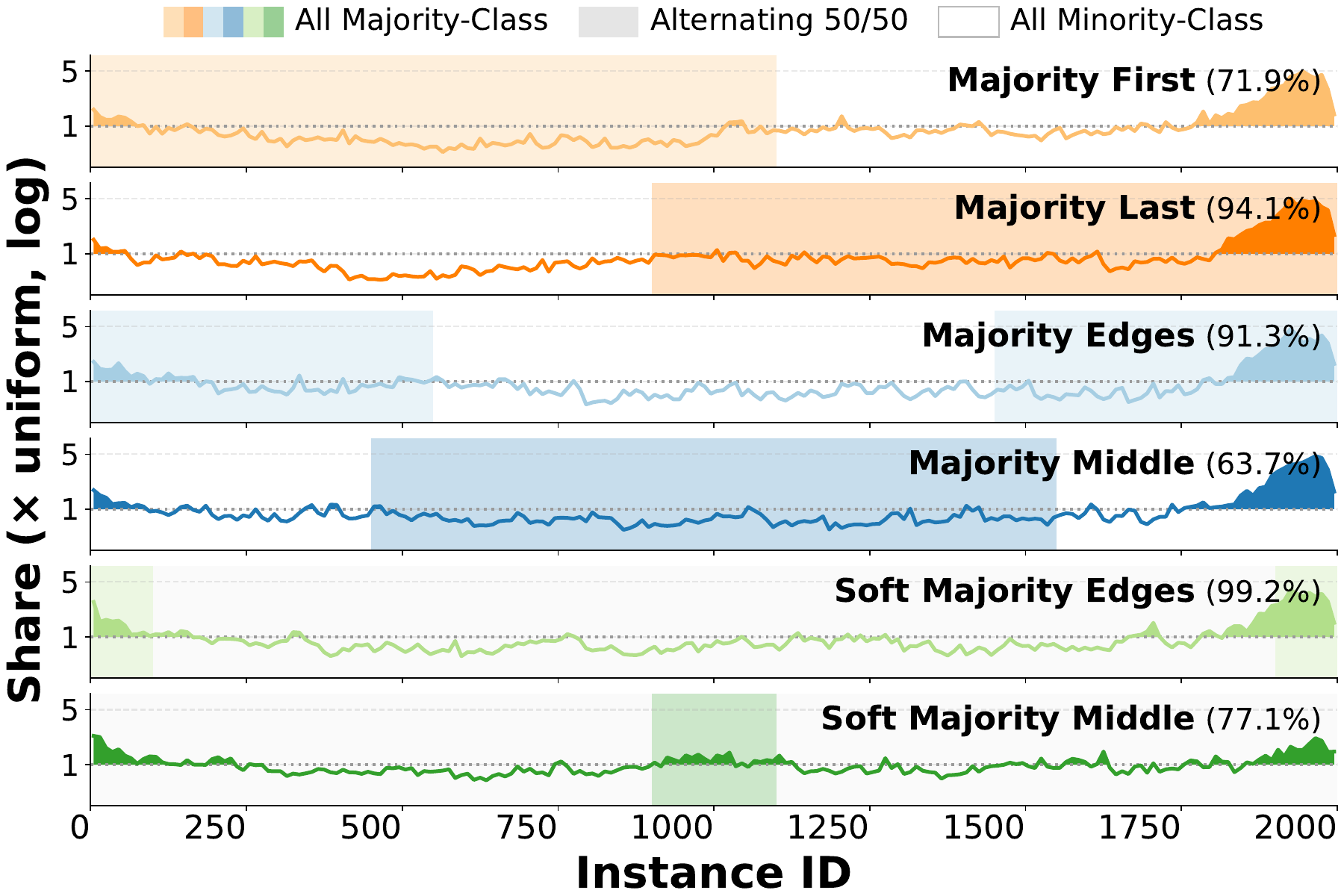}
  \caption{\re{IxG position profiles for \textit{Gemma 4 31B} on \textit{NER}, separated by order pattern.}}
  \label{fig:ixg_gemma4_31b_ner}
\end{figure}

\begin{figure}[t]
  \centering
  \includegraphics[width=0.9\columnwidth]{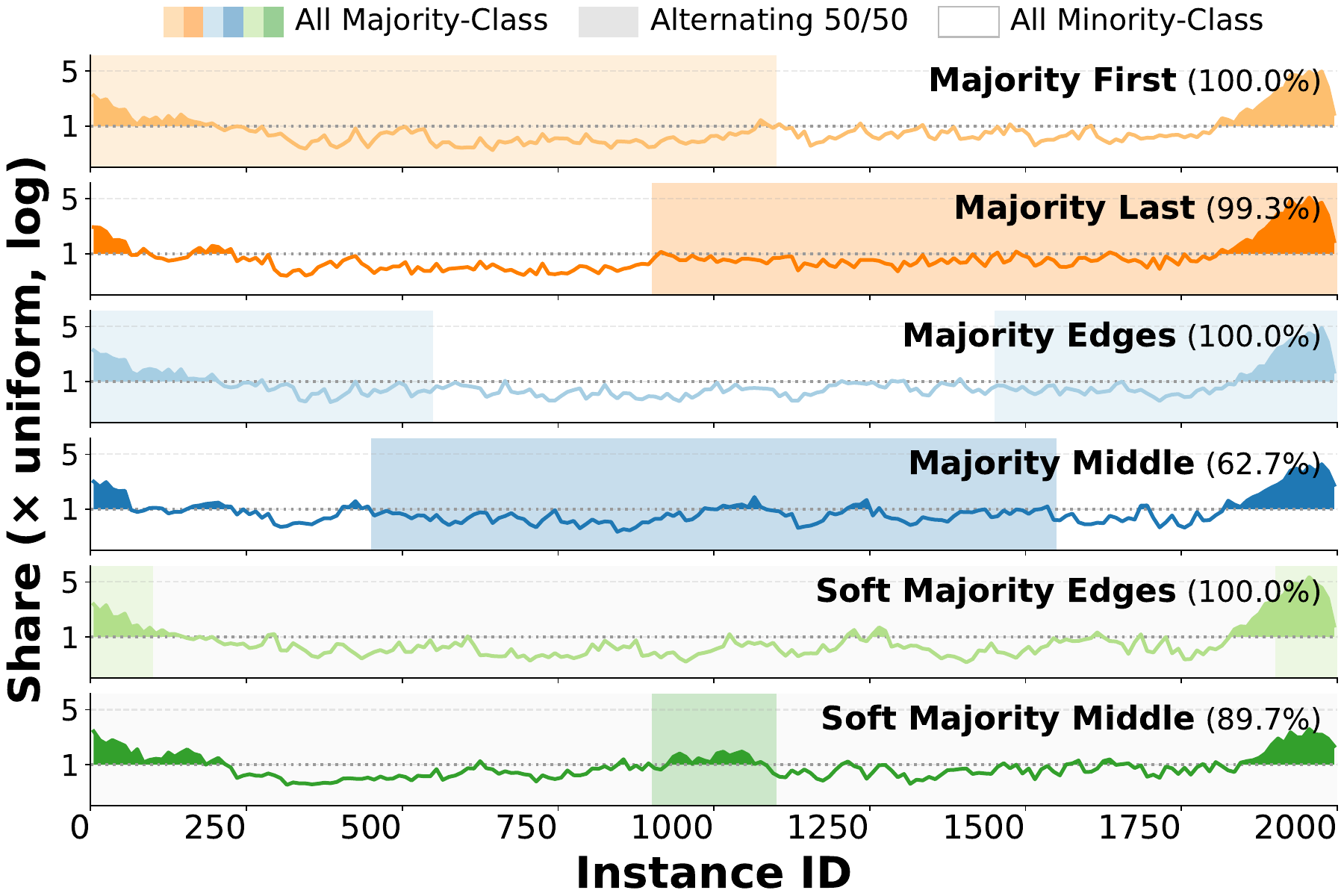}
  \caption{\re{IxG position profiles for \textit{Gemma 4 31B} on \textit{Sentiment}, separated by order pattern.}}
  \label{fig:ixg_gemma4_31b_sentiment}
\end{figure}

\begin{figure}[t]
  \centering
  \includegraphics[width=0.9\columnwidth]{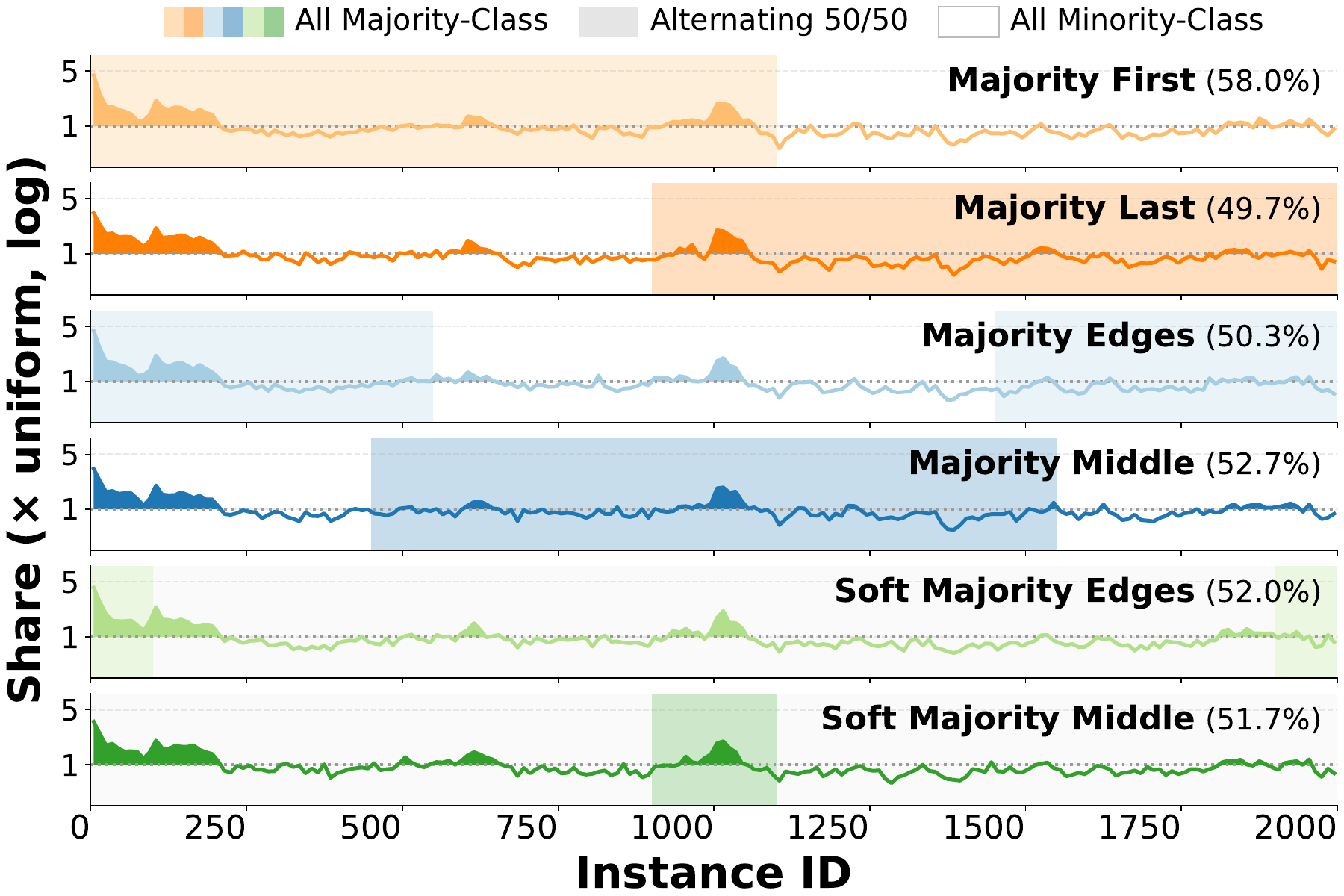}
  \caption{\re{IxG position profiles for \textit{Qwen3-4B-Instruct} on \textit{Math}, separated by order pattern.}}
  \label{fig:IxG_qwen3_4b_math}
\end{figure}

\begin{figure}[t]
  \centering
  \includegraphics[width=0.9\columnwidth]{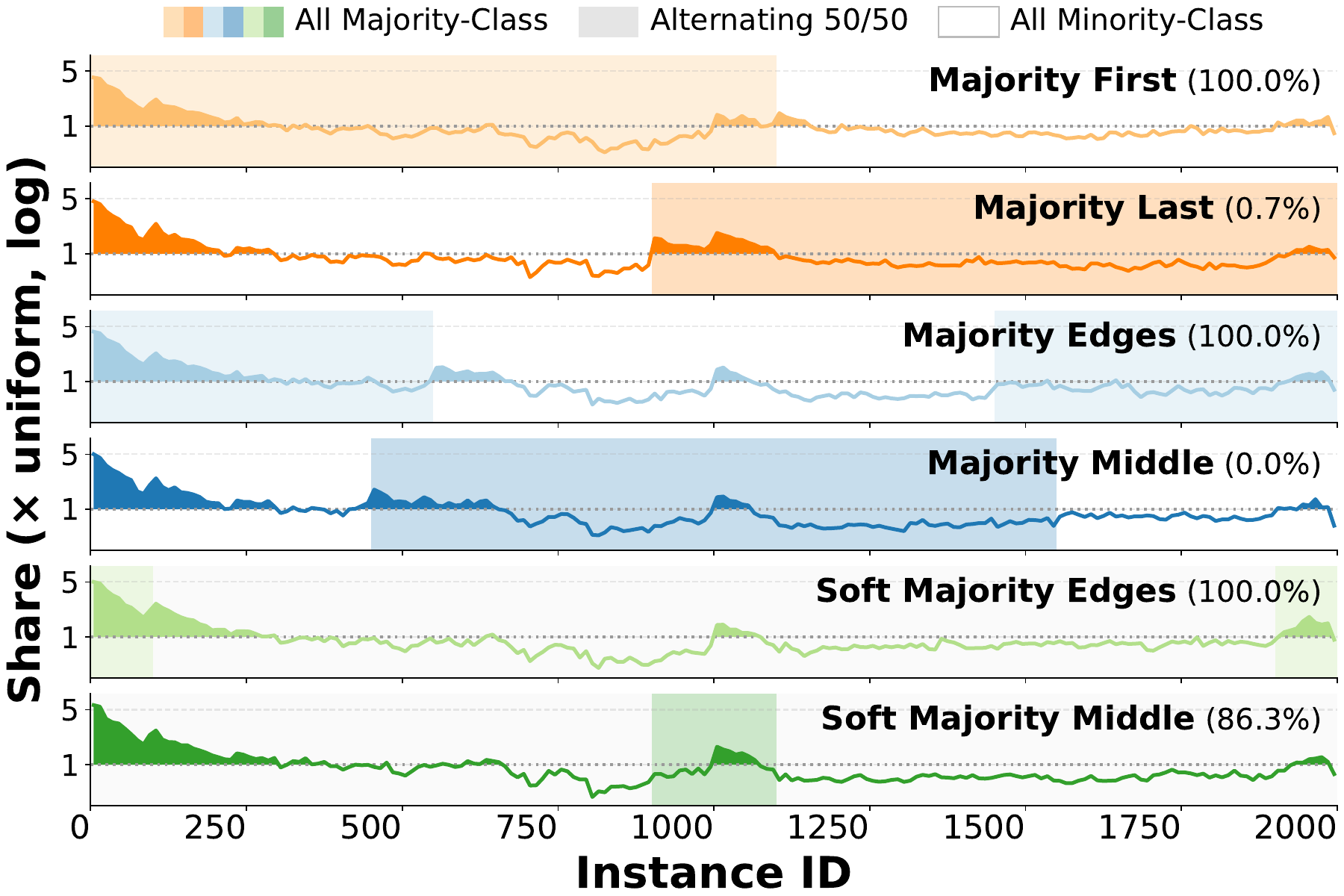}
  \caption{\re{IxG position profiles for \textit{Qwen3-4B-Instruct} on \textit{NER}, separated by order pattern.}}
  \label{fig:IxG_qwen3_4b_ner}
\end{figure}

\begin{figure}[t]
  \centering
  \includegraphics[width=0.9\columnwidth]{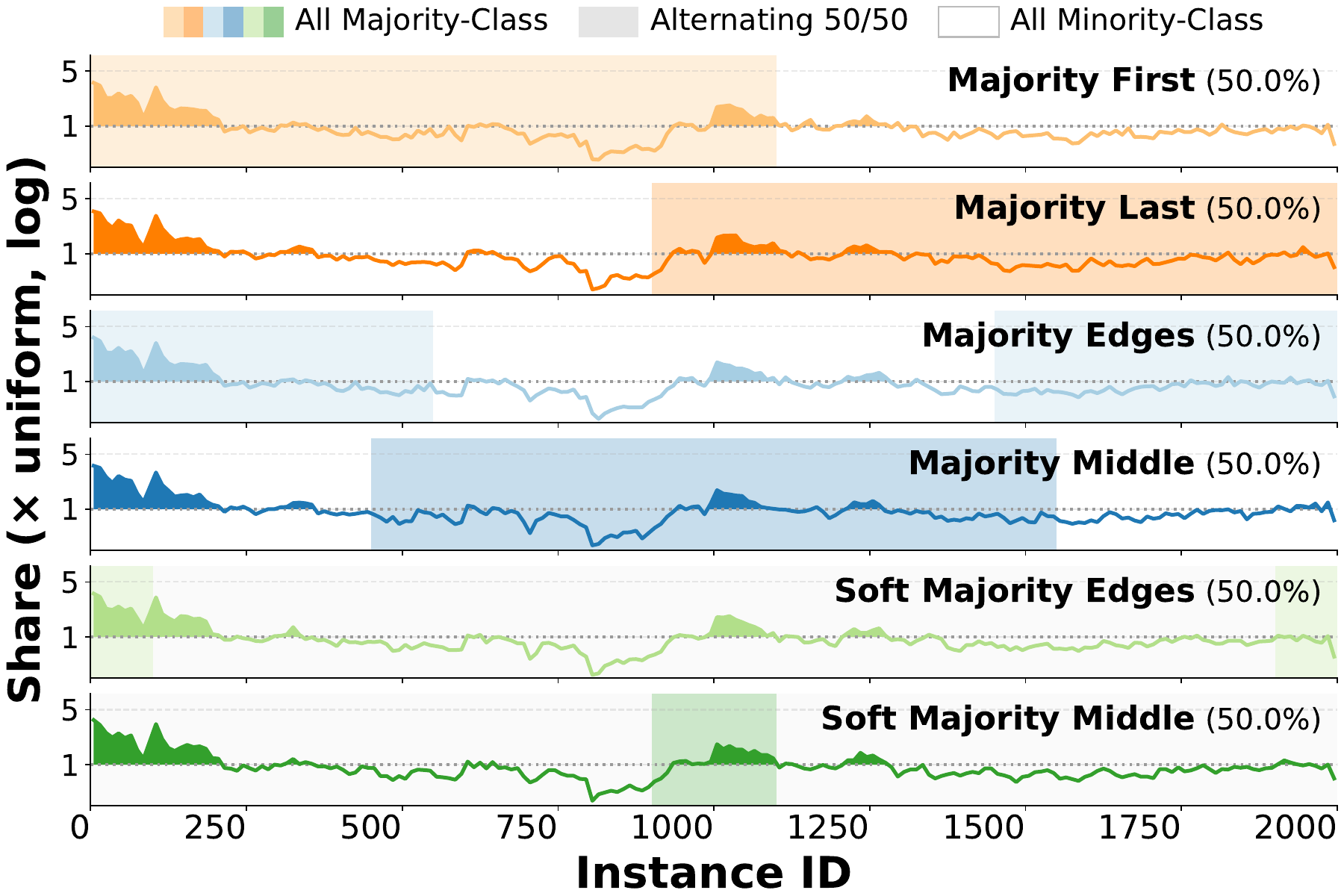}
  \caption{\re{IxG position profiles for \textit{Qwen3-4B-Instruct} on \textit{Occurrence}, separated by order pattern.}}
  \label{fig:IxG_qwen3_4b_occurrence}
\end{figure}

\begin{figure}[t]
  \centering
  \includegraphics[width=0.9\columnwidth]{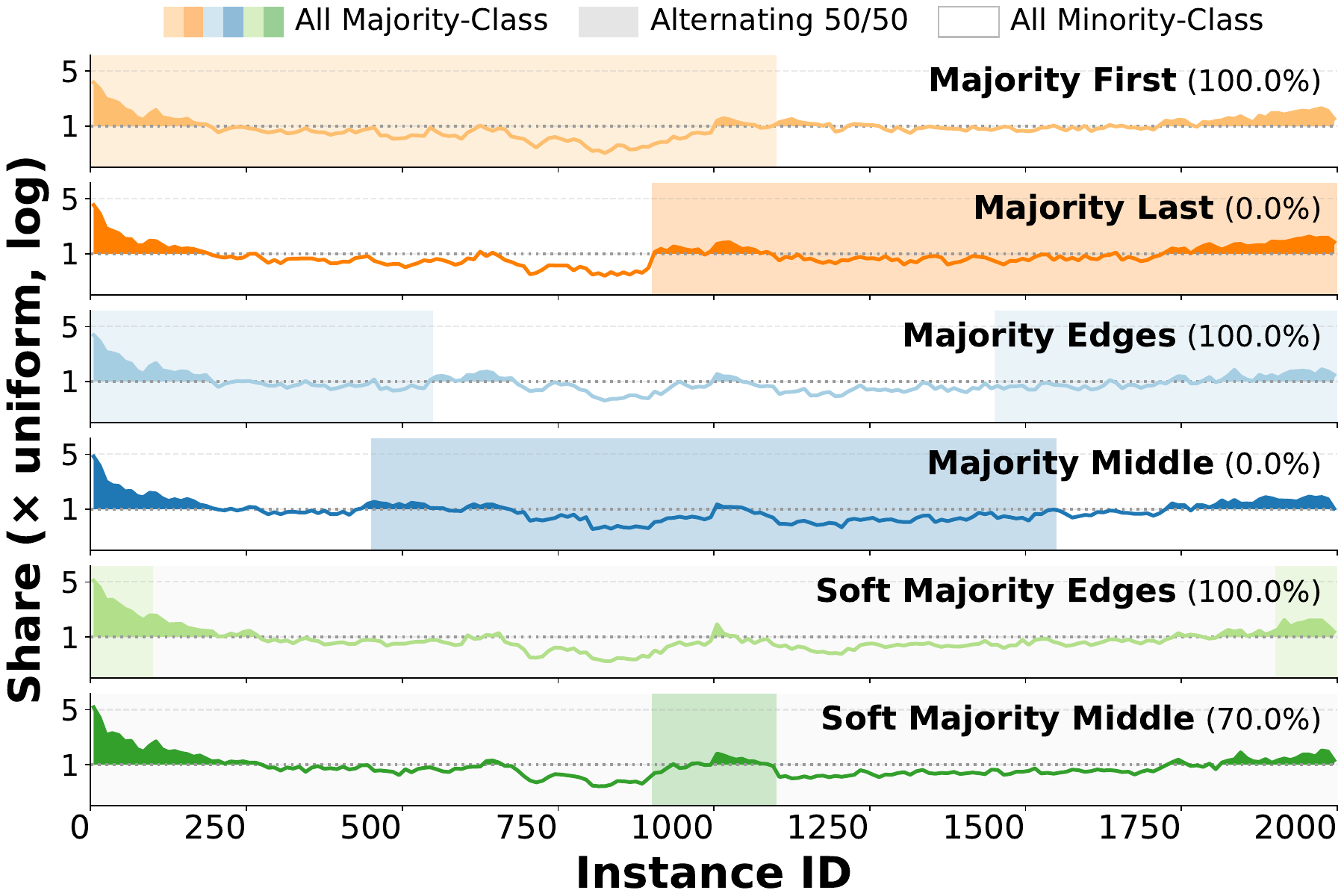}
  \caption{\re{IxG position profiles for \textit{Qwen3-4B-Instruct} on \textit{Sentiment}, separated by order pattern.}}
  \label{fig:IxG_qwen3_4b_sentiment}
\end{figure}

\begin{figure}[t]
  \centering
  \includegraphics[width=0.9\columnwidth]{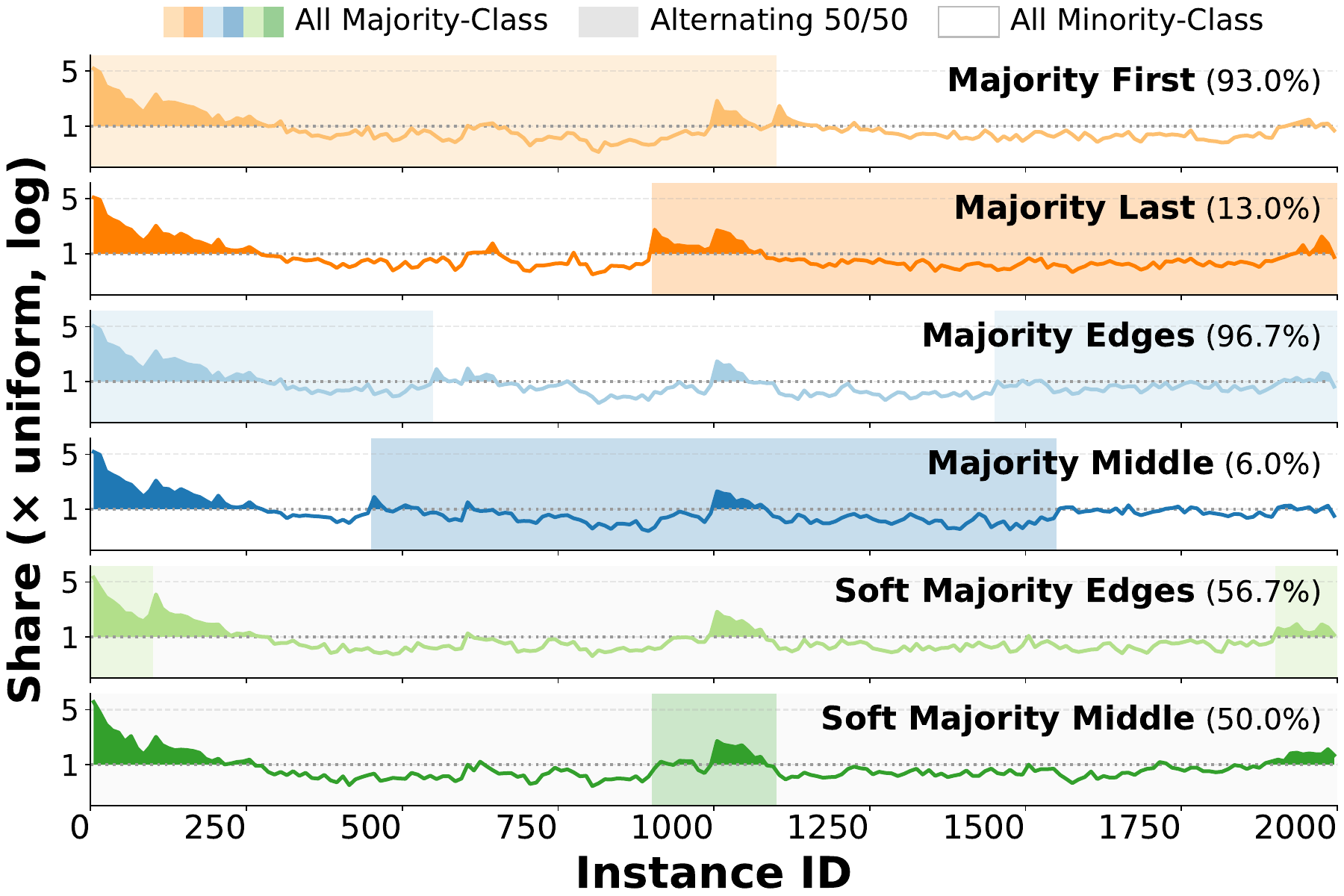}
  \caption{\re{IxG position profiles for \textit{Qwen3-4B-Instruct} on \textit{WSD}, separated by order pattern.}}
  \label{fig:IxG_qwen3_4b_wsd}
\end{figure}

\re{Figures~\ref{fig:ixg_gemma4_26b_a4b_ner}--\ref{fig:IxG_qwen3_4b_wsd}
% disaggregate the \textit{Gemma} profiles in Figure~\ref{fig:interp-position-profiles-both} by task and order pattern; the corresponding \textit{Qwen3-4B-Instruct} profiles appear in Figure~\ref{fig:interp-position-profiles-pattern-qwen} TODOTODO HERE ALL PATTERNS.
\last{show all six structured order patterns separately for each model and task. They include \textit{NER} and \textit{Sentiment} for the two \textit{Gemma} models and all five tasks for \textit{Qwen3-4B-Instruct}, extending the results in Figures~\ref{fig:interp-position-profiles-both} and~\ref{fig:interp-position-profiles-pattern-qwen}.}
Boundary preferences vary across models and patterns, while below-uniform attribution across substantial portions of the middle remains common.}
\last{The additional \textit{Math} and \textit{Occurrence} profiles for \textit{Qwen3-4B-Instruct} retain an emphasis on early positions and a local peak around instance 1,000. Yet their accuracy varies much less across patterns than on \textit{NER} and \textit{Sentiment}, illustrating that similar positional profiles can accompany different accuracy levels.}

\last{All profiles use the expanded \textit{core} orderings and ten-instance bins from Section~\ref{sec:gradient_attribution}. Dotted lines mark the uniform reference at $1$ on a logarithmic $y$-axis. Background bands identify majority-label, minority-label and alternating regions. Curves use only outputs with available attribution and their fills are visual aids, not uncertainty intervals.}
\last{Accuracy annotations here and in Figures~\ref{fig:interp-position-profiles-both}--\ref{fig:interp-position-profiles-pattern-qwen} use all parseable generated answers. These subsets coincide for \textit{Qwen3-4B-Instruct}, but differ for some \textit{Gemma} conditions because valid outputs beyond the two candidate labels have no attribution.}

\subsection{\re{Input $\times$ Gradient in \last{the} Generalisability Setting}}
\label{append:inter_ixg_generalisability}
\begin{figure}[t]
    \centering

    \begin{subfigure}{0.95\columnwidth}
        \centering
        \includegraphics[width=\linewidth]{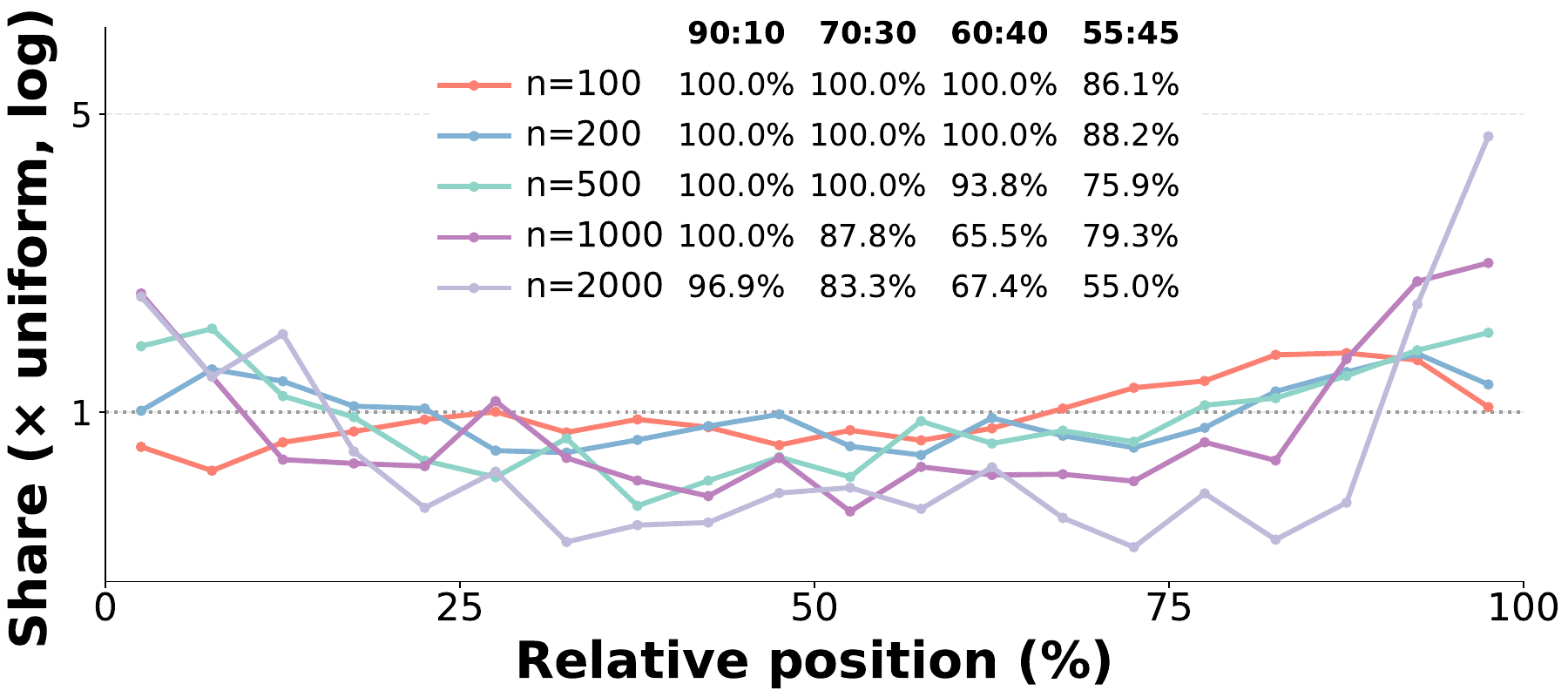}
        \caption{Gemma 4 26B A4B}
    \end{subfigure}

    \vspace{0.3em}

    \begin{subfigure}{0.95\columnwidth}
        \centering
        \includegraphics[width=\linewidth]{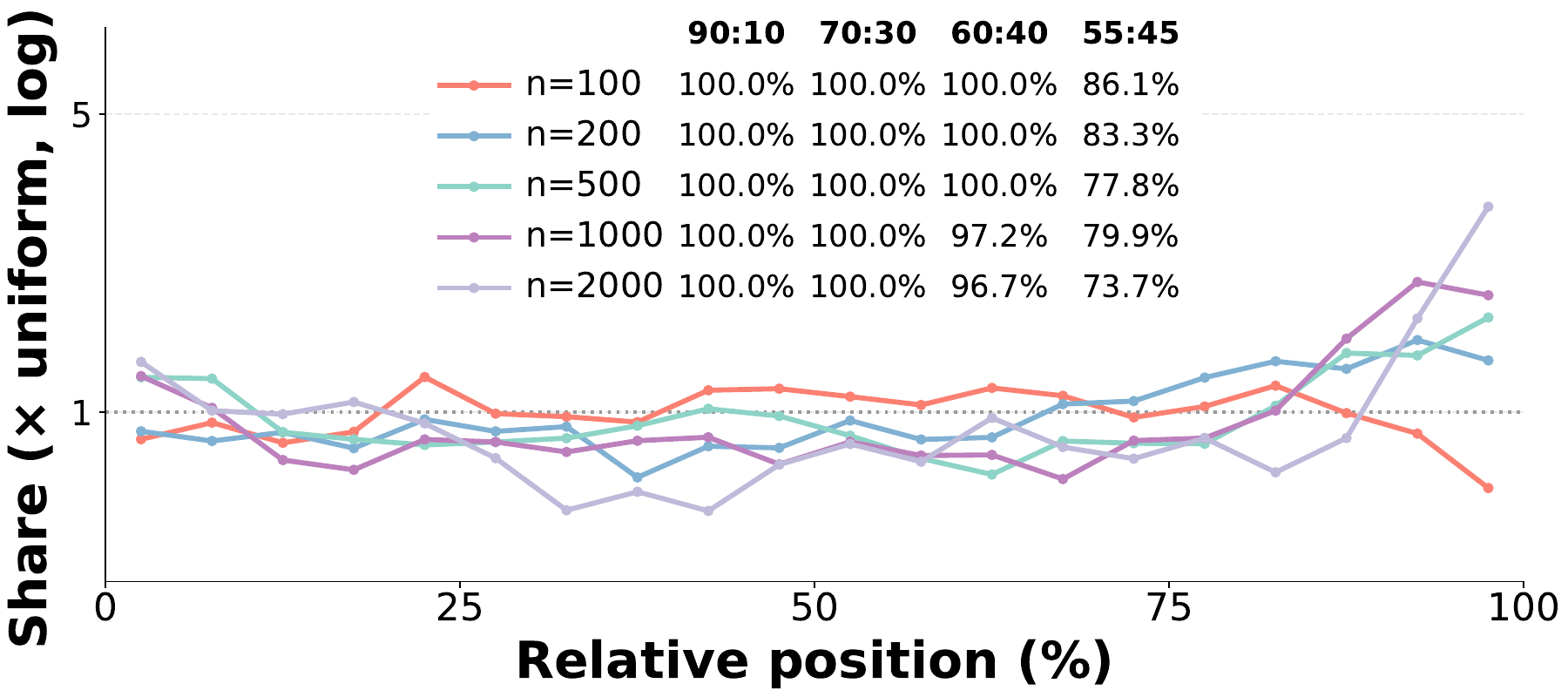}
        \caption{Gemma 4 31B}
    \end{subfigure}

    \vspace{0.3em}

    \begin{subfigure}{0.95\columnwidth}
        \centering
        \includegraphics[width=\linewidth]{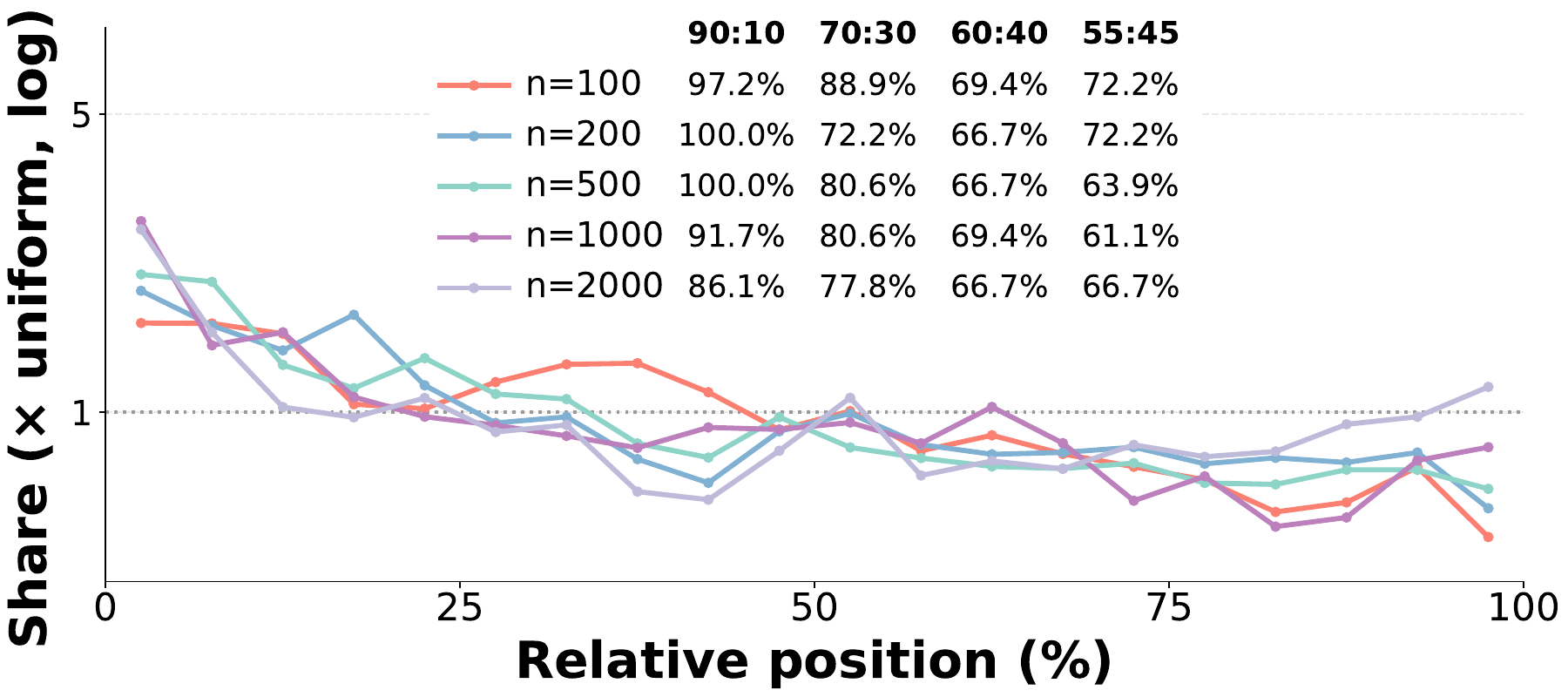}
        \caption{Qwen3-4B-Instruct}
    \end{subfigure}

    \caption{\re{IxG position profiles \last{by instance count}, pooled over valid replays from \textit{NER}, \textit{Sentiment}, all
    % the
    seven order patterns, and
    % label distribution.
    \last{four label distributions. Each point averages a 5\% position bin. The inset reports accuracy (\%) on the attribution subset for each count and distribution. The dotted line at $1$ is the uniform reference, and the $y$-axis is logarithmic.}}}
    \label{fig:base_ixg_by_count}
\end{figure}

\begin{figure}[t]
    \centering

    \begin{subfigure}{0.95\columnwidth}
        \centering
        \includegraphics[width=\linewidth]{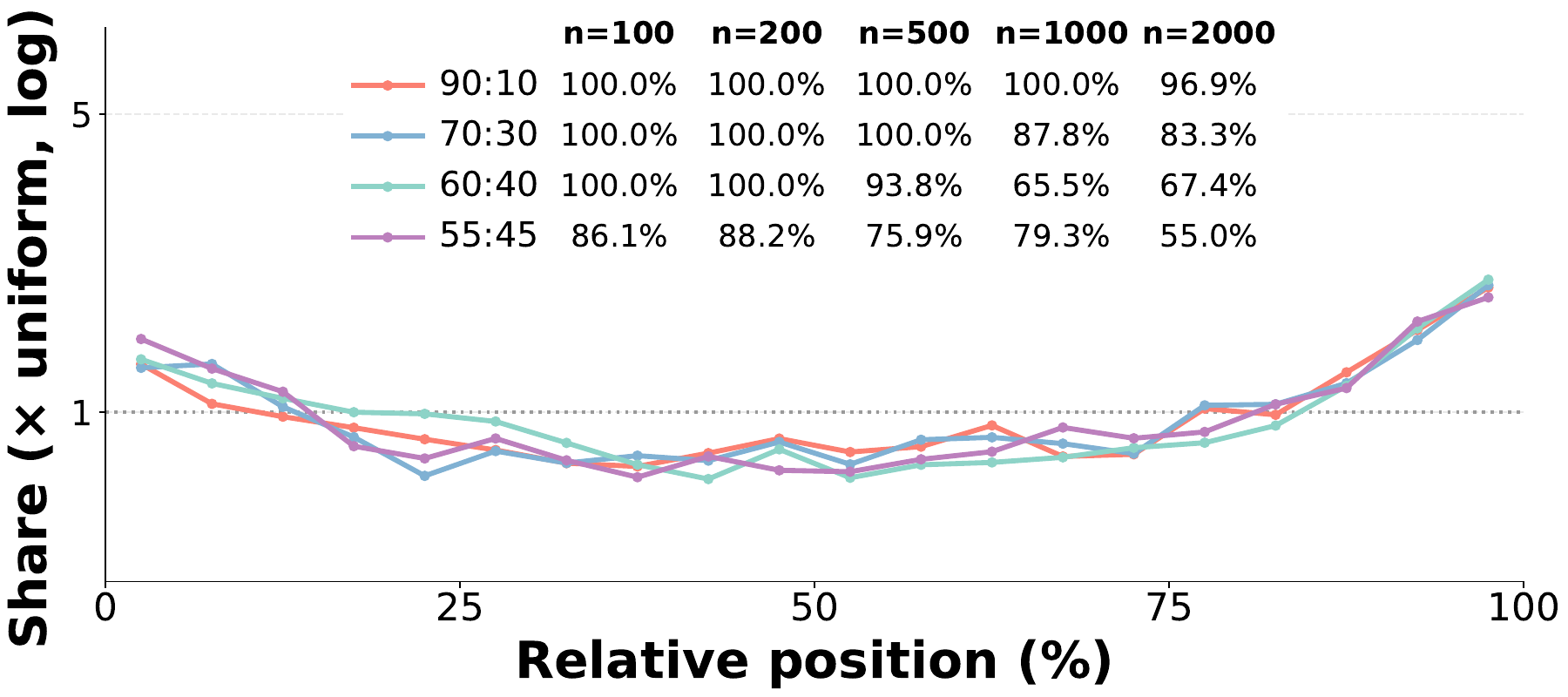}
        \caption{Gemma 4 26B A4B}
    \end{subfigure}

    \vspace{0.3em}

    \begin{subfigure}{0.95\columnwidth}
        \centering
        \includegraphics[width=\linewidth]{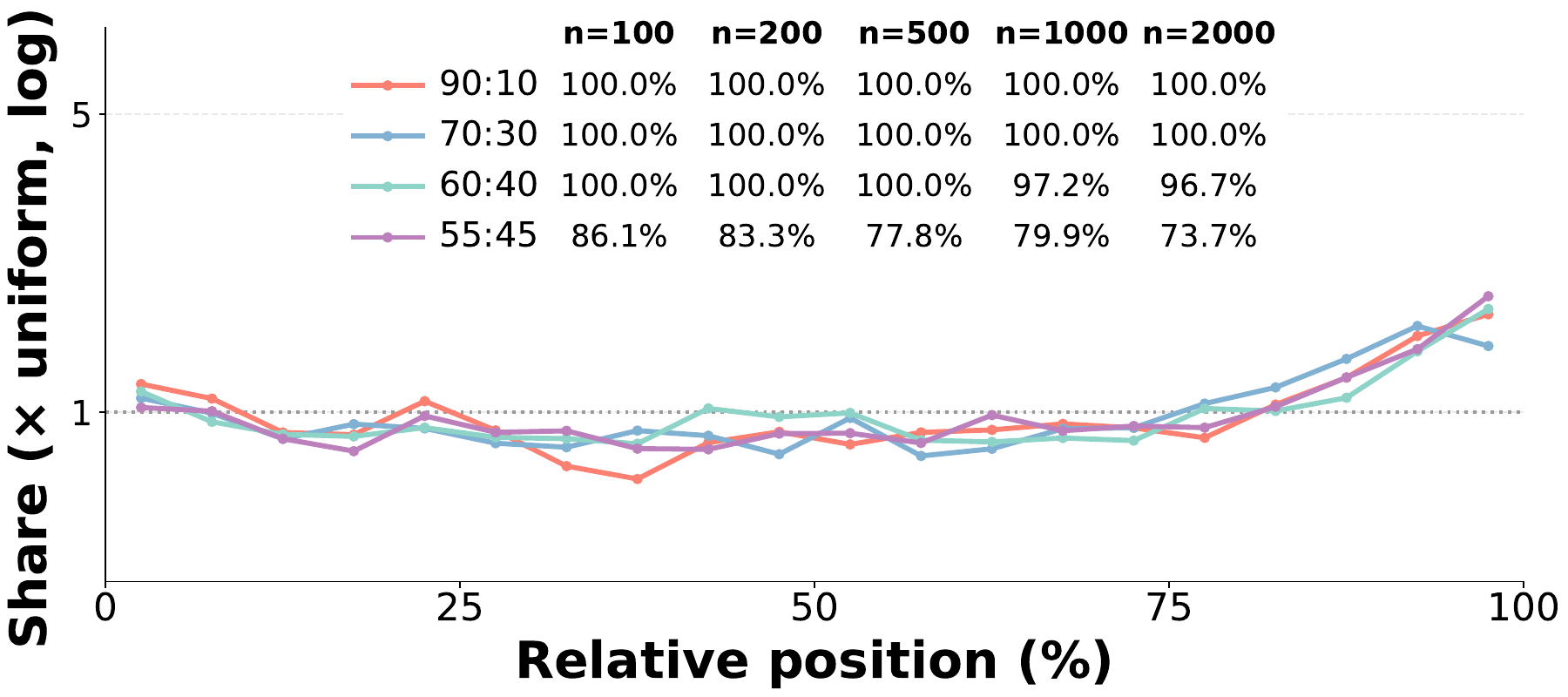}
        \caption{Gemma 4 31B}
    \end{subfigure}

    \vspace{0.3em}

    \begin{subfigure}{0.95\columnwidth}
        \centering
        \includegraphics[width=\linewidth]{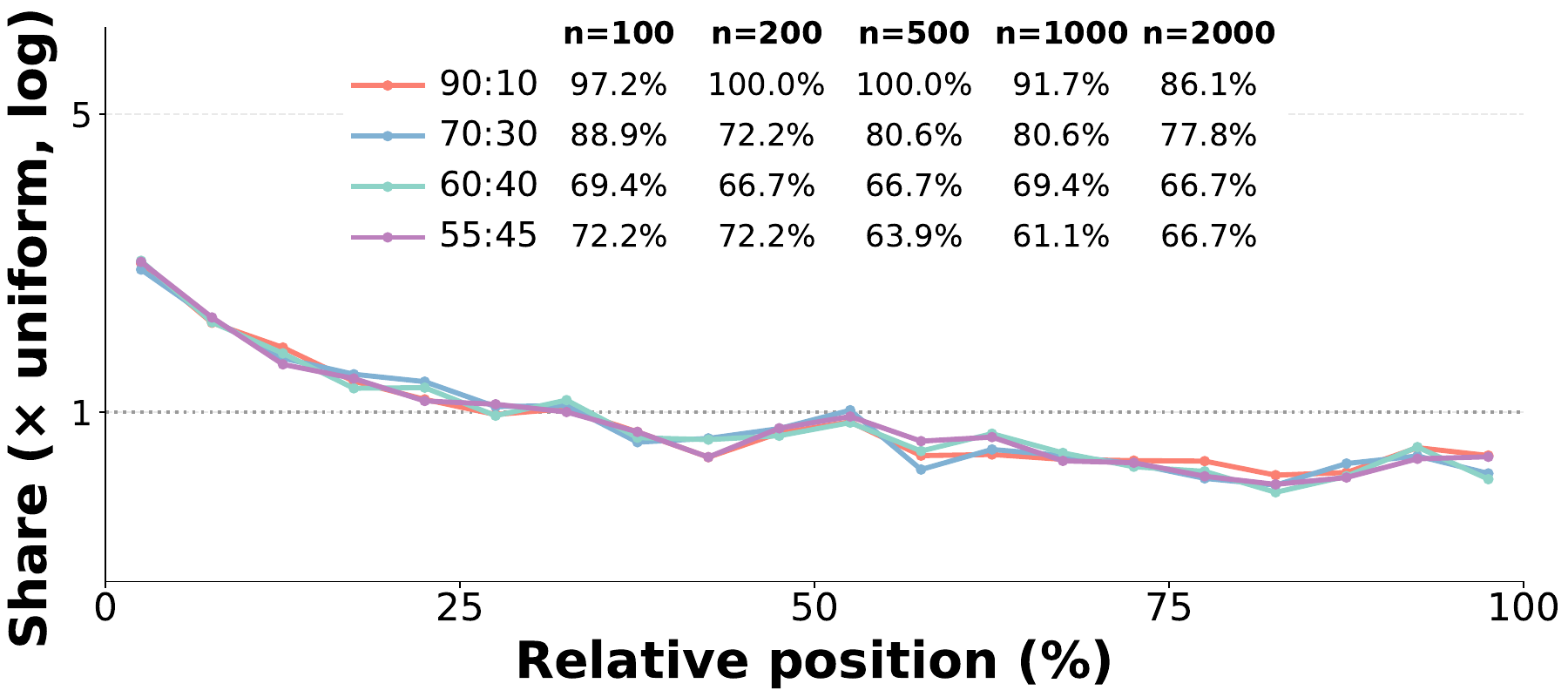}
        \caption{Qwen3-4B-Instruct}
    \end{subfigure}

    \caption{\re{IxG position profiles \last{by label distribution}, pooled over valid replays from \textit{NER}, \textit{Sentiment}, all
    % the
    seven order patterns, and
    % instance count.
    \last{five instance counts. The inset reports accuracy (\%) on the attribution subset for each distribution and count. Binning, axes and normalisation follow Figure~\ref{fig:base_ixg_by_count}.}}}
    \label{fig:base_ixg_by_label}
\end{figure}

% Figure~\ref{fig:base_ixg_by_count} and Figure~\ref{fig:base_ixg_by_label}
\last{We apply the same IxG measure to the \textit{generalisability} setting for the three models on \textit{NER} and \textit{Sentiment}. Figure~\ref{fig:base_ixg_by_count} compares instance counts, pooling across label distributions. Figure~\ref{fig:base_ixg_by_label} compares label distributions, pooling across instance counts. Both include all seven order patterns and use relative instance position to compare inputs of different lengths.} Each task and condition has up to 18 replays.

% \last{We use selection seed 1, taking the 2,000-instance, 55/45 condition from the \textit{core} setting and the other 19 conditions from the \textit{generalisability} setting. Each task and condition has up to 18 replays: two majority labels, three random orderings (order seeds 0--2) and one ordering for each structured pattern, whose within-label order is fixed by the selection seed.}

\last{After length normalisation and within-replay scaling by $N$, we average scores within 20 equal relative-position bins, each containing $N/20$ consecutive instances. For each curve, we first average replays equally within each task over the conditions being pooled, then average the two task profiles equally. The accuracy insets use only these replays with available attribution.}
% , excluding ties and failed replays. This differs from the accuracy annotations for the expanded \textit{core} analysis above.}

\last{The effect of instance count is model dependent. For both \textit{Gemma} models, longer inputs generally show a stronger rise near the end, most clearly at 2,000 instances. This change is more pronounced for \textit{Gemma 4 26B A4B}, whose profiles also lie further below the uniform reference across much of the middle at larger instance counts. The curves do not follow a strictly monotonic ordering at every position. \textit{Qwen3-4B-Instruct} remains more strongly weighted towards the beginning across all instance counts, with a smaller rise near the end at 2,000.}

\last{By contrast, the profiles for different label distributions largely overlap within each model. Both \textit{Gemma} models retain their increase near the end, while \textit{Qwen3-4B-Instruct} retains its emphasis on early positions. The accuracy annotations nevertheless show poorer performance for more balanced label distributions in many settings, especially for \textit{Qwen3-4B-Instruct}. Thus, the pooled positional profiles can remain similar even when aggregation accuracy changes.}

\subsection{\re{Input $\times$ Gradient under Prompt Variants}}
\label{append:inter_ixg_prompts}
\begin{figure}[t]
  \centering
  \includegraphics[width=0.9\columnwidth]{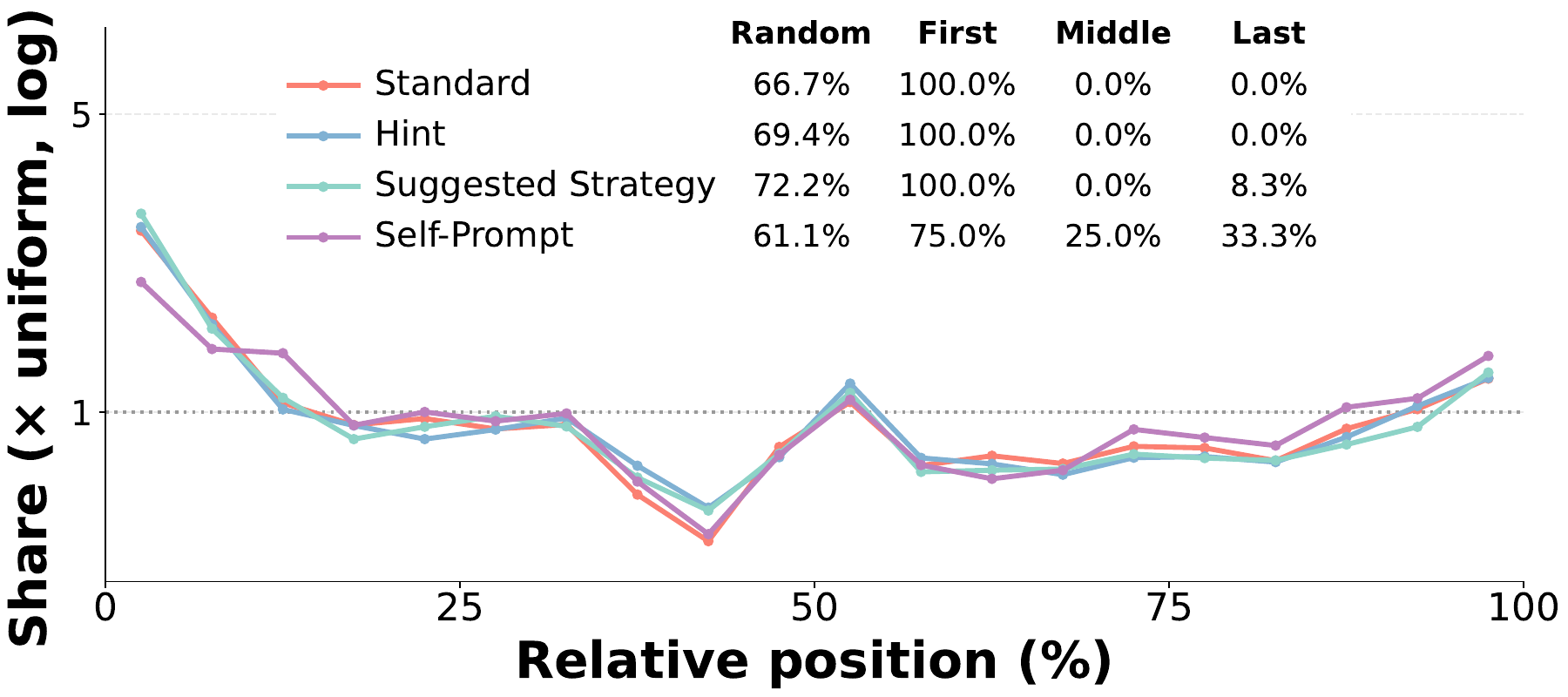}
  \caption{\re{IxG position profiles \last{for \textit{Qwen3-4B-Instruct}}, pooled over valid replays from \textit{NER}, \textit{Sentiment} and all
  % the
  seven order patterns,
  % for each prompt variant.
  \last{for the standard prompt and three variants. The inset reports accuracy (\%) on the same outputs for four representative patterns. Binning, axes and normalisation follow Figure~\ref{fig:base_ixg_by_count}.}}}
  \label{fig:IxG_qwen3_4b_prompt_variant}
\end{figure}

% Figure~\ref{fig:IxG_qwen3_4b_prompt_variant}
\last{Figure~\ref{fig:IxG_qwen3_4b_prompt_variant} compares the standard prompt and three prompting variants for \textit{Qwen3-4B-Instruct} on \textit{NER} and \textit{Sentiment}. Each prompt contributes 108 valid replays with attribution.
% across three selection seeds (0--2), two tasks, two majority labels and nine orderings per instance set.
The runs follow Section~\ref{sec:prompt_variant_can}, relative-position bins and task averaging follow Appendix~\ref{append:inter_ixg_generalisability}.}

\last{All four profiles place greater weight near the beginning and share a local increase around the middle and a smaller rise near the end. Their broad shapes remain similar despite changes in accuracy. In particular, self-prompting has lower accuracy under \textit{majority-first} and higher accuracy under \textit{majority-middle} and \textit{majority-last}, without a comparable change in the pooled attribution profile.}

\end{document}